\documentclass{article}
\usepackage{ijcai24}

\usepackage{times}
\usepackage{soul}
\usepackage{url}
\usepackage[hidelinks]{hyperref}
\usepackage[utf8]{inputenc}
\usepackage[small]{caption}
\usepackage{graphicx}
\usepackage{amsmath}
\usepackage{amsthm}
\usepackage{booktabs}
\usepackage{algorithm}
\usepackage{algorithmic}
\usepackage[switch]{lineno}
\usepackage{subcaption}
\usepackage{cleveref}
\usepackage{longtable}
\usepackage[dvipsnames]{xcolor}

\newcommand{\G}[1]{{\color{ForestGreen} \textbf{#1}}}
\newcommand{\R}[1]{{\color{BrickRed} \textit{#1}}}

\title{Trusting Semantic Segmentation Networks}

\author{
Samik Some$^1$
\And
Vinay P. Namboodiri$^2$
\affiliations
$^1$IIT Kanpur\\
$^2$University of Bath\\
\emails
samiks@iitk.ac.in,
vpn22@bath.ac.uk
}

\begin{document}

\makeatletter
\g@addto@macro\@maketitle{
\begin{figure}[H]
    \setlength{\linewidth}{\textwidth}
    \setlength{\hsize}{\textwidth}
    \begin{subfigure}{0.247\textwidth}
        \centering
        \includegraphics[width=\linewidth]{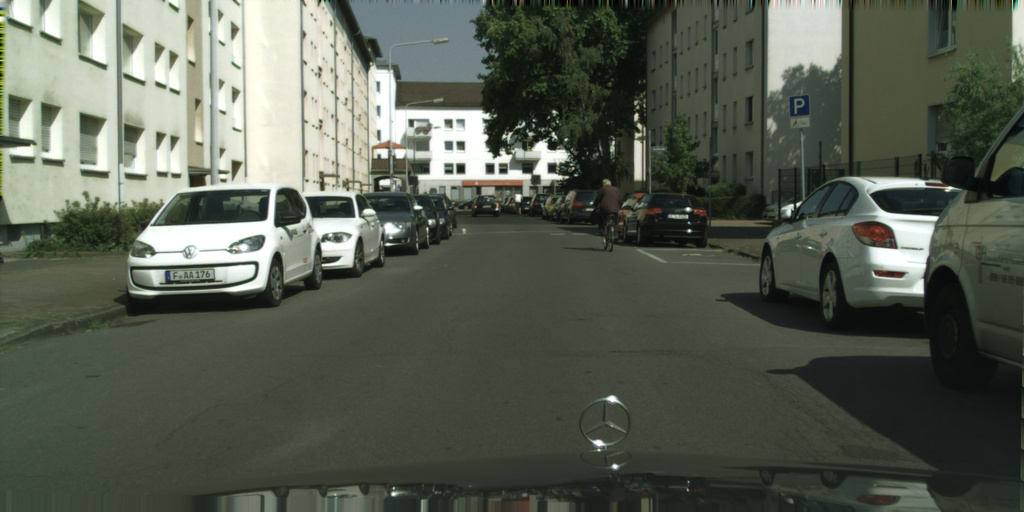}
    \end{subfigure}
    \begin{subfigure}{0.247\textwidth}
        \centering
        \includegraphics[width=\linewidth]{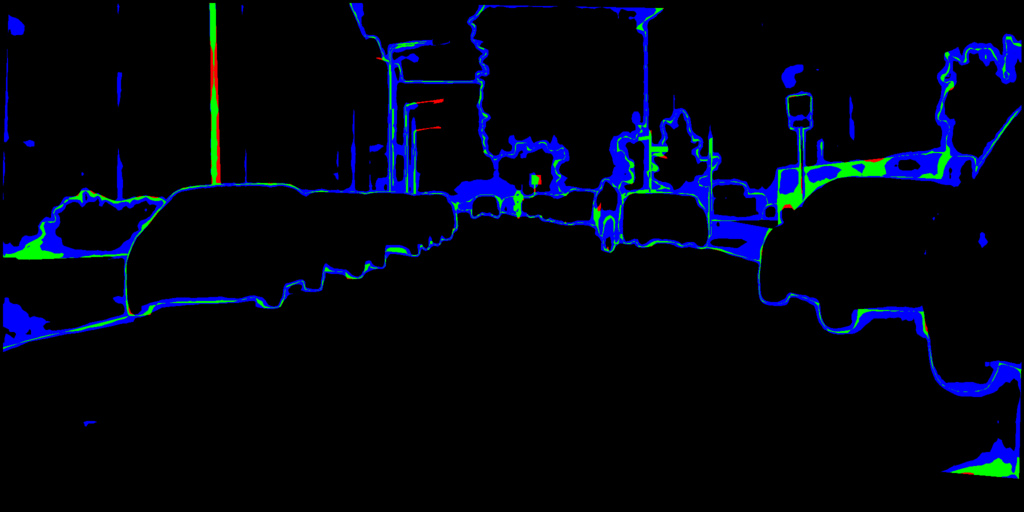}
    \end{subfigure}
    \begin{subfigure}{0.247\textwidth}
        \centering
        \includegraphics[width=\linewidth]{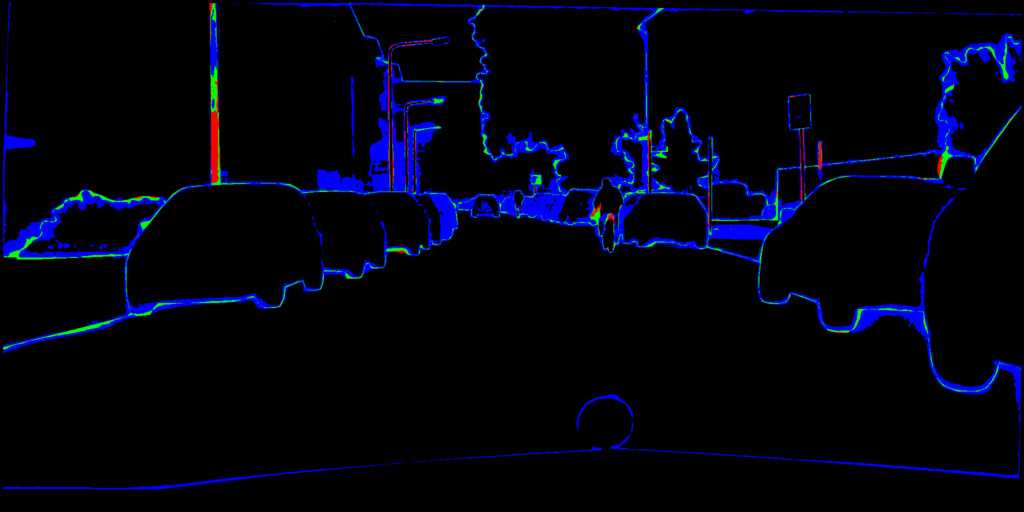}
    \end{subfigure}
    \begin{subfigure}{0.247\textwidth}
        \centering
        \includegraphics[width=\linewidth]{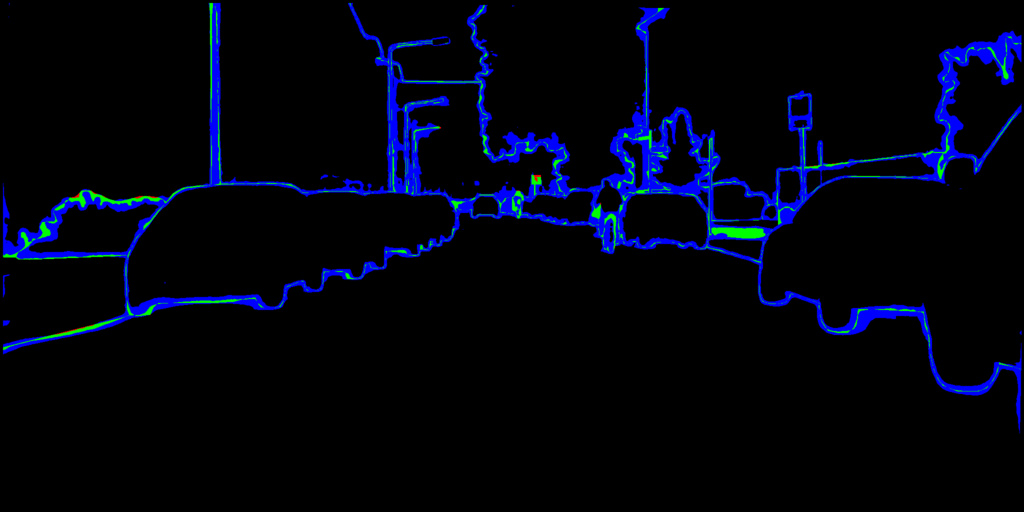}
    \end{subfigure} \\
    \begin{subfigure}{0.247\textwidth}
        \centering
        \includegraphics[width=\linewidth]{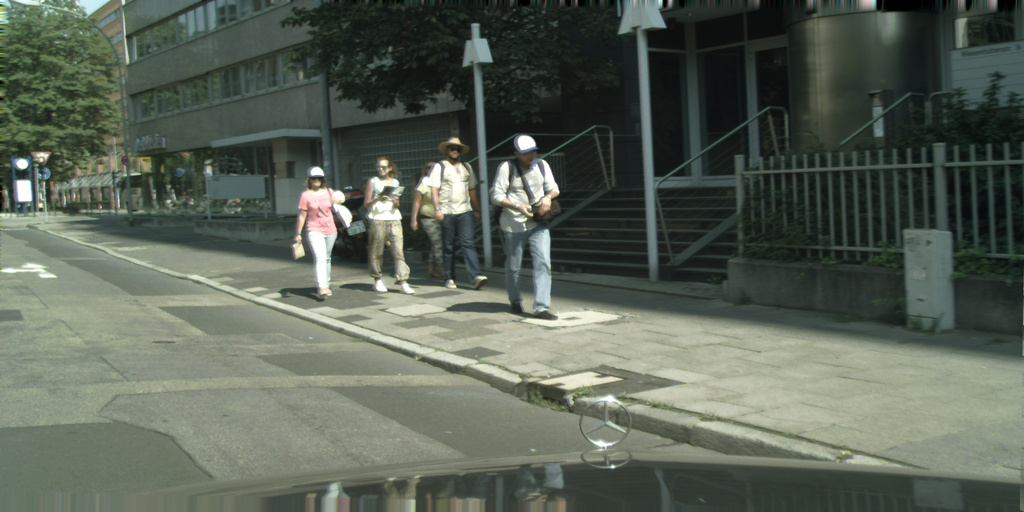}
    \end{subfigure}
    \begin{subfigure}{0.247\textwidth}
        \centering
        \includegraphics[width=\linewidth]{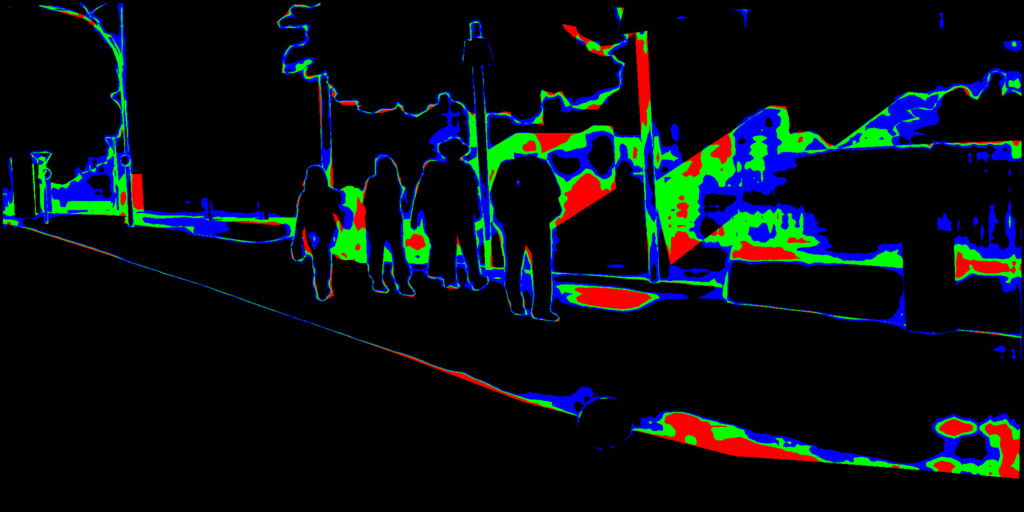}
    \end{subfigure}
    \begin{subfigure}{0.247\textwidth}
        \centering
        \includegraphics[width=\linewidth]{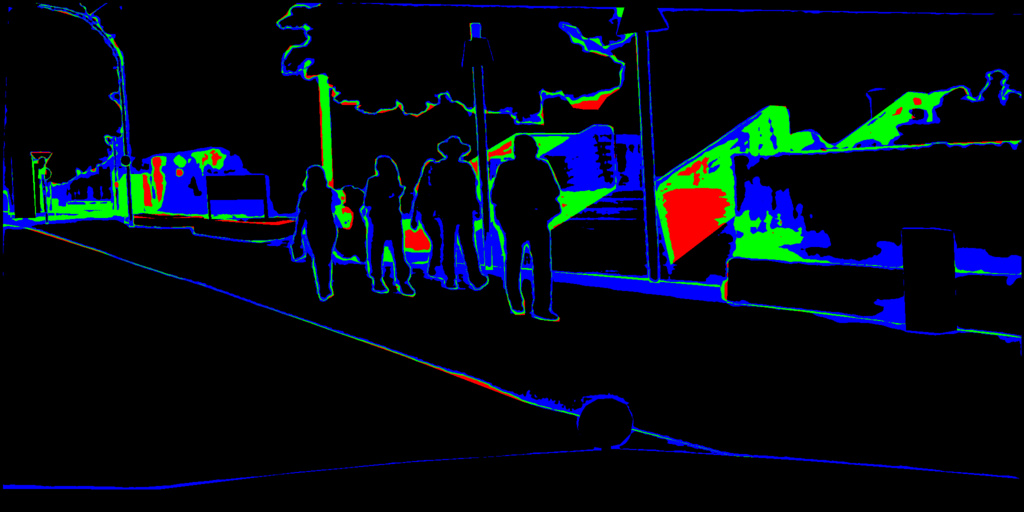}
    \end{subfigure}
    \begin{subfigure}{0.247\textwidth}
        \centering
        \includegraphics[width=\linewidth]{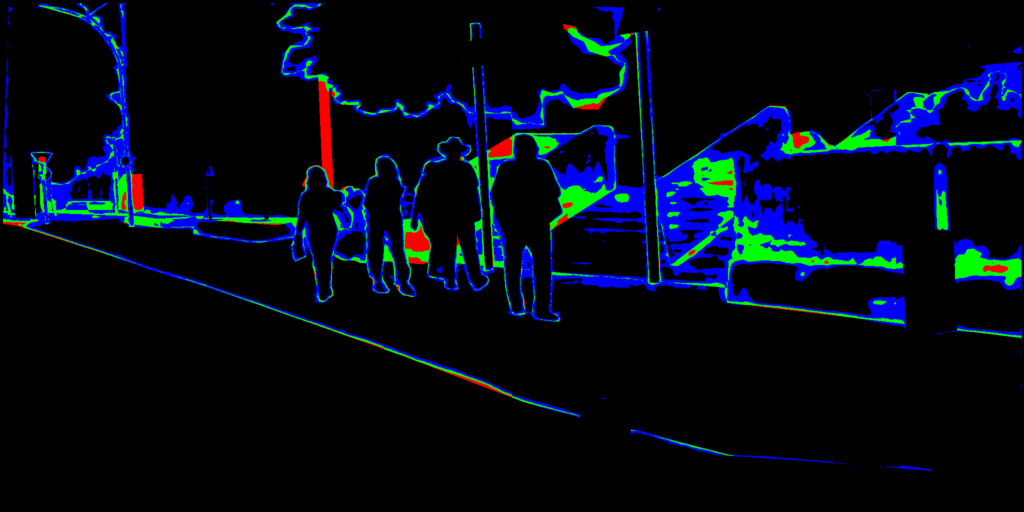}
    \end{subfigure}
    \caption{A couple of examples show how well the entropy of the predicted outputs correlates with misclassified pixels. Pixels in green indicate misclassified pixels with high entropy, red shows misclassified pixels with low entropy, and blue shows correctly classified pixels with high entropy. Images in each row show the input and outputs from DRN, OneFormer and SegFormer, respectively. We note that entropy captures misclassified pixels with high recall rates.}
    \label{fig:visual}
\end{figure}
}
\makeatother

\maketitle

\begin{abstract}
Semantic segmentation has become an important task in computer vision with the growth of self-driving cars, medical image segmentation, etc. Although current models provide excellent results, they are still far from perfect and while there has been significant work in trying to improve the performance, both with respect to accuracy and speed of segmentation, there has been little work which analyses the failure cases of such systems. In this work, we aim to provide an analysis of how segmentation fails across different models and consider the question of whether these can be predicted reasonably at test time. To do so, we explore existing uncertainty-based metrics and see how well they correlate with misclassifications, allowing us to define the degree of trust we put in the output of our prediction models. Through several experiments on three different models across three datasets, we show that simple measures such as entropy can be used to capture misclassification with high recall rates.
\end{abstract}

\section{Introduction}
Semantic segmentation is defined as a task which involves taking an image and labelling each pixel of the image as belonging to one of a set of predefined classes. With the advent of self-driving cars, path-finding robots, and various other such tasks, semantic segmentation has emerged as an important task in the field of computer vision, as all of these applications rely on being able to segment, parse and understand the scene they see in order to take appropriate action. Semantic segmentation is even used in the medical field to segment areas of medical images for analysis. When used for such critical tasks, it is of utmost importance that the models we propose provide us with accurate results or some confidence metrics which can be used to rely on their outputs.

Most of the earlier approaches to semantic segmentation involve using Convolutional Neural Networks (CNN) to extract features and predict the segmentation class for each pixel. Examples of such approaches include \cite{chen2014semantic,yu2017dilated,chen2017deeplab}. Some of these approaches also utilise Conditional Random Fields (CRF) to improve their results further. More recent approaches use Transformer-based architectures such as \cite{cheng2021per,xie2021segformer,jain2023semask,jain2023oneformer}. Several of these approaches also utilise masked predictions for each class rather than simple pixel-wise multi-class predictions.

Although we now have a wealth of approaches that target semantic segmentation, almost none provide any form of uncertainty quantification for their predictions, leaving us with only the softmax probability as a confidence metric to trust their decisions. This is especially true during test time since we cannot judge the model against ground truth predictions. For example, consider the case when there is a domain shift in the inputs during test time; how do we trust the output of our network without any form of confidence or trust metric?

While there have been papers which look at uncertainty in deep neural networks in general \cite{gawlikowski2023survey}, including relatively new ones like \cite{vazhentsev2022uncertainty}, which specifically considers uncertainty in Transformer networks, there has not been much work focusing on estimation of trustworthiness for semantic segmentation in a black box manner. By this, we mean that without any knowledge of the architecture, at test time, we wish to know whether our model has succeeded in semantic segmentation or not. Considering the gravity of this task, we believe that there should be a study of the same for the sake of understanding these networks better.

In this paper, we first look at where these segmentation networks fail to predict the correct classes. We then ask questions such as, \textit{``Knowing where these networks fail, can we somehow predict these failures?''}, and \textit{``How well do current approaches of gauging uncertainty correlate with misclassification?''}. To this end, we perform a series of experiments, analyse the results and show that certain uncertainty methods can indeed be utilised to decide how much we should trust the output of our networks, even when we consider domain shifts in the input. For instance, when performing transfer learning on a network trained on Cityscapes to the Dark Zurich dataset, we validate that we can predict the specific cases of failure in segmentation well. We hope our work provides insight into existing segmentation networks and helps drive further research to ensure their trustworthiness. \Cref{fig:visual} shows some examples of our approach where we use entropy to judge whether a pixel is likely to be misclassified. We note that simple entropy can identify the likely misclassified regions with high recall. Further detailed analysis is presented in \Cref{sec:exp}.

\section{Related Works}
Semantic segmentation has been researched by the vision community for well over a decade, and has been well documented in survey papers such as \cite{guo2018review,hao2020brief}. However, we will primarily concern ourselves with relatively recent works that utilise deep neural networks for performing this task. Among these, most earlier works, such as \cite{chen2014semantic,long2015fully,yu2017dilated,zhao2017pyramid,chen2017deeplab,yang2018denseaspp}, are convolutional in structure and generally make use of existing Convolutional Neural Networks (CNN) such as VGG~\cite{simonyan2014very}, ResNet~\cite{he2016deep}, ResNeXt~\cite{xie2017aggregated}, for feature extraction purposes followed by combining extracted features (possibly at multiple scales) in novel ways. Generally, the classification is done at a lower resolution than the input image and is upsampled back to the original dimensions near the end of the pipeline. The improvement in performance across these approaches mainly derives from coming up with new novel ways of pooling and combining the extracted features, such as the usage of Dilated Convolutions, Pyramid Pooling and Atrous Spatial Pyramids, among others.

Newer approaches such as \cite{cheng2021per,xie2021segformer,strudel2021segmenter,gu2022multi,jain2023semask,jain2023oneformer}, are based on Transformer networks introduced in \cite{vaswani2017attention}. Initially proposed for natural language processing tasks, Transformers also proved effective in computer vision. These approaches generally follow two major pipelines for extracting features from the input. Either they utilise CNN feature extractors, much like earlier works, followed by Transformer decoders for refinement. Alternatively, they utilise ViT-like~\cite{dosovitskiy2020image} architectures where the input image is divided into patches, which are directly fed to a Transformer encoder for feature generation, followed by a Transformer decoder and segmentation head.

We also consider how to extract uncertainty information from deep neural networks and refer to some existing works. Works like Deep Ensembles~\cite{lakshminarayanan2017simple} and Bayesian Networks~\cite{blundell2015weight} allow us to measure the uncertainty of predictions relatively easily, but they require special training procedures and are not applicable to already existing networks. However, \cite{gal2016dropout} shows that Dropout layers can be used to generate multiple predictions for a given input and thus allow us to compute a certain measure of model uncertainty. Similarly, \cite{houlsby2011bayesian} proposed a metric originally meant for active learning but can also be utilised as an uncertainty measure for existing networks.

To the best of our knowledge, this type of work focusing on semantic segmentation has not been done earlier. Works such as ~\cite{jammalamadaka2012has,parikh2011finding,vazhentsev2022uncertainty} are probably closest to our work. \cite{vazhentsev2022uncertainty} proposes and evaluates several uncertainty metrics for gauging the uncertainty of Transformer-based architectures and has similar goals of misclassification detection using uncertainty; however, the paper only explores these within the setting of Transformer architectures, and while most newer segmentation architectures are Transformer-based, older CNN-based networks are still used. Thus, we focus on making our evaluation and analysis as general as possible.

\begin{figure*}[t]
    \begin{subfigure}{0.247\textwidth}
        \centering
        \includegraphics[width=\linewidth]{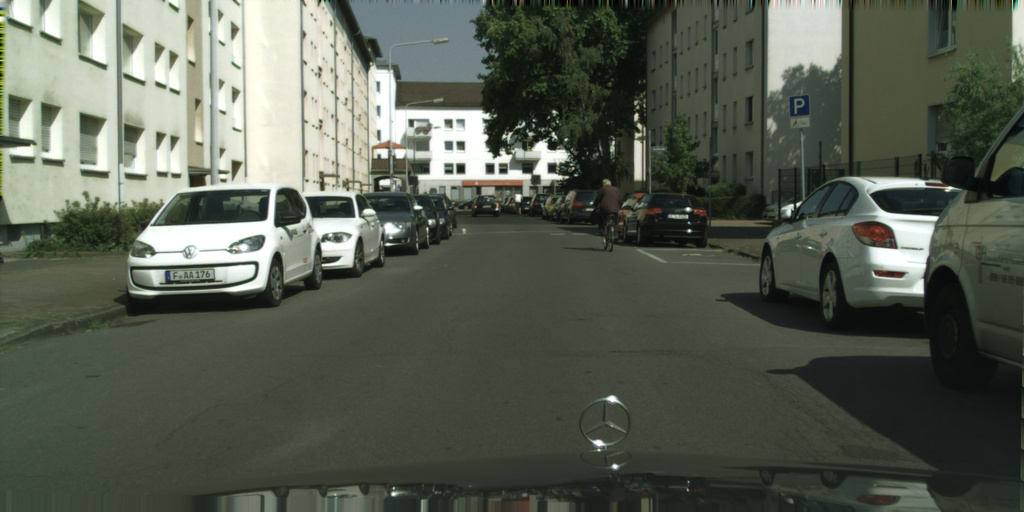}
        \caption{Image}
        \label{fig:sub:image}
    \end{subfigure}
    \begin{subfigure}{0.247\textwidth}
        \centering
        \includegraphics[width=\linewidth]{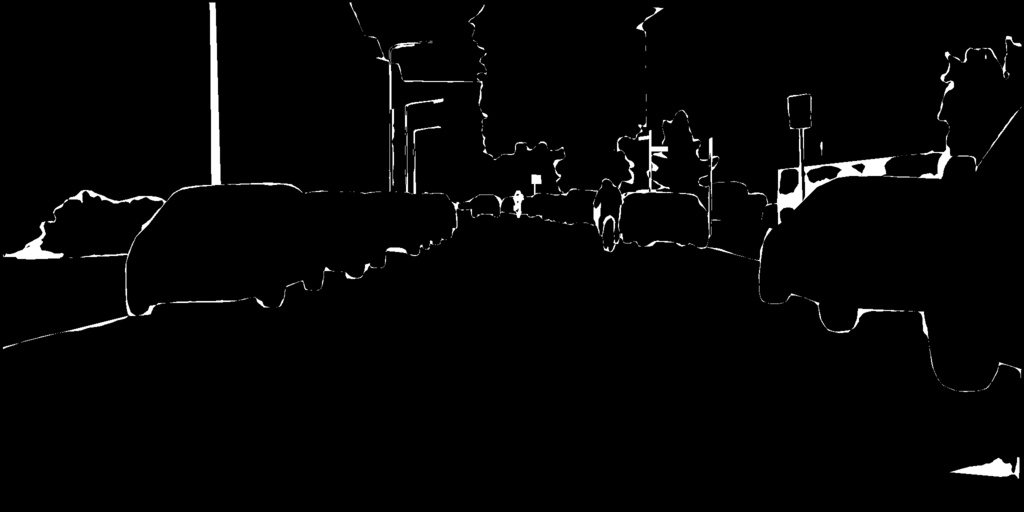}
        \caption{DRN - Misclassification}
        \label{fig:sub:drn_mis}
    \end{subfigure}
    \begin{subfigure}{0.247\textwidth}
        \centering
        \includegraphics[width=\linewidth]{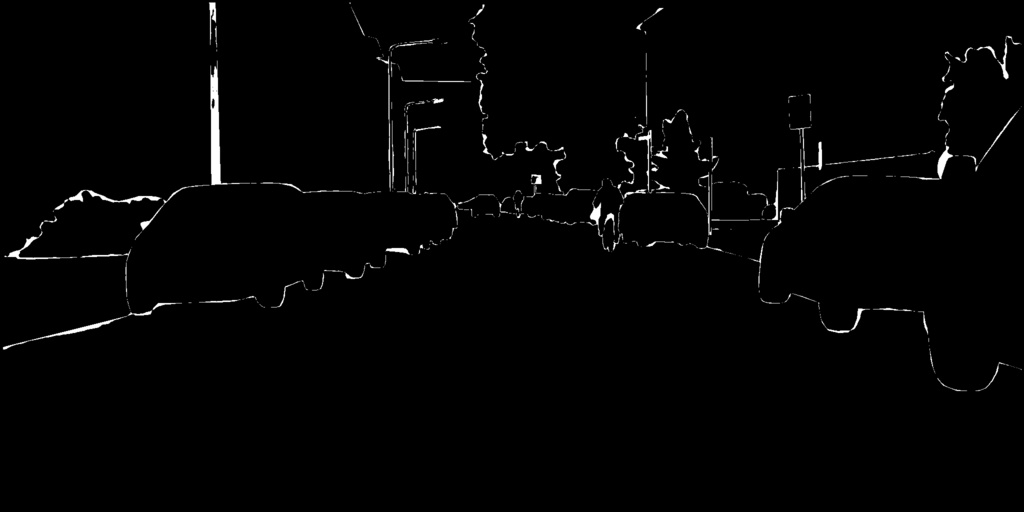}
        \caption{OneFormer - Misclassification}
        \label{fig:sub:of_mis}
    \end{subfigure}
    \begin{subfigure}{0.247\textwidth}
        \centering
        \includegraphics[width=\linewidth]{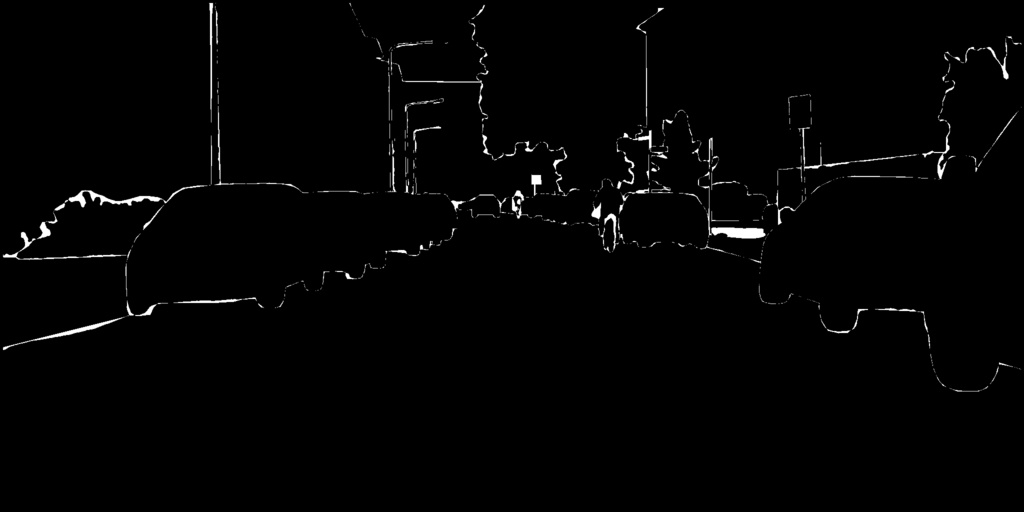}
        \caption{SegFormer - Misclassification}
        \label{fig:sub:sf_mis}
    \end{subfigure} \\
    \begin{subfigure}{0.247\textwidth}
        \centering
        \includegraphics[width=\linewidth]{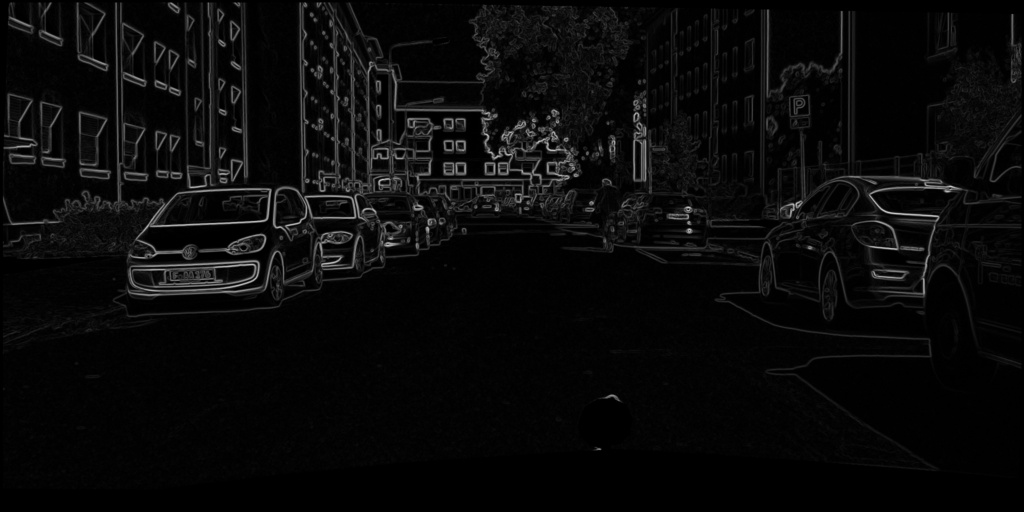}
        \caption{Scharr Edges}
        \label{fig:sub:edges}
    \end{subfigure}
    \begin{subfigure}{0.247\textwidth}
        \centering
        \includegraphics[width=\linewidth]{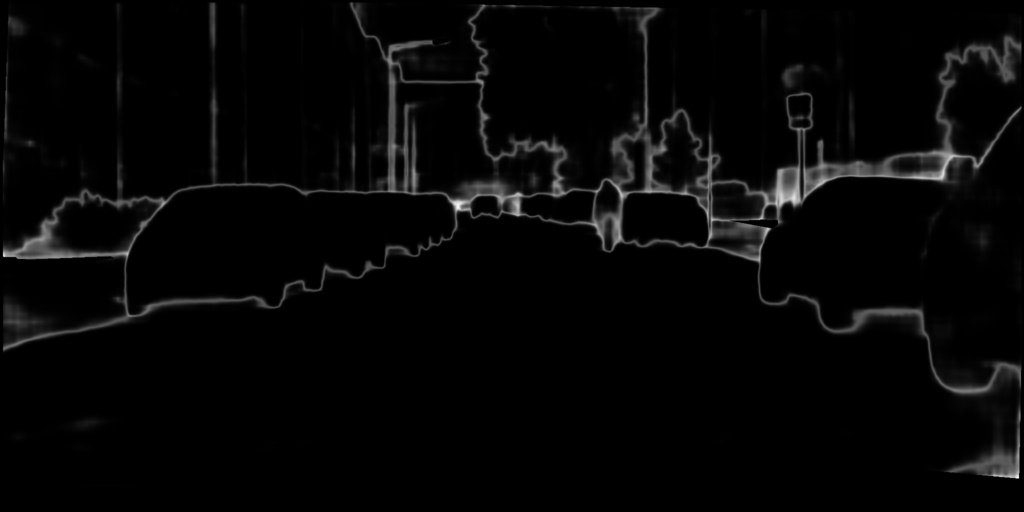}
        \caption{DRN - Entropy}
        \label{fig:sub:drn_ent}
    \end{subfigure}
    \begin{subfigure}{0.247\textwidth}
        \centering
        \includegraphics[width=\linewidth]{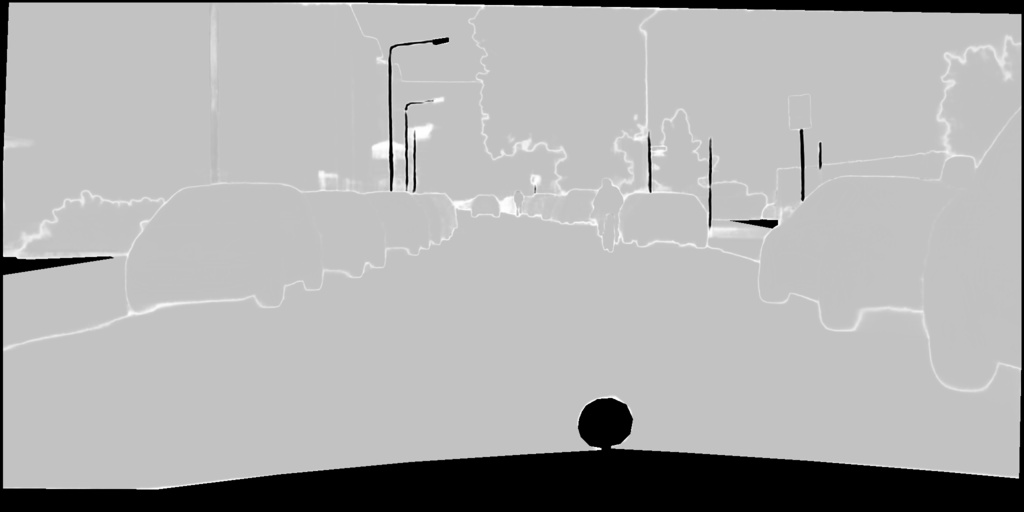}
        \caption{OneFormer - Entropy}
        \label{fig:sub:of_ent}
    \end{subfigure}
    \begin{subfigure}{0.247\textwidth}
        \centering
        \includegraphics[width=\linewidth]{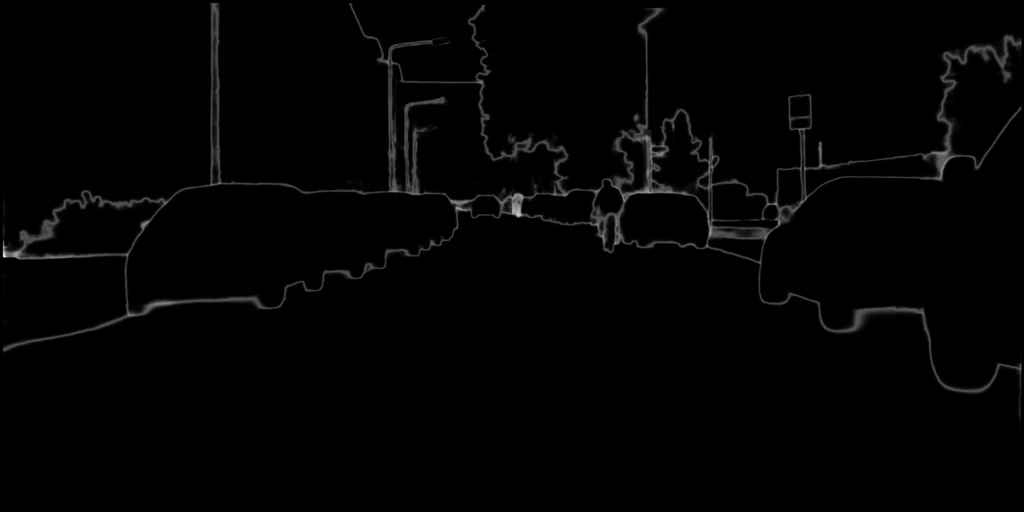}
        \caption{SegFormer - Entropy}
        \label{fig:sub:sf_ent}
    \end{subfigure} \\\\
    \begin{subfigure}{0.247\textwidth}
        \centering
        \includegraphics[width=\linewidth]{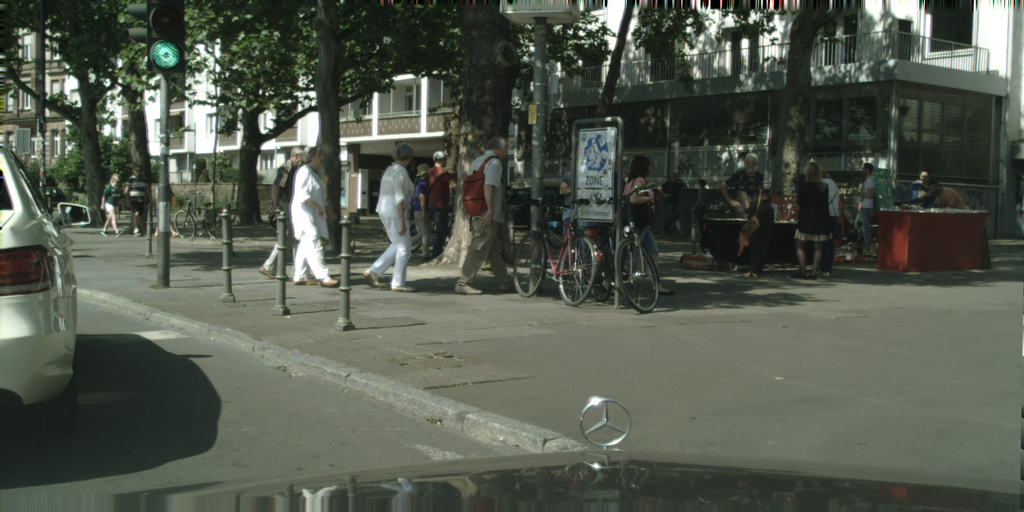}
        \caption{Image}
        \label{fig:sub:image2}
    \end{subfigure}
    \begin{subfigure}{0.247\textwidth}
        \centering
        \includegraphics[width=\linewidth]{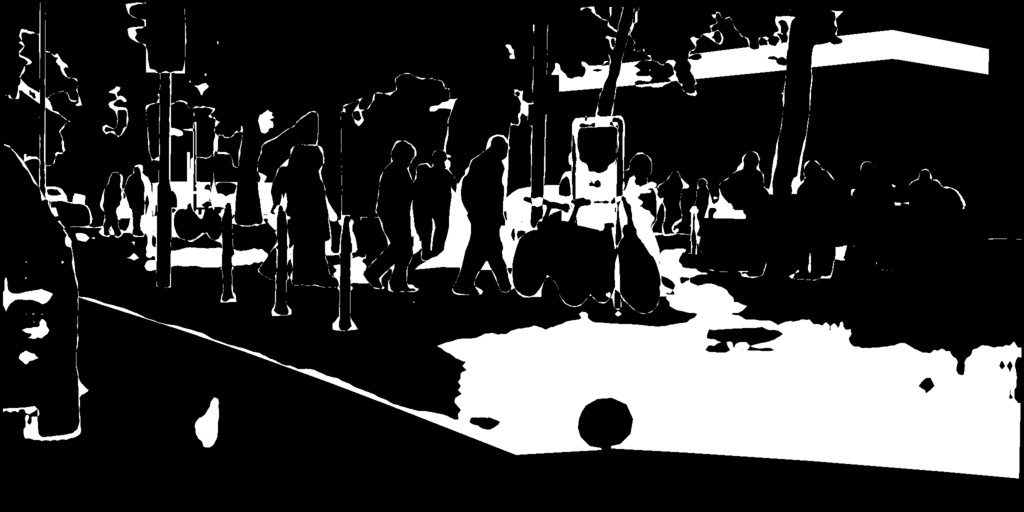}
        \caption{DRN - Misclassification}
        \label{fig:sub:drn_mis2}
    \end{subfigure}
    \begin{subfigure}{0.247\textwidth}
        \centering
        \includegraphics[width=\linewidth]{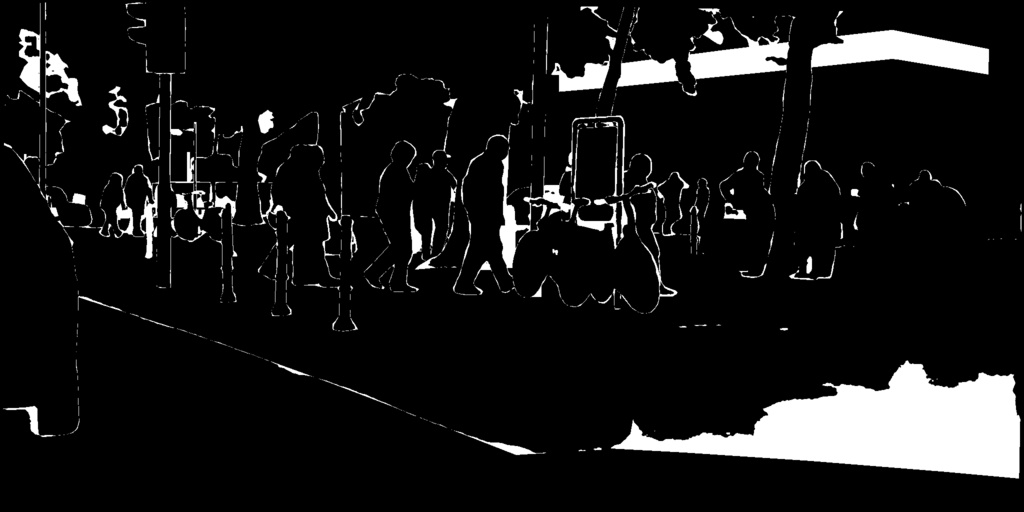}
        \caption{OneFormer - Misclassification}
        \label{fig:sub:of_mis2}
    \end{subfigure}
    \begin{subfigure}{0.247\textwidth}
        \centering
        \includegraphics[width=\linewidth]{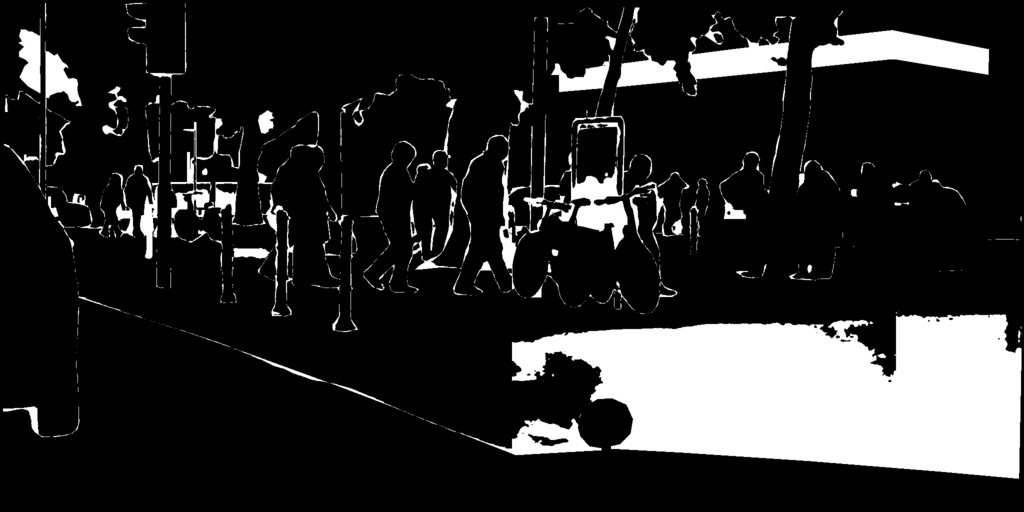}
        \caption{SegFormer - Misclassification}
        \label{fig:sub:sf_mis2}
    \end{subfigure} \\
    \begin{subfigure}{0.247\textwidth}
        \centering
        \includegraphics[width=\linewidth]{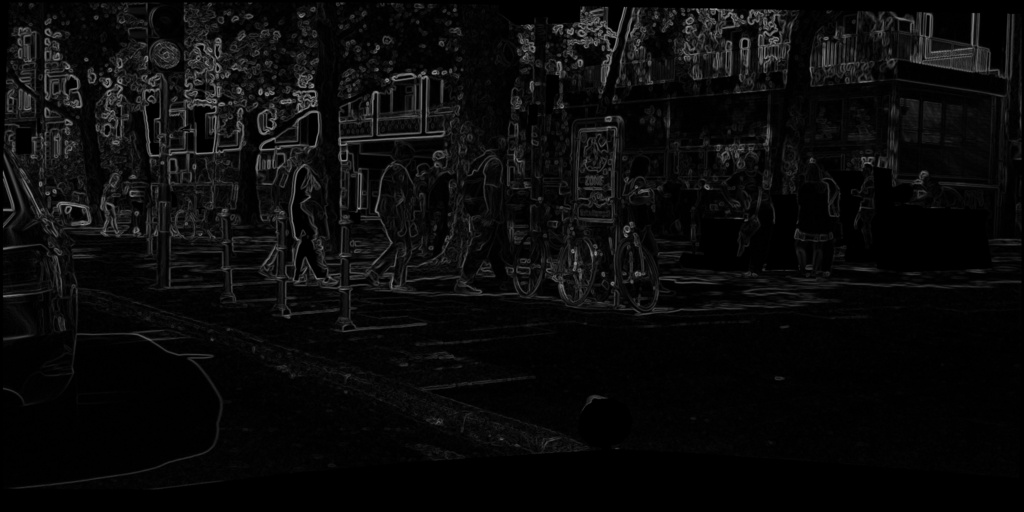}
        \caption{Scharr Edges}
        \label{fig:sub:edges2}
    \end{subfigure}
    \begin{subfigure}{0.247\textwidth}
        \centering
        \includegraphics[width=\linewidth]{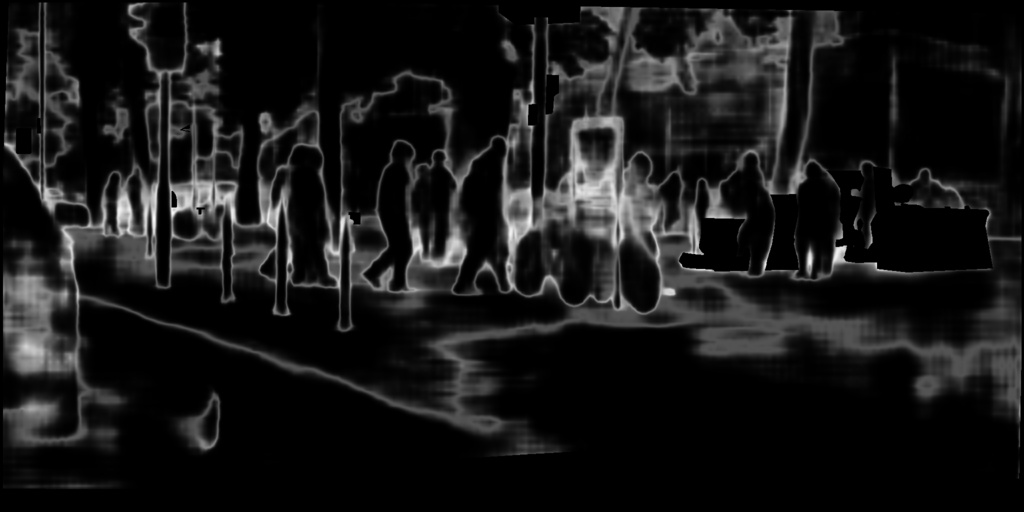}
        \caption{DRN - Entropy}
        \label{fig:sub:drn_ent2}
    \end{subfigure}
    \begin{subfigure}{0.247\textwidth}
        \centering
        \includegraphics[width=\linewidth]{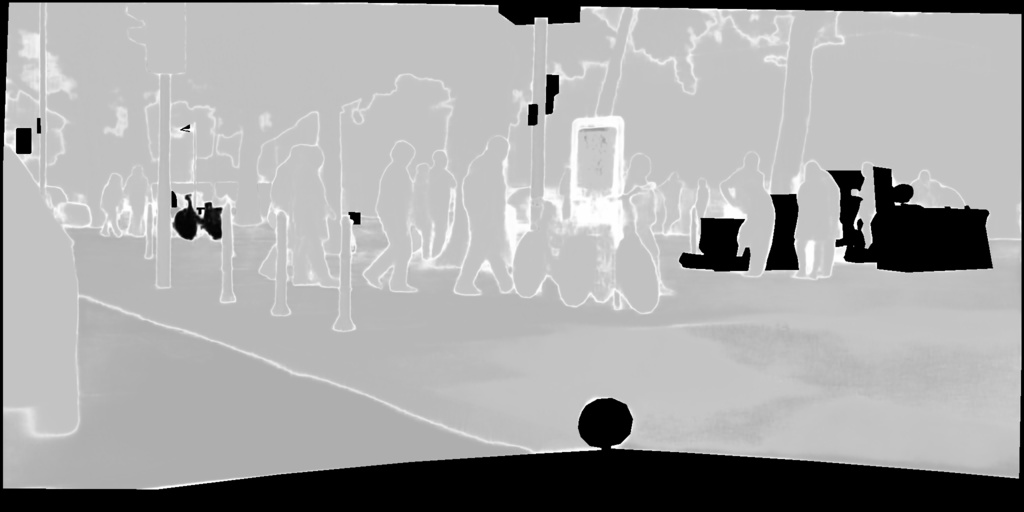}
        \caption{OneFormer - Entropy}
        \label{fig:sub:of_ent2}
    \end{subfigure}
    \begin{subfigure}{0.247\textwidth}
        \centering
        \includegraphics[width=\linewidth]{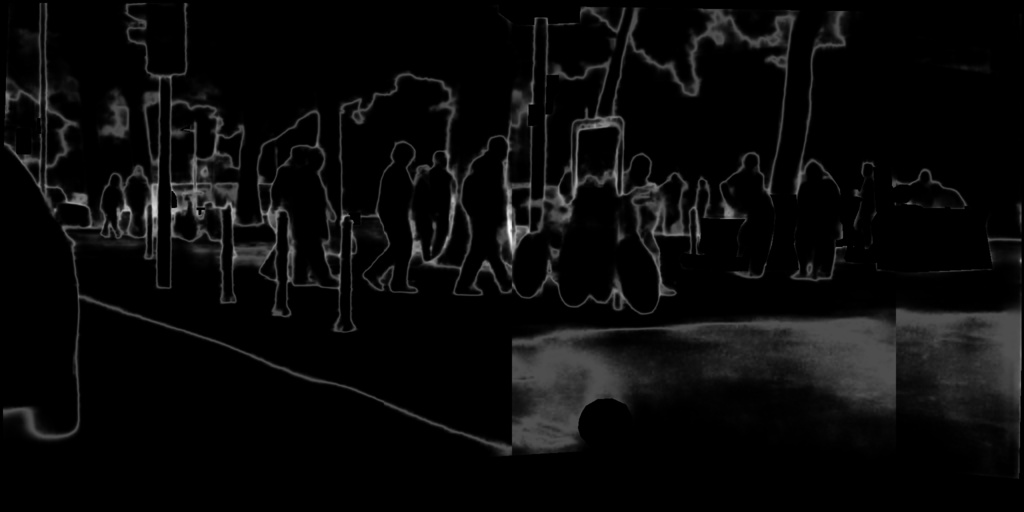}
        \caption{SegFormer - Entropy}
        \label{fig:sub:sf_ent2}
    \end{subfigure}
    \caption{Comparison of misclassified pixels and entropy of DRN, OneFormer and SegFormer networks on a couple of Cityscapes validation images, along with the images themselves and edges detected using the Scharr operator. It is to be noted that OneFormer generally produces high entropy outputs, which cause most of the image to be grey. However, we can still see that the highest entropy regions (in bright white) still correspond well to misclassified pixels.}
    \label{fig:misclassification}
\end{figure*}

\section{Failure Analysis}
Before we develop a trust score or metric, we must first look at and try to understand the failure cases of our models. We consider a few standard segmentation networks - Dilated Residual Networks~\cite{yu2017dilated} (DRN), OneFormer~\cite{jain2023oneformer} and SegFormer~\cite{xie2021segformer} and look at their outputs. We specifically choose a mix of older and newer networks for this since they are quite different architecturally and should allow us to focus on generalities rather than the peculiarities of any single architectural family. We use pre-trained weights on the Cityscapes dataset~\cite{cordts2016cityscapes} for all networks and observe their outputs on the validation set. ~\Cref{fig:sub:drn_mis,fig:sub:drn_mis2,fig:sub:of_mis,fig:sub:of_mis2,fig:sub:sf_mis,fig:sub:sf_mis2} highlight the misclassified pixels for the networks on a couple of such images. Black indicates correctly classified or ignored pixels, whereas white indicates misclassified pixels.

From the images, it is clear that OneFormer and SegFormer are better than DRN at segmenting images. Indeed, mIoU for DRN, OneFormer, and SegFormer are 0.5479, 0.6733, and 0.6875, respectively. However, we note that the failure modes for the networks are very similar. All of these networks misclassify pixels in similar regions of the images. It is also evident that several of these misclassifications lie along the edges of various objects in the images. This makes intuitive sense since it can be challenging, even for humans, to assign the boundary pixels to a particular object exactly.

This naturally brings us to the question, \textit{``Can we reasonably predict which pixels are most likely to be misclassified?''}. Even if we cannot accurately predict which pixels will be misclassified, we would like to have some measure of confidence for these classifications. Since we would like to apply whatever approach we come up with on existing networks without any re-training or finetuning, we need to consider what kind of information we can extract and use from just the images and their segmentation predictions since that is all we have during test time. Edge detection may seem an obvious first choice, as we have already established that several misclassified pixels lie along object boundaries. A better choice is some measure of uncertainty, such as entropy.

\Cref{fig:sub:edges,fig:sub:edges2} show the edges obtained when the Scharr filter is applied to the images. The gradient magnitudes thus obtained are scaled between \([0,1]\), with \(0\) being black and \(1\) being white. All ignored pixels are also in black. Since the edges are obtained from just the image rather than the predictions, the network used does not matter. This already tells us that edge detection may not be particularly effective in predicting misclassifications, which depend on the segmentation network used. While some edges align well with the misclassifications, most are correctly classified, especially the edges within each object. The fact remains that only the edges that align with ground truth label boundaries are misclassified. Moreover, by its very nature, edge detection completely fails to capture any misclassified pixels within objects; this is especially evident in the second image.

On the other hand, the simple entropy of predicted labels, as shown in \Cref{fig:sub:drn_ent,fig:sub:drn_ent2,fig:sub:of_ent,fig:sub:of_ent2,fig:sub:sf_ent,fig:sub:sf_ent2}, is far more aligned with the misclassifications. Similar to edges, the entropy is scaled between \([0,1]\) for each image and ignored pixels are set to \(0\). Here, we see a marked difference between OneFormer and the other two networks; DRN and SegFormer provide very confident predictions and result in most pixels having very low entropy, whereas OneFormer's classifications are under-confident, resulting in most pixels having higher entropy. However, even then, at a glance, we notice that the highest entropy pixels still align well with misclassifications.

\begin{figure*}[t]
    \centering
    \begin{subfigure}{0.425\linewidth}
        \centering
        \includegraphics[width=\linewidth]{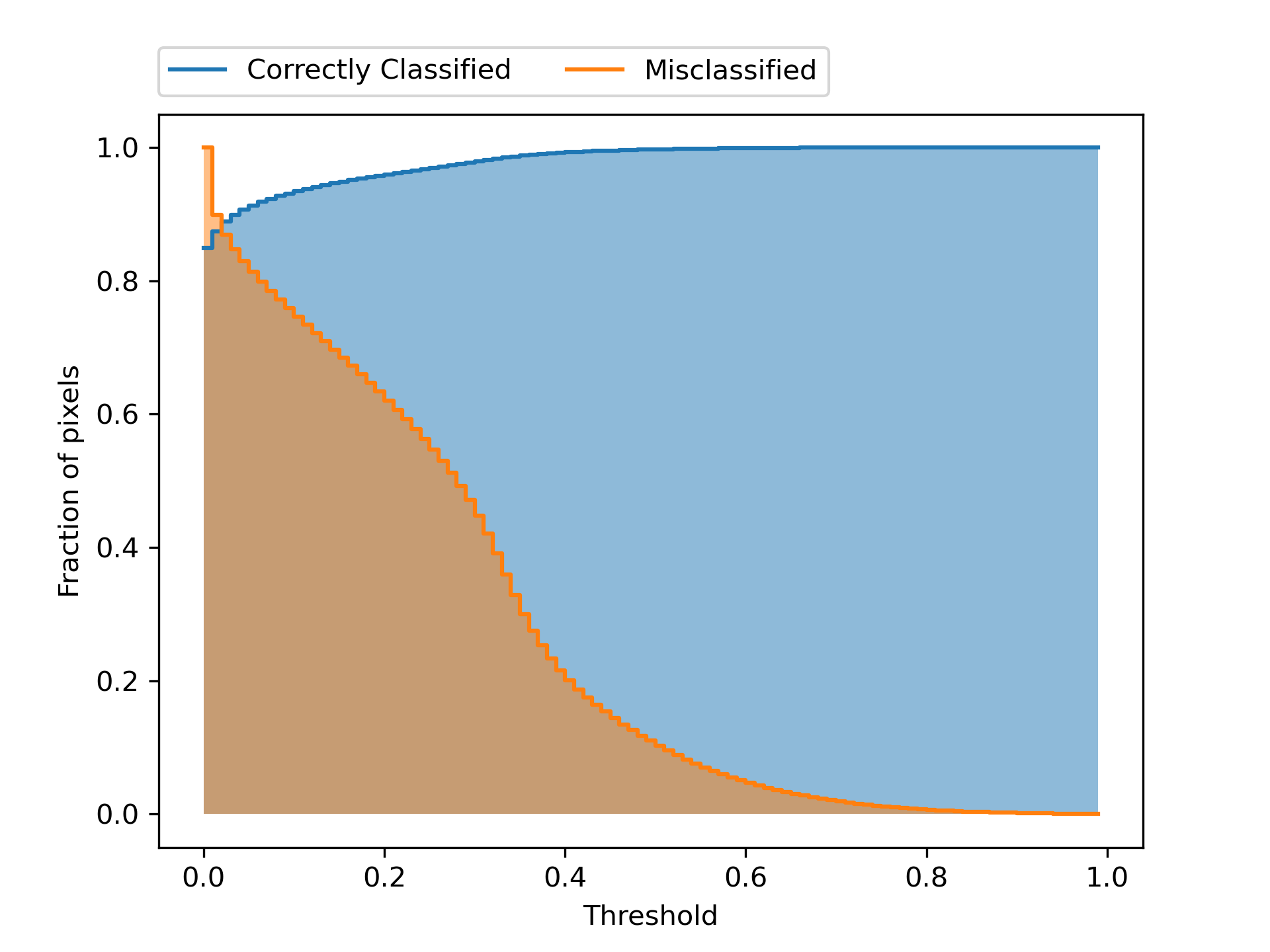}
        \caption{DRN D-105, Cityscapes}
    \end{subfigure}
    \begin{subfigure}{0.425\linewidth}
        \centering
        \includegraphics[width=\linewidth]{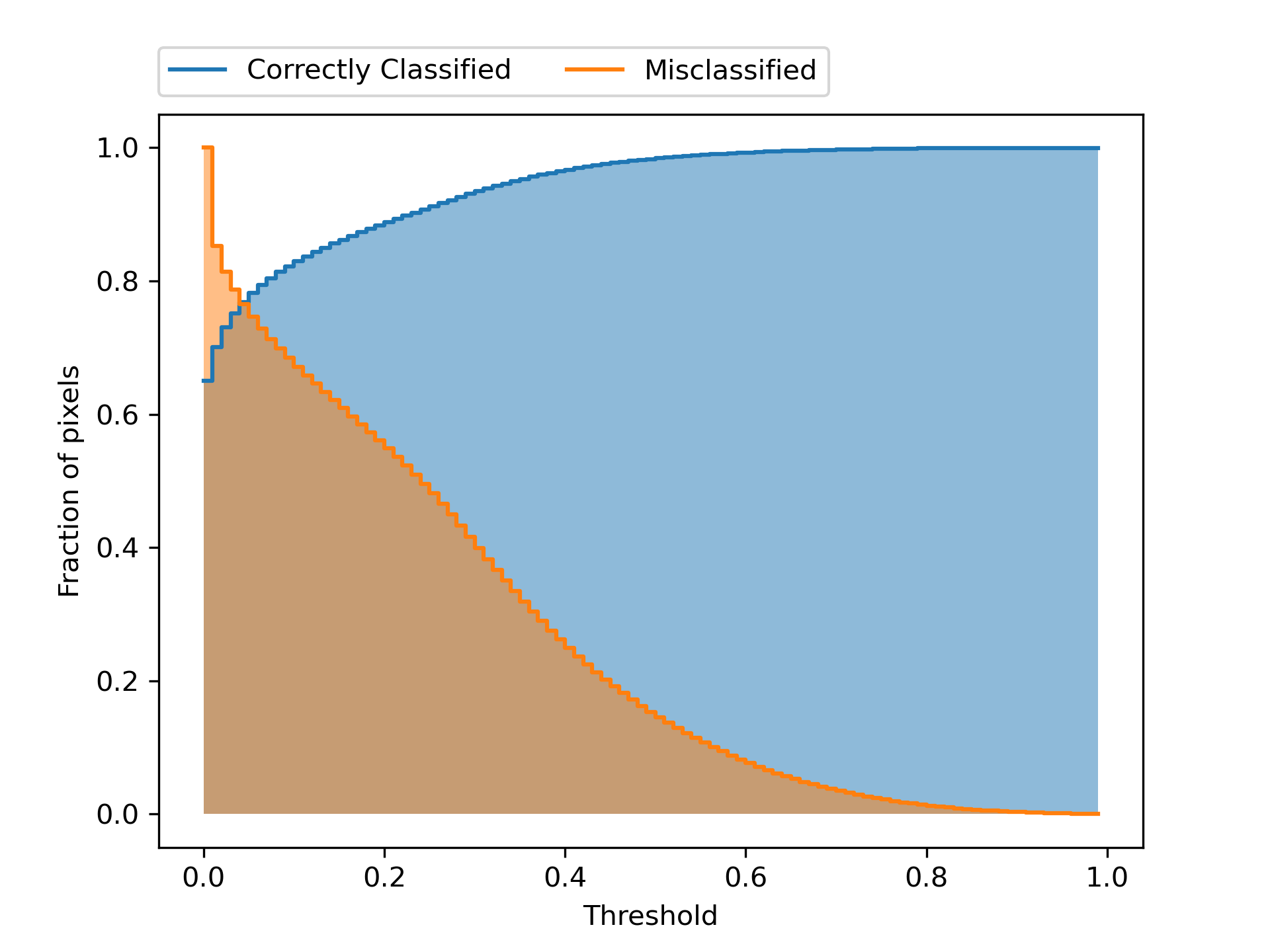}
        \caption{SegFormer B5, ADE20K}
    \end{subfigure}
    \caption{Cumulative histogram of entropy values for correctly classified (blue) and misclassified pixels (orange) for a couple of experimental settings.}
    \label{fig:histogram}
\end{figure*}

For a more quantitative look, we plot the cumulative histogram of entropy values for correctly classified and misclassified pixels in \Cref{fig:histogram}. The plot accumulates histogram counts for correctly classified pixels as the threshold increases and accumulates counts for misclassified pixels as it decreases. This allows us to directly read the percentage of correct and incorrect pixels predicted if we threshold entropy at any given value and consider anything below the threshold to be correctly classified and anything above it to be misclassified.

With these observations in mind, we focus on quantitative analysis of various uncertainty measures, which can be derived at test time, on pre-trained networks and consider how well they manage to predict misclassified pixels.

\section{Uncertainty Metrics}
We now describe the uncertainty metrics used for our experiments and how they are calculated. We only consider metrics that can be obtained purely at test time and those that work with nearly all neural networks. This rules out any approach for gauging uncertainty that requires us to change the network structure or fine-tune it. Specifically, we consider the following techniques.
\begin{itemize}
    \item \textbf{Probability Margins} - Some of the simplest possible uncertainty metrics we can obtain involve directly using the highest probability scores output by the network. We consider two different but related metrics. The first one is often called the variation ratio and is defined as follows:
    \begin{equation}
        VR = 1 - p_m
    \end{equation}
    where, \(p_m = \max_{i=1}^{K}(p_i)\) represents the highest probability across all classes.
    
    The second metric takes into account the difference between the highest and the second highest probability and is defined as:
    \begin{equation}
        PM = 1 - \left\{p_m - \max_{i=1, i\neq m}^{K}(p_i)\right\}
    \end{equation}
    \item \textbf{Entropy} - Another simple and possibly the most commonly used uncertainty metric we can obtain from a model. The only requirement is that the model outputs a probability distribution per pixel over all classes, which is true for nearly all (if not all) semantic segmentation models. Formally, the entropy for a single pixel is defined as:
    \begin{equation}
        H = -\sum_{k=1}^K p_k \log p_k
    \end{equation}
    where, \(K\) is the number of classes and \(p_k\) represents the probability of the pixel belonging to class \(k\).
    \item \textbf{Monte Carlo Dropout (MC Dropout)~\cite{gal2016dropout}} - Typically, Dropout layers~\cite{hinton2012improving} in neural networks are disabled during test time and replaced with a simple scaling of the features. MC Dropout utilises these layers to generate multiple predictions for the same image by keeping the Dropout layers active during test time. This introduces a source of randomness in the network, which allows it to produce different outputs for the same input on each forward pass. The variations in these predictions can be used as a form of uncertainty. However, one drawback of the technique is that it can only be used with networks trained with Dropout layers.
    \item \textbf{Noise} - Yet another approach to obtain uncertainty is to introduce a tiny amount of random noise in the input, which perturbs the output. Like MC Dropout, this allows us to generate multiple predictions for the same input by using multiple forward passes (each with a different noise). For our purposes, we use zero mean Gaussian noise, with a small standard deviation.
    \item \textbf{Scaling} - Segmentation networks already use multi-scale inputs to improve their performance~\cite{jain2023oneformer,yu2017dilated}. Essentially, it involves running forward passes on multiple scales of the input, which generate multiple segmentations for the same image. The networks can then scale back the outputs to the original resolution and average them to get better results than a single pass. We can, however, also use these predictions as a source of uncertainty.
\end{itemize}

Except for Probability Margins and Entropy, all the other methods give us multiple predictions for each input image. We can extract uncertainty information from these predictions in a few ways, described as follows:
\begin{itemize}
    \item \textbf{Averaged Probability Margins} - Given multiple probability distributions over a pixel, we can take their mean to obtain a single representative probability distribution for said pixel. Mathematically we have:
    \begin{equation}
        \overline{p_k} = \frac{1}{N}\sum_{i=1}^N p_k^i
        \label{eqn:avg_p}
    \end{equation}
    where \(\overline{p_k}\) represents the average probability of a pixel belonging to class \(k\), and \(p_k^i\) represents the probability of the pixel belonging to class \(k\) for the \(i^\textrm{th}\) prediction. \(N\) is the total number of predictions. Once we have the average probability distribution per pixel, we can compute probability margin values over them as follows:
    \begin{equation}
        VR = 1 - \overline{p_m},\quad\mathrm{where}\quad\overline{p_m} = \max_{i=1}^{K}(\overline{p_i})
    \end{equation}
    \begin{equation}
        PM = 1 - \left\{\overline{p_m} - \max_{i=1, i\neq m}^{K}(\overline{p_i})\right\}
    \end{equation}
    \item \textbf{Averaged Entropy} - Similar to averaged probability margins, we can use \cref{eqn:avg_p} to obtain the average probability distribution per pixel and compute entropy over it as:
    \begin{equation}
        H = -\sum_{k=1}^K \overline{p_k} \log \overline{p_k}
    \end{equation}
    \item \textbf{Variance} - Variance of the probability distribution has also been used as a measure of uncertainty in Bayesian settings~\cite{gal2017deep,smith2018understanding}. Since we have multiple predictions, we can compute the empirical variance of class \(k\) for any given pixel as:
    \begin{equation}
        \sigma^2_k = \frac{1}{N}\sum_{i=1}^N (p_k^i - \overline{p_k})^2
    \end{equation}
    where, \(\overline{p_k}\) is calculated as in \cref{eqn:avg_p}. To obtain a single value per pixel, we consider taking both the average as well as the maximum across across all classes.
    \item \textbf{Bayesian Active Learning by Disagreement (BALD)}~\cite{houlsby2011bayesian} - Originally proposed as an acquisition function for selecting samples in a Bayesian active learning setting, BALD can also be used as an uncertainty metric. The BALD score is defined as:
    \begin{equation}
        S_{\textrm{BALD}} = -\sum_{k=1}^K\overline{p_k}\log \overline{p_k} + \frac{1}{N}\sum_{k=1,i=1}^{K,N}p_k^i\log p_k^i
    \end{equation}
    where, \(\overline{p_k}\) and \(p_k^i\) are same as defined earlier.
    Essentially, it checks the disagreement between the entropy of the expected prediction and the expected entropy across all predictions. Higher disagreement represents more uncertainty.
\end{itemize}

\section{Experiments and Observations}
\label{sec:exp}

We now look at various experiments to ascertain the usefulness of the uncertainty metrics described in the last section in determining whether a pixel is likely to be misclassified. First, we describe the datasets used for our experiments.

\textbf{Cityscapes~\cite{cordts2016cityscapes}} - The primary dataset used for our experiments consists of dashcam videos from cars in several German cities targeted towards understanding urban street scenes. The validation set is comprised of 500 finely annotated images of \(2048\times1024\) resolution. Pixels in each image are annotated as belonging to one of 30 defined classes, of which 19 are considered for classification and evaluation. Any pixels not belonging to one of these classes are effectively ignored for evaluation purposes.

\textbf{Dark Zurich~\cite{sakaridis2019guided}} - A dataset analogous to Cityscapes with a small validation set of only 50 \(1920\times1080\) images; it consists of urban scenes from the city of Zurich at night time as opposed to Cityscapes where the images are taken at day time. We use this dataset to check how well our metrics perform when there is a domain shift of the input.

\textbf{ADE20K~\cite{zhou2019semantic}} - A rather large dataset with a validation set of 2000 images of varying resolutions. Unlike Cityscapes and Dark Zurich, ADE20K has a much larger number of segmentation classes, 150 and the images are not only from dashcam videos but cover a wide range of subjects.

We now describe the segmentation networks that were used in our experiments. We choose our models such that they range from small (DRN D-22) to large (OneFormer Swin-L) and cover a wide variety of architectures.

\textbf{Dilated Residual Networks~\cite{yu2017dilated}} - A Convolutional Neural Network (CNN) based on ResNet~\cite{he2016deep} but using Dilated Convolutions instead, which helps it achieve better performance in semantic segmentation tasks. We use the pre-trained weights for DRN D-22 and DRN D-105 variants provided by the authors for the Cityscapes dataset.

\textbf{OneFormer~\cite{jain2023oneformer}} - A Transformer-based architecture that utilises features from a backbone network and passes them through a Transformer decoder to get pixel-level classifications. We use pre-trained weights with ConvNeXt-L and Swin-L backbones on Cityscapes and ADE20K datasets, as provided by the authors.

\textbf{SegFormer~\cite{xie2021segformer}} - Another Transformer-based architecture that uses a Hierarchical Transformer encoder and a Multi-Layer Perceptron (MLP) decoder. We use pre-trained weights of the B5 variant for Cityscapes and ADE20K datasets as provided by the authors.

In order to measure how well a particular uncertainty metric corresponds to misclassifications, we use precision, recall and area under the receiver operating characteristic (AUROC) metrics. AUROC involves plotting the False Positive Rate against the True Positive Rate at different thresholds and then computing the area covered by the curve. While AUROC takes into account multiple threshold values, for computing precision and recall, we need to choose a threshold value to binarise the uncertainty scores. Unfortunately, as seen in \Cref{fig:misclassification}, the scales of these uncertainty values can vary quite a lot across different models and even across different images for a single model. As such, we cannot use a single constant threshold value across all models or images. Therefore, we define a couple of processes to dynamically choose a threshold for each image. Both methods rely on the fact that the fraction of misclassified pixels is not very high.
\begin{enumerate}
    \item For the first method, we check the ratio of predicted misclassified pixels to total pixels at multiple uniformly spaced threshold levels and select the threshold that causes the largest change in this value.
    \item We note that the ratio of predicted misclassified pixels to total pixels increases as the threshold increases. For the second method, we specify a maximum value allowed for this ratio and choose the highest threshold, which results in a ratio lower than the specified value.
\end{enumerate}

Before we discuss the results, we clarify the meaning of each scenario:
\begin{itemize}
    \item \textbf{Base} - The simplest scenario where we only consider a single forward pass. As such, only Probability Margins and Entropy can be used as uncertainty metrics for this setting.
    \item \textbf{Noise} - The images are injected with a Gaussian noise of zero mean and \(0.01\) standard deviation. The noise is injected after normalizing the images. We use ten forward passes per image in this setting.
    \item \textbf{Scale} - This represents using multi-scaled inputs for generating multiple outputs. We use scales of \(\{0.5, 0.75, 1.0, 1.25, 1.5\}\) for our purposes.
    \item \textbf{Drop} - MC Dropout is applied in this scenario. Since it requires the network to have Dropout layers, we did not use this technique with DRN. Similar to the Noise scenario, we use ten passes per image.
\end{itemize}

Before moving ahead, we would like to state that although a large number of experiments under various scenarios were performed, we only highlight some of those results for brevity and compactness in the following text. The complete set of results for all the experiments is provided in the supplementary materials for this paper. Similarly, since we obtain the metrics for each pixel rather than each image, we consider both micro-averaging and macro-averaging over the images. However, we only show the micro-averaged results in this text and provide the rest in the supplementary material. The supplementary material also includes classwise results for several experiments.

\begin{table}[t]
    \centering
    \caption{Area under the Receiver Operating Characteristics (AUROC) on the Cityscapes dataset. VR, PM, EN, AV, MV and BALD represent Variation Ratio, Probability Margin, Entropy, Average Variance, Maximum Variance and BALD metrics respectively.}
    \begin{tabular}{lcccccc}
    \toprule
    Model & VR & PM & EN & AV & MV & BALD \\
    \midrule
    \multicolumn{7}{c}{DRN D-22} \\
    Base & 0.928 & 0.930 & 0.934 \\
    Noise & 0.928 & 0.930 & 0.934 & 0.772 & 0.772 & 0.828 \\
    Scale & \textbf{0.947} & \textbf{0.947} & 0.944 & 0.900 & 0.900 & 0.908 \\
    \midrule
    \multicolumn{7}{c}{OneFormer ConvNeXt-L} \\
    Base & 0.882 & 0.907 & 0.857 \\
    Noise & 0.885 & 0.909 & 0.861 & 0.647 & 0.632 & 0.674 \\
    Scale & 0.897 & \textbf{0.922} & 0.874 & 0.788 & 0.768 & 0.821 \\
    Drop & 0.894 & 0.917 & 0.871 & 0.726 & 0.702 & 0.674 \\
    \midrule
    \multicolumn{7}{c}{SegFormer B5} \\
    Base & 0.909 & 0.914 & 0.923 \\
    Noise & 0.909 & 0.914 & 0.924 & 0.715 & 0.714 & 0.797 \\
    Scale & 0.939 & 0.942 & \textbf{0.946} & 0.883 & 0.883 & 0.909 \\
    Drop & 0.910 & 0.915 & 0.924 & 0.793 & 0.792 & 0.852 \\
    \bottomrule
    \end{tabular}
    \label{tab:auroc_cityscapes}
\end{table}

\begin{table}[t]
    \centering
    \caption{Area under the Receiver Operating Characteristics (AUROC) on the ADE20K dataset. VR, PM, EN, AV, MV and BALD represent Variation Ratio, Probability Margin, Entropy, Average Variance, Maximum Variance and BALD metrics respectively.}
    \begin{tabular}{lcccccc}
    \toprule
    Model & VR & PM & EN & AV & MV & BALD \\
    \midrule
    \multicolumn{7}{c}{OneFormer ConvNeXt-L} \\
    Base & 0.607 & 0.705 & 0.616 \\
    Noise & 0.605 & 0.704 & 0.615 & 0.624 & 0.618 & 0.667 \\
    Scale & 0.624 & \textbf{0.729} & 0.630 & 0.673 & 0.664 & 0.728 \\
    Drop & 0.614 & 0.715 & 0.617 & 0.664 & 0.652 & 0.667 \\
    \midrule
    \multicolumn{7}{c}{SegFormer B5} \\
    Base & 0.800 & 0.802 & 0.848 \\
    Noise & 0.800 & 0.803 & 0.848 & 0.749 & 0.712 & 0.795 \\
    Scale & 0.836 & 0.835 & \textbf{0.866} & 0.810 & 0.793 & 0.827 \\
    Drop & 0.802 & 0.804 & 0.849 & 0.796 & 0.752 & 0.830 \\
    \bottomrule
    \end{tabular}
    \label{tab:auroc_ade20k}
\end{table}

\begin{table*}[tb]
    \centering
    \caption{Classwise AUROC scores on Cityscapes and AUROC datasets. DRN refers to DRN D-105, OF refers to OneFormer ConvNeXt-L and SF refers to SegFormer B5 models. All scores use entropy as the uncertainty metric.}
    \begin{tabular}{lllllllllll}
        \toprule
        Model & \multicolumn{5}{c}{Top 5} & \multicolumn{5}{c}{Bottom 5} \\
        \midrule
        \multicolumn{11}{c}{Cityscapes} \\
        DRN & sky & car & vege. & road & building & pole & rider & terrain & fence & wall \\
        Base & 0.9681 & 0.9672 & 0.9362 & 0.9241 & 0.9235 & 0.7529 & 0.7214 & 0.7133 & 0.6946 & 0.6513 \\
        DRN & car & sky & road & vege. & bus & terrain & rider & train & fence & wall \\
        Scale & 0.9727 & 0.9708 & 0.9615 & 0.9534 & 0.9349 & 0.7487 & 0.7483 & 0.7414 & 0.7400 & 0.6980 \\
        SegFormer & sky & car & vege. & bus & building & rider & pole & fence & wall & terrain \\
        Base & 0.9660 & 0.9464 & 0.9286 & 0.9203 & 0.9148 & 0.8223 & 0.7565 & 0.7387 & 0.7152 & 0.7117 \\
        SegFormer & car & bus & sky & vege. & truck & rider & fence & terrain & pole & wall \\
        Scale & 0.9654 & 0.9631 & 0.9628 & 0.9511 & 0.9379 & 0.8396 & 0.8026 & 0.7751 & 0.7680 & 0.7167 \\
        \midrule
        \multicolumn{11}{c}{ADE20K} \\
        OneFormer & washer & pier & boat & gr.stand & barrel & shower & column & runway & fountain & lake \\
        Base & 0.9583 & 0.8935 & 0.8885 & 0.8782 & 0.8543 & 0.3092 & 0.3089 & 0.2676 & 0.2365 & 0.0234 \\
        OneFormer & washer & boat & dishw. & sea & hill & scr. door & buffet & runway & clock & lake \\
        Scale & 0.9575 & 0.9060 & 0.8802 & 0.8649 & 0.8445 & 0.3557 & 0.3454 & 0.3359 & 0.3102 & 0.0860 \\
        SegFormer & tent & bus & ship & runway & bed & crt & dirt track & monitor & shower & land \\
        Base & 0.9881 & 0.9765 & 0.9629 & 0.9490 & 0.9456 & 0.3179 & 0.2451 & 0.1745 & 0.1334 & 0.1227 \\
        SegFormer & tent & bus & bed & runway & pool table & crt & dirt track & land & shower & monitor \\
        Scale & 0.9890 & 0.9795 & 0.9646 & 0.9581 & 0.9528 & 0.2905 & 0.2267 & 0.1449 & 0.1273 & 0.1027 \\
        \bottomrule
    \end{tabular}
    \label{tab:classwise_auroc}
\end{table*}

\begin{table}[!h]
    \centering
    \caption{Precision and Recall on Cityscapes dataset using largest difference as thresholding strategy. All of the results shown use Entropy as the uncertainty metric.}
    \begin{tabular}{lcccc}
        \toprule
        Metric & Base & Noise & Scale & Drop \\
        \midrule
        \multicolumn{5}{c}{DRN D-22} \\
        Precision & \textbf{16.74} & \textbf{16.74} & 12.52 & \\
        Recall & 96.89 & 96.90 & \textbf{99.19} & \\
        Pixel \% & 29.21 & 29.22 & 40.72 & \\
        \midrule
        \multicolumn{5}{c}{OneFormer ConvNeXT-L} \\
        Precision & 16.23 & \textbf{16.54} & 14.36 & 15.12 \\
        Recall & 84.88 & 85.05 & \textbf{87.42} & 87.21 \\
        Pixel \% & 18.65 & 18.21 & 21.90 & 19.40 \\
        \midrule
        \multicolumn{5}{c}{SegFormer B5} \\
        Precision & \textbf{20.89} & 20.84 & 16.85 & 20.74 \\
        Recall & 90.47 & 90.55 & \textbf{95.29} & 90.66 \\
        Pixel \% & 13.20 & 13.26 & 17.48 & 13.33 \\
        \bottomrule
    \end{tabular}
    \label{tab:largest_diff_cityscapes}
\end{table}

As AUROC allows us to get a good overview of how our uncertainty metrics correlate with misclassification, we first look at these in \Cref{tab:auroc_cityscapes} for the Cityscapes dataset. We observe that simple metrics like Entropy and Probability Margins perform better than more involved metrics such as Variance and BALD. We also look at the values of AUROC on ADE20K dataset in \Cref{tab:auroc_ade20k}. The differences are smaller between various metrics on this dataset. Due to the large number of classes in this dataset, the overall entropy of each pixel is higher, making it challenging to separate correctly classified pixels from misclassified ones.

Let us also look at classwise AUROC results in \Cref{tab:classwise_auroc}. Due to space constraints, we only show the five best-performing and worst-performing classes here and provide results for all classes in the supplementary material. We observe that for Cityscapes, classes that perform best are also classes which are more common and occupy larger areas in images. In contrast, poorly performing classes are relatively less common or have smaller areas. For ADE20K, the results are more challenging to explain as there are 150 classes to consider. However, in general, they seem to correspond well to the overall performance of the models, such that the classes with better AUROC also have better IoU and vice versa.

\begin{table}[tb]
    \centering
    \caption{Precision, Recall, and AUROC on the Dark Zurich dataset, with Entropy as the uncertainty metric.}
    \begin{tabular}{lcccc}
        \toprule
        Scenario & \multicolumn{3}{c}{Largest Difference} & AUROC \\
        & Pr & Rc & Px\% & \\
        \midrule
        \multicolumn{5}{c}{DRN D 22} \\
        Base & 69.46 & 81.24 & 77.15 & 0.6062 \\
        Noise & 69.35 & \textbf{81.28} & 77.32 & 0.6038 \\
        Scale & \textbf{72.34} & 79.24 & 72.67 & \textbf{0.6714} \\
        \midrule
        \multicolumn{5}{c}{OneFormer ConNeXt-L} \\
        Base &  39.55 & 60.16 & 42.79 & 0.5949 \\
        Noise & 42.15 & 58.65 & 39.84 & 0.6036 \\
        Scale & \textbf{43.69} & \textbf{77.11} & 49.79 & \textbf{0.7012} \\
        Drop & 40.23 & 61.24 & 42.61 & 0.6089 \\
        \bottomrule
    \end{tabular}
    \label{tab:pr_rc_dz}
\end{table}

\begin{figure*}[tb]
    \begin{subfigure}{0.196\textwidth}
        \centering
        \includegraphics[width=\linewidth]{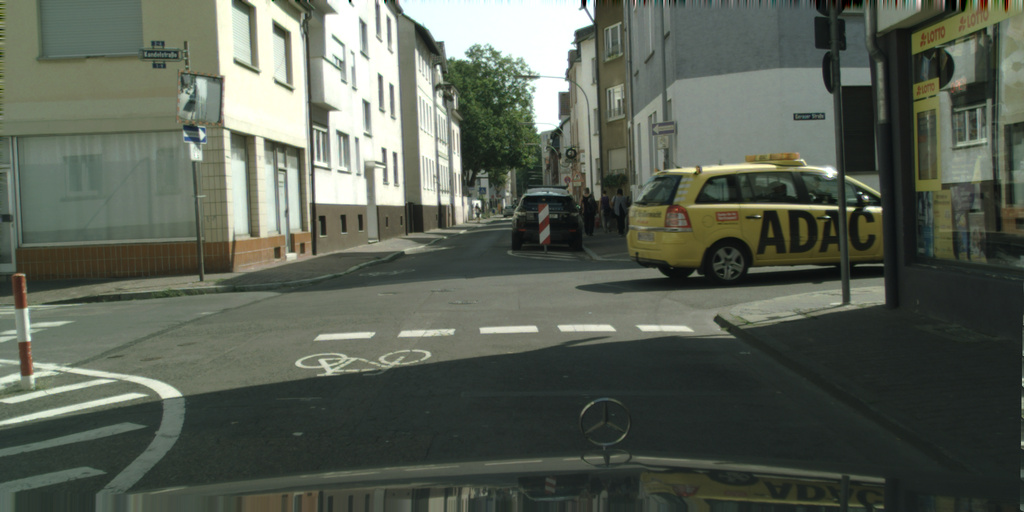}
    \end{subfigure}
    \begin{subfigure}{0.196\textwidth}
        \centering
        \includegraphics[width=\linewidth]{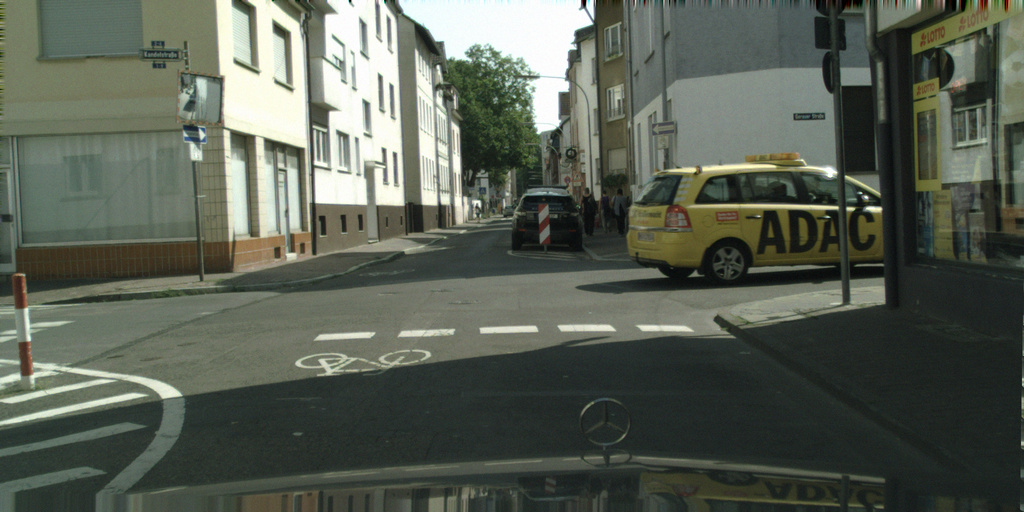}
    \end{subfigure}
    \begin{subfigure}{0.196\textwidth}
        \centering
        \includegraphics[width=\linewidth]{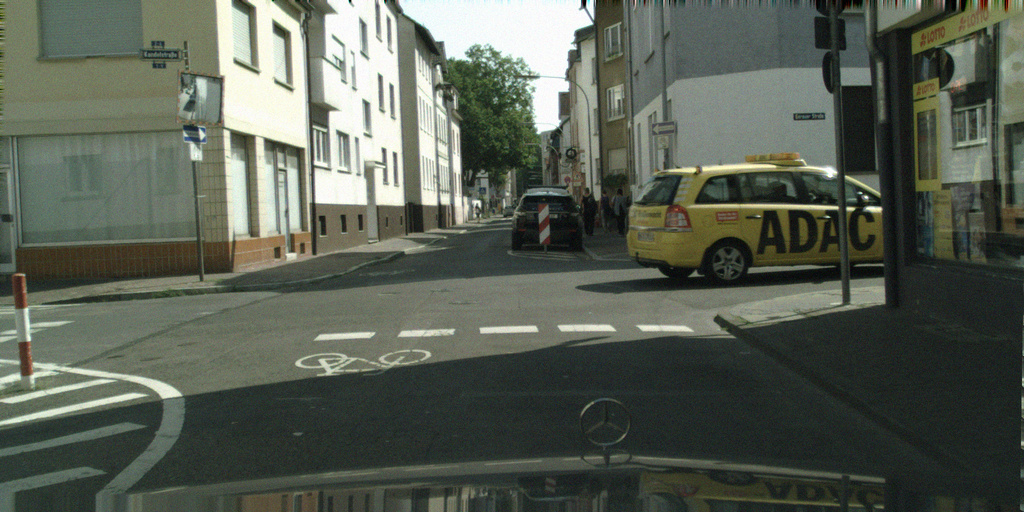}
    \end{subfigure}
    \begin{subfigure}{0.196\textwidth}
        \centering
        \includegraphics[width=\linewidth]{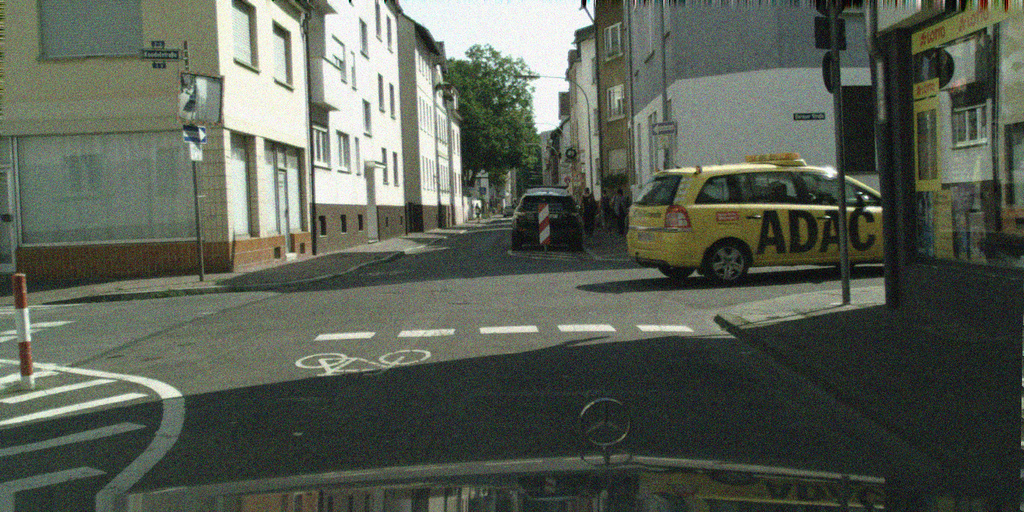}
    \end{subfigure}
    \begin{subfigure}{0.196\textwidth}
        \centering
        \includegraphics[width=\linewidth]{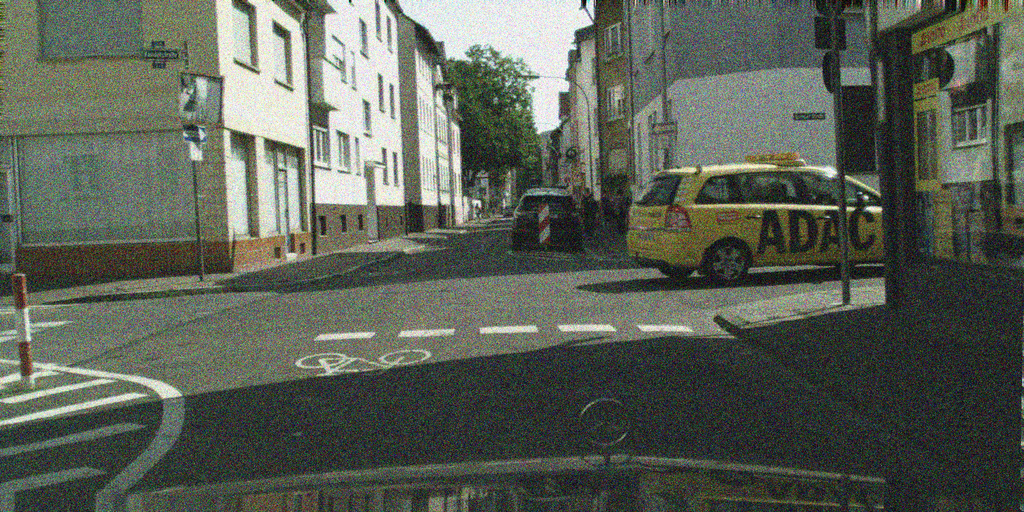}
    \end{subfigure} \\
    \begin{subfigure}{0.196\textwidth}
        \centering
        \includegraphics[width=\linewidth]{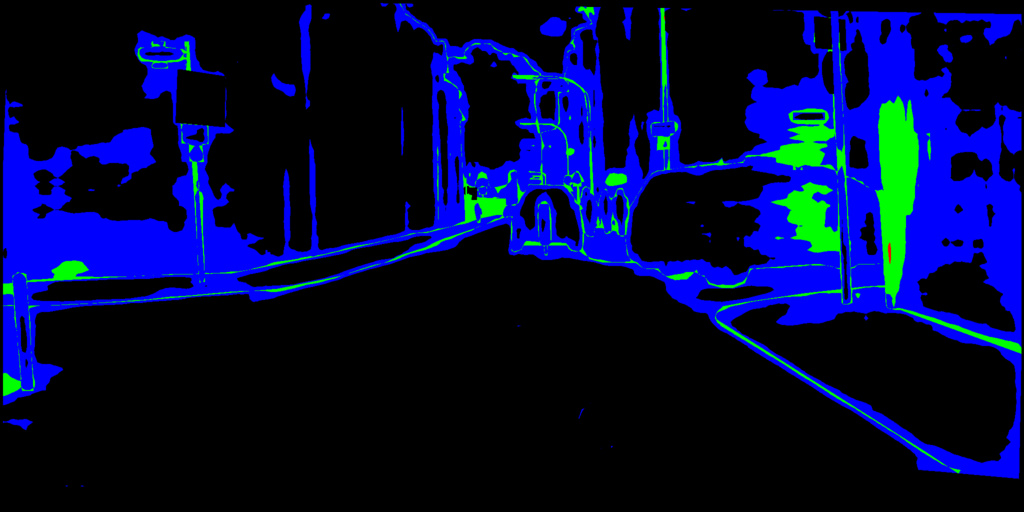}
    \end{subfigure}
    \begin{subfigure}{0.196\textwidth}
        \centering
        \includegraphics[width=\linewidth]{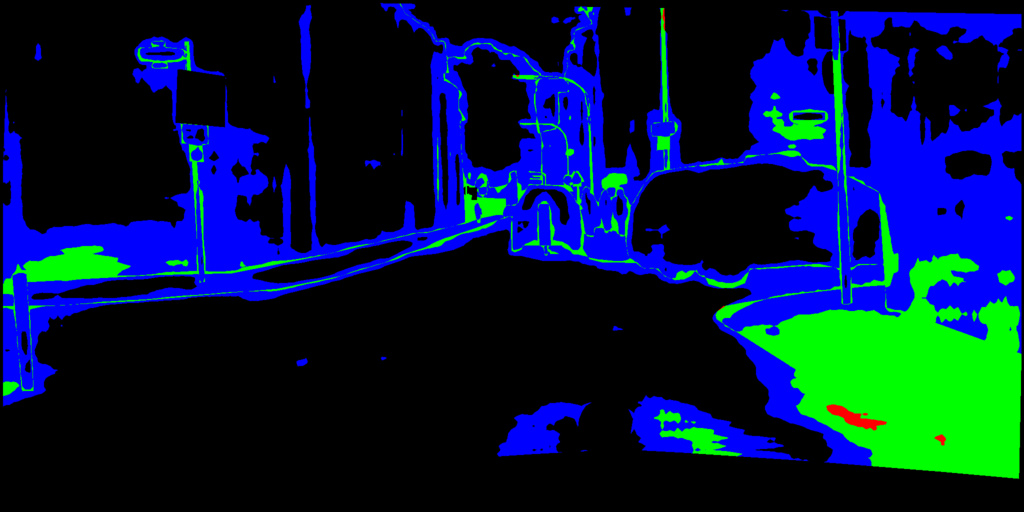}
    \end{subfigure}
    \begin{subfigure}{0.196\textwidth}
        \centering
        \includegraphics[width=\linewidth]{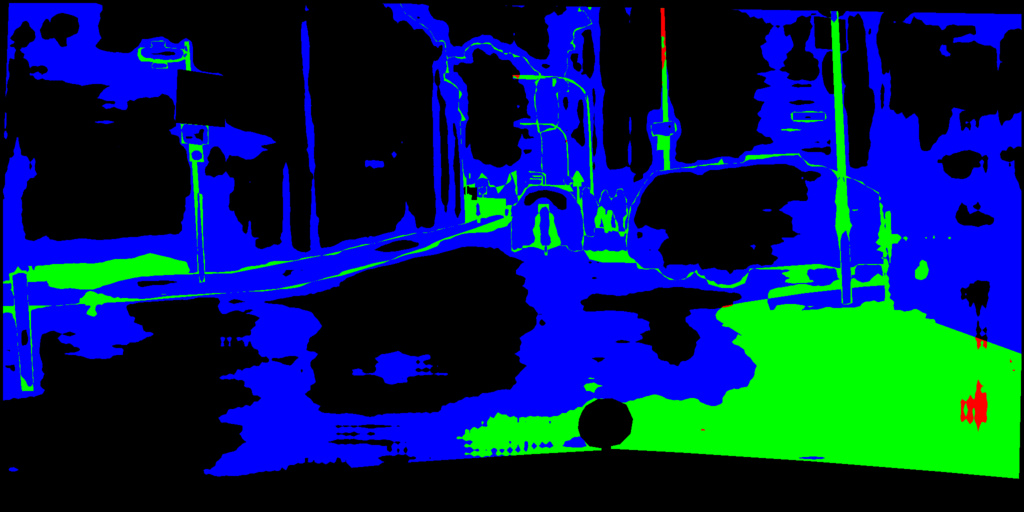}
    \end{subfigure}
    \begin{subfigure}{0.196\textwidth}
        \centering
        \includegraphics[width=\linewidth]{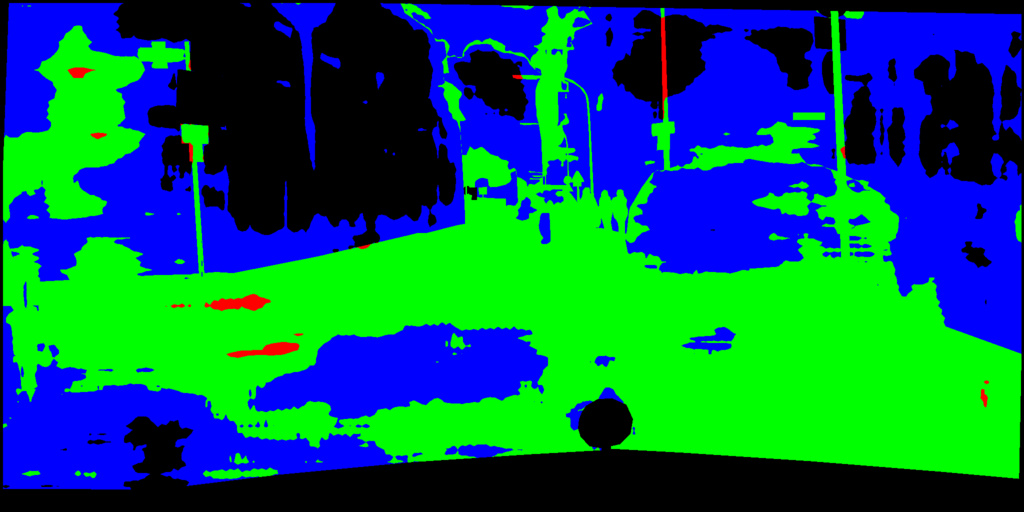}
    \end{subfigure}
    \begin{subfigure}{0.196\textwidth}
        \centering
        \includegraphics[width=\linewidth]{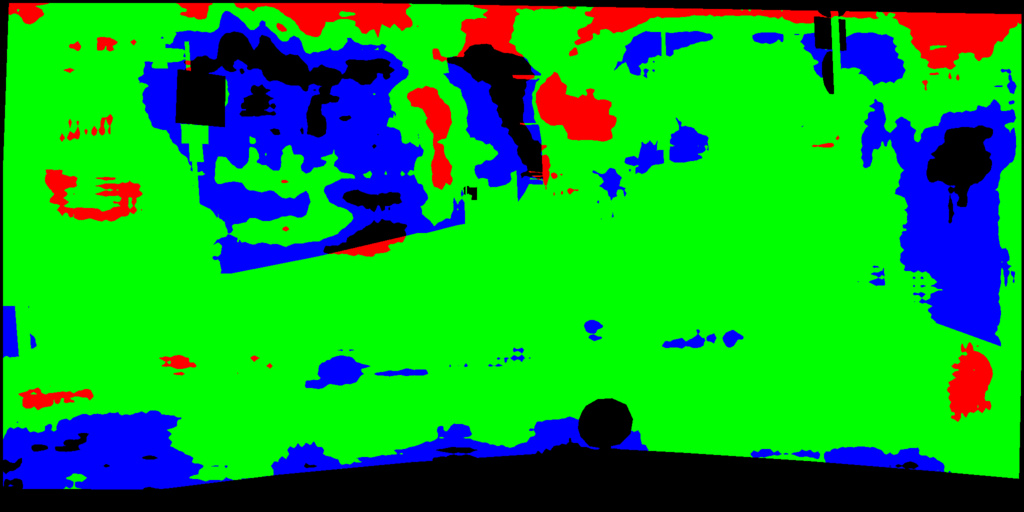}
    \end{subfigure}
    \caption{An example of how DRN misclassifies inputs when noise is added and how well entropy can capture it. From left to right, we have the original image followed by increasing amounts of noise added to it. The bottom row shows misclassified pixels with high entropy in green, misclassified pixels with low entropy in red and correctly classified pixels with high entropy in blue.}
    \label{fig:noise}
\end{figure*}

\begin{table*}[!h]
    \centering
    \caption{Precision and Recall on Cityscapes dataset using maximum percentage of pixels as thresholding strategy. All of the results shown use Entropy as the uncertainty metric.}
    \begin{tabular}{lrrrrrrrrr}
        \toprule
        Scenario & \multicolumn{3}{c}{Max 5\% pixels} & \multicolumn{3}{c}{Max 10\% pixels}& \multicolumn{3}{c}{Max 15\% pixels} \\
        & Pr & Rc & Px\% & Pr & Rc & Px\% & Pr & Rc & Px\% \\
        \midrule
        \multicolumn{10}{c}{DRN D-22} \\
        Base & \textbf{44.23} & 48.74 & 4.82 & 35.03 & 73.61 & 9.75 & 28.23 & 84.73 & 14.51 \\
        Noise & 44.20 & 48.77 & 4.82 & 35.02 & 73.64 & 9.76 & 28.23 & 84.78 & 14.51 \\
        Scale & 43.86 & 47.72 & 4.83 & 35.54 & 73.68 & 9.68 & 28.92 & \textbf{86.18} & 14.62 \\
        \midrule
        \multicolumn{10}{c}{OneFormer ConNeXt-L} \\
        Base & 31.47 & 54.09 & 4.60 & 24.11 & 74.15 & 8.61 & 20.51 & 79.86 & 11.28 \\
        Noise & 31.41 & 54.88 & 4.60 & 24.24 & 74.79 & 8.63 & 20.73 & 80.37 & 11.21 \\
        Scale & \textbf{31.81} & 55.45 & 4.69 & 23.77 & 76.48 & 9.01 & 19.49 & \textbf{82.95} & 12.32 \\
        Drop & 31.47 & 54.88 & 4.65 & 23.72 & 75.90 & 8.92 & 19.59 & 82.24 & 12.02 \\
        \midrule
        \multicolumn{10}{c}{SegFormer B5} \\
        Base & 36.71 & 63.98 & 4.89 & 25.99 & 83.01 & 9.38 & 22.02 & 88.66 & 12.05 \\
        Noise & 36.60 & 63.83 & 4.88 & 25.94 & 82.98 & 9.39 & 21.96 & 88.71 & 12.09 \\
        Scale & \textbf{38.15} & 65.23 & 4.84 & 26.57 & 85.68 & 9.67 & 20.54 & \textbf{92.12} & 13.69 \\
        Drop & 36.73 & 63.96 & 4.88 & 25.96 & 83.06 & 9.40 & 21.92 & 88.80 & 12.13 \\
        \bottomrule
    \end{tabular}
    \label{tab:px_max_cityscapes}
\end{table*}

\begin{table}[t]
    \centering
    \caption{AUROC for increasing noise levels on the Cityscapes dataset for DRN D-22 and OneFormer ConvNeXt-L models.}
    \begin{tabular}{lcc}
    \toprule
    Scenario & DRN D-22 & OneFormer ConvNeXt-L \\
    \midrule
    Base & 0.9337 & 0.8596 \\
    Noise 5 & 0.8732 & 0.8504 \\
    Noise 10 & 0.7934 & 0.8431 \\
    Noise 25 & 0.7041 & 0.8173 \\
    Noise 50 & 0.6213 & 0.7544 \\
    \bottomrule
    \end{tabular}
    \label{tab:noise}
\end{table}

We now take a look at how well our first method for dynamic thresholding works. \Cref{tab:largest_diff_cityscapes} shows the results of using the largest difference as a thresholding technique on the Cityscapes dataset. We use Entropy as the uncertainty metric. We observe that while we get very high recall rates, we also select quite a large fraction of pixels with values reaching as high as 40\% for the Scale setting with DRN. As a result, precision suffers quite a bit in these cases. However, we will see later that this technique is useful when the network produces highly inaccurate outputs, such as when there is a domain shift in the input.

We now consider the performance of the second thresholding method. \Cref{tab:px_max_cityscapes} shows the results for the same on the Cityscapes dataset. As expected, recall improves at the cost of precision across all scenarios when we increase the maximum fraction of pixels allowed. However, even with as few as 10\% pixels, we can get over 70\% recall using entropy as an uncertainty metric. Among the different scenarios, Scale performs better than the others, especially for Transformer-based networks.

\Cref{tab:pr_rc_dz} shows the results of using Cityscapes trained DRN and OneFormer models on the Dark Zurich dataset using Entropy as the uncertainty metric. This experiment is meant to test whether the uncertainty metrics work reliably when there is a domain shift of the inputs. The models perform poorly on this dataset, with mIoU of 0.0737 for DRN and 0.3752 for OneFormer. Even though the models perform very poorly, we note that we can still use Entropy to get good precision and recall rates using the largest difference method for dynamic thresholding. We also note that using multiple scales generally performs better than other algorithms.

As another test of input domain shift, we experiment by introducing varying levels of noise in the input and checking the performance of the networks on them. The noise introduced is Gaussian in nature with standard deviations of \(\{5, 10, 25, 50\}\). The noise is introduced into the image before any preprocessing. \Cref{tab:noise} shows the AUROC values in this scenario on DRN D-22 and OneFormer ConvNeXt-L models. We also provide a visual example of the same in \Cref{fig:noise}. As can be seen with increasing noise levels, the outputs become highly inaccurate, but we can still capture most of these using Entropy as the uncertainty metric.

\section{Conclusion}
In this paper, we present a new form of analysis for semantic segmentation tasks focused on trusting the output of such networks during test time. Our work is primarily concerned with evaluating existing uncertainty metrics and seeing how well they correlate with and are able to capture misclassifications.

Our analysis suggests that it is indeed possible to obtain valid evaluation measures that can be used to predict the test time performance of existing segmentation networks. We have shown that even simple measures such as Probability Margins and Entropy can be used as a proxy for capturing misclassified pixels with high recall rates. Because of their simplicity and generality, such tests can be used with nearly all existing architectures without any computational overhead and can help us know when the outputs are to be trusted.  

We believe that our work will provide new insights into the nature of semantic segmentation and help drive further research in evaluating the trustworthiness of these networks.

\bibliographystyle{named}
\bibliography{ijcai24}

\newpage
\onecolumn

\section*{Supplementary Material}

\subsection*{Cityscapes and ADE20K}
First, we provide all results related to Cityscapes and ADE20K datasets. We provide results for Dark Zurich in a separate section, as experiments on it involve input domain shift. Table~\ref{tab:perf} collates all the performance metrics for the Base setting of all networks. We compute Mean Intersection Over Union (mIoU), Expected Calibration Error (ECE) and Brier Score (BS). For all of the metrics, we show both micro-averaged (\(\mu\) Avg.) as well as macro-averaged (M Avg.) results.

As expected, DRN performs the worst among all the models, while OneFormer and SegFormer perform much better. We also note that DRN and SegFormer are relatively well-calibrated, but OneFormer is not. Results on Cityscapes are much better than on ADE20K since the latter is a much more challenging dataset with more variety and a much larger number of classes.

\begin{table}[h!]
    \centering
    \begin{tabular}{lcccccc}
        \toprule
        Model & \multicolumn{2}{c}{mIoU \(\uparrow\)} & \multicolumn{2}{c}{ECE \(\downarrow\)} & \multicolumn{2}{c}{BS \(\downarrow\)} \\
        & \(\mu\) Avg. & M Avg. & \(\mu\) Avg. & M Avg. & \(\mu\) Avg. & M Avg. \\
        \midrule
        \multicolumn{7}{c}{Cityscapes} \\
        DRN D-22 & 0.6790 & 0.5479 & 0.0380 & 0.0177 & 0.0808 & 0.0824 \\
        DRN D-105 & 0.7551 & 0.6330 & \textbf{0.0337} & 0.0171 & 0.0617 & 0.0631 \\
        OneFormer ConvNeXt-L & 0.8287 & 0.6733 & 0.7887 & 0.8383 & 0.8012 & 0.8011 \\
        OneFormer Swin-L & \textbf{0.8288} & 0.6682 & 0.7973 & 0.8400 & 0.8034 & 0.8034 \\
        SegFormer B-5 & 0.8239 & \textbf{0.6875} & 0.0631 & \textbf{0.0160} & \textbf{0.0516} & \textbf{0.0524} \\
        \midrule
        \multicolumn{7}{c}{ADE20K} \\
        OneFormer ConvNeXt-L & 0.5684 & 0.4854 & 0.8753 & 0.8195 & 0.9690 & 0.9686 \\
        OneFormer Swin-L & \textbf{0.5698} & \textbf{0.4921} & 0.8948 & 0.8261 & 0.9689 & 0.9687 \\
        SegFormer B-5 & 0.5100 & 0.4434 & \textbf{0.1051} & \textbf{0.1068} & \textbf{0.2646} & \textbf{0.2845} \\
        \bottomrule
    \end{tabular}
    \caption{Micro and macro-averaged (\(\mu\) and M, respectively) Mean Intersection Over Union (mIoU), Expected Calibration Error (ECE) and Brier Score (BS) on Cityscapes and ADE20K datasets.}
    \label{tab:perf}
\end{table}

Table~\ref{tab:cityscapes_auroc_d} and Table~\ref{tab:ade20k_auroc_d} show the Area Under Receiver Operating Characteristic (AUROC) as well as the Precision, Recall and fraction of selected pixels on Cityscapes and ADE20K datasets, respectively.

We observe that in most cases, Scale seems to provide the best AUROC and Recall, while Noise generally provides good Precision. However, in most cases, Base comes quite close to these values and has the advantage of being computationally cheaper. We also note that simpler metrics such as Variation Ratio, Probability Margin and Entropy are often better than or very close to more involved metrics such as Variance and BALD. The Recall values often reach higher than 90\% for Cityscapes and 70\% for ADE20K, although often at the cost of selecting too many pixels in the image.

\begin{longtable}{llcccccccc}
    \toprule
    Scenario & Uncertainty & \multicolumn{2}{c}{AUROC \(\uparrow\)} & \multicolumn{6}{c}{Largest Difference Thresholding} \\
    & Metric & & & \multicolumn{2}{c}{Precision \(\uparrow\)} & \multicolumn{2}{c}{Recall \(\uparrow\)} & \multicolumn{2}{c}{Pixel \%} \\
    & & \(\mu\) Avg. & M Avg. & \(\mu\) Avg. & M Avg. & \(\mu\) Avg. & M Avg. & \(\mu\) Avg. & M Avg. \\
    \midrule
    \multicolumn{10}{c}{DRN D-22} \\
    Base & Var. Ratio & 0.9118 & 0.9277 & 21.10 & 20.18 & 92.13 & 94.71 & 23.29 & 23.51 \\
    & Prob. Margin & 0.9152 & 0.9302 & 19.71 & 18.84 & 93.31 & 95.62 & 25.26 & 25.48 \\
    & Entropy & 0.9205 & 0.9337 & 17.52 & 16.74 & 95.04 & 96.89 & 28.94 & 29.21 \\
    Noise & Var. Ratio & 0.9119 & 0.9278 & 21.10 & 20.18 & 92.15 & 94.73 & 23.30 & 23.52 \\
    & Prob. Margin & 0.9152 & 0.9303 & 19.71 & 18.85 & 93.32 & 95.63 & 25.26 & 25.49 \\
    & Entropy & 0.9205 & 0.9337 & 17.52 & 16.74 & 95.05 & 96.90 & 28.95 & 29.22 \\
    & Avg. Var. & 0.7609 & 0.7722 & \textbf{37.97} & \textbf{36.82} & 57.26 & 59.58 & 08.04 & 08.13 \\
    & Max. Var. & 0.7614 & 0.7724 & 37.71 & 36.55 & 57.42 & 59.71 & 08.12 & 08.21 \\
    & BALD & 0.8196 & 0.8276 & 29.28 & 28.36 & 72.32 & 73.90 & 13.18 & 13.30 \\
    Scale & Var. Ratio & 0.9410 & \textbf{0.9466} & 15.90 & 15.22 & 97.74 & 98.46 & 32.80 & 33.08 \\
    & Prob. Margin & \textbf{0.9414} & 0.9465 & 14.80 & 14.19 & 98.14 & 98.76 & 35.36 & 35.66 \\
    & Entropy & 0.9398 & 0.9444 & 13.04 & 12.52 & \textbf{98.74} & \textbf{99.19} & 40.38 & 40.72 \\
    & Avg. Var. & 0.8973 & 0.8998 & 23.50 & 22.51 & 91.38 & 92.38 & 20.74 & 20.95 \\
    & Max. Var. & 0.8976 & 0.8995 & 23.18 & 22.20 & 91.69 & 92.65 & 21.11 & 21.31 \\
    & BALD & 0.9066 & 0.9075 & 16.49 & 15.82 & 96.37 & 97.10 & 31.17 & 31.44 \\
    \midrule
    \multicolumn{10}{c}{DRN D-105} \\
    Base & Var. Ratio & 0.8939 & 0.9206 & 23.04 & 22.19 & 85.84 & 90.61 & 14.76 & 14.87 \\
    & Prob. Margin & 0.8995 & 0.9247 & 21.77 & 20.94 & 87.36 & 91.77 & 15.90 & 16.02 \\
    & Entropy & 0.9094 & 0.9322 & 19.75 & 18.95 & 89.88 & 93.64 & 18.03 & 18.17 \\
    Noise & Var. Ratio & 0.8949 & 0.9209 & 23.09 & 22.22 & 86.03 & 90.70 & 14.76 & 14.87 \\
    & Prob. Margin & 0.9004 & 0.9250 & 21.82 & 20.98 & 87.54 & 91.84 & 15.90 & 16.01 \\
    & Entropy & 0.9103 & 0.9325 & 19.79 & 18.98 & 90.05 & 93.70 & 18.03 & 18.16 \\
    & Avg. Var. & 0.7471 & 0.7666 & \textbf{39.53} & \textbf{38.65} & 52.68 & 56.64 & 05.28 & 05.34 \\
    & Max. Var. & 0.7472 & 0.7664 & 39.46 & 38.56 & 52.71 & 56.61 & 05.29 & 05.36 \\
    & BALD & 0.8077 & 0.8249 & 32.48 & 31.63 & 66.59 & 70.05 & 08.12 & 08.21 \\
    Scale & Var. Ratio & 0.9341 & 0.9474 & 19.48 & 18.60 & 93.73 & 95.92 & 19.07 & 19.21 \\
    & Prob. Margin & 0.9366 & 0.9490 & 18.33 & 17.51 & 94.52 & 96.48 & 20.43 & 20.57 \\
    & Entropy & \textbf{0.9400} & \textbf{0.9509} & 16.45 & 15.71 & \textbf{95.79} & \textbf{97.33} & 23.07 & 23.23 \\
    & Avg. Var. & 0.8893 & 0.9007 & 27.81 & 26.58 & 84.51 & 87.00 & 12.04 & 12.15 \\
    & Max. Var. & 0.8900 & 0.9009 & 27.59 & 26.37 & 84.74 & 87.17 & 12.17 & 12.28 \\
    & BALD & 0.9111 & 0.9197 & 20.60 & 19.70 & 91.20 & 93.16 & 17.54 & 17.67 \\
    \midrule
    \multicolumn{10}{c}{OneFormer ConvNeXt-L} \\
    Base & Var. Ratio & 0.7265 & 0.8822 & 15.02 & 16.52 & 84.68 & 87.75 & 18.03 & 18.17 \\
    & Prob. Margin & 0.7893 & 0.9073 & 14.37 & 15.63 & 89.19 & 91.32 & 19.84 & 19.98 \\
    & Entropy & 0.7089 & 0.8569 & 14.13 & 16.23 & 81.30 & 84.88 & 18.40 & 18.65 \\
    Noise & Var.Ratio & 0.7334 & 0.8851 & 15.04 & 16.66 & 84.72 & 87.85 & 18.00 & 18.13 \\
    & Prob. Margin & 0.7927 & 0.9087 & 14.15 & 15.72 & 89.07 & 91.37 & 20.12 & 20.25 \\
    & Entropy & 0.7153 & 0.8612 & 14.25 & 16.54 & 80.60 & 85.05 & 18.08 & 18.21 \\
    & Avg. Var. & 0.6458 & 0.6474 & 18.71 & 28.78 & 34.52 & 35.06 & 05.90 & 05.99 \\
    & Max. Var. & 0.6323 & 0.6324 & 17.11 & 27.49 & 31.93 & 32.20 & 05.97 & 06.05 \\
    & BALD & 0.6727 & 0.6742 & 20.21 & \textbf{29.16} & 40.01 & 40.55 & 06.33 & 06.42 \\
    Scale & Var. Ratio & 0.7647 & 0.8972 & 12.96 & 14.53 & 88.64 & 90.63 & 21.87 & 22.06 \\
    & Prob. Margin & \textbf{0.8305} & \textbf{0.9215} & 12.58 & 13.53 & \textbf{92.36} & \textbf{93.71} & 23.46 & 23.68 \\
    & Entropy & 0.7350 & 0.8736 & 12.11 & 14.36 & 82.76 & 87.42 & 21.84 & 21.90 \\
    & Avg. Var. & 0.7835 & 0.7880 & 16.78 & 21.70 & 65.40 & 67.80 & 12.46 & 12.54 \\
    & Max. Var. & 0.7637 & 0.7677 & 15.75 & 20.82 & 62.04 & 64.34 & 12.59 & 12.67 \\
    & BALD & 0.8173 & 0.8213 & 17.66 & 21.76 & 71.71 & 74.12 & 12.98 & 13.05 \\
    Drop & Var. Ratio & 0.7471 & 0.8936 & 14.89 & 15.82 & 85.54 & 89.39 & 18.36 & 18.44 \\
    & Prob. Margin & 0.8059 & 0.9166 & 13.51 & 14.47 & 91.12 & 93.09 & 21.56 & 21.71 \\
    & Entropy & 0.7278 & 0.8712 & 13.76 & 15.12 & 82.96 & 87.21 & 19.28 & 19.40 \\
    & Avg. Var. & 0.7219 & 0.7264 & 21.08 & 27.12 & 49.93 & 52.42 & 07.57 & 07.68 \\
    & Max. Var. & 0.6994 & 0.7023 & 19.55 & 25.98 & 45.49 & 47.68 & 07.44 & 07.54 \\
    & BALD & 0.7624 & 0.7667 & \textbf{21.71} & 27.02 & 58.30 & 60.90 & 08.58 & 08.69 \\
    \midrule
    \multicolumn{10}{c}{OneFormer Swin-L} \\
    Base & Var. Ratio & 0.7379 & 0.8909 & 13.68 & 15.60 & 86.62 & 89.80 & 20.24 & 20.37 \\
    & Prob. Margin & 0.7937 & 0.9127 & 13.30 & 14.59 & 89.82 & 92.61 & 21.57 & 21.72 \\
    & Entropy & 0.7238 & 0.8677 & 13.57 & 15.52 & 82.85 & 86.61 & 19.51 & 19.73 \\
    Noise & Var.Ratio & 0.7432 & 0.8927 & 14.34 & 15.84 & 86.49 & 89.67 & 19.28 & 19.48 \\
    & Prob. Margin & 0.7976 & 0.9139 & 13.76 & 14.87 & 90.09 & 92.60 & 20.93 & 21.10 \\
    & Entropy & 0.7288 & 0.8709 & 13.95 & 15.60 & 83.60 & 87.08 & 19.15 & 19.39 \\
    & Avg. Var. & 0.6603 & 0.6611 & 23.56 & 29.87 & 36.01 & 36.33 & 04.88 & 05.00 \\
    & Max. Var. & 0.6467 & 0.6463 & 22.08 & 28.63 & 33.36 & 33.46 & 04.83 & 04.95 \\
    & BALD & 0.6866 & 0.6874 & \textbf{25.56} & \textbf{30.50} & 41.36 & 41.61 & 05.17 & 05.24 \\
    Scale & Var. Ratio & 0.7816 & 0.9109 & 12.76 & 13.84 & 89.92 & 92.58 & 22.52 & 22.69 \\
    & Prob. Margin & \textbf{0.8415} & \textbf{0.9304} & 11.87 & 12.90 & \textbf{93.29} & \textbf{95.15} & 25.12 & 25.40 \\
    & Entropy & 0.7557 & 0.8951 & 12.30 & 13.65 & 87.69 & 90.83 & 22.78 & 22.95 \\
    & Avg. Var. & 0.8093 & 0.8159 & 19.67 & 22.86 & 70.28 & 71.57 & 11.42 & 11.59 \\
    & Max. Var. & 0.7929 & 0.7985 & 18.86 & 22.18 & 67.53 & 68.59 & 11.44 & 11.61 \\
    & BALD & 0.8352 & 0.8421 & 20.03 & 22.73 & 75.31 & 76.74 & 12.02 & 12.20 \\
    Drop & Var. Ratio & 0.7543 & 0.8972 & 13.39 & 14.67 & 87.99 & 90.76 & 20.99 & 21.20 \\
    & Prob. Margin & 0.8078 & 0.9179 & 12.95 & 14.15 & 90.91 & 93.25 & 22.43 & 22.68 \\
    & Entropy & 0.7396 & 0.8779 & 12.60 & 14.31 & 85.42 & 88.66 & 21.65 & 21.85 \\
    & Avg. Var. & 0.7211 & 0.7397 & 23.02 & 28.08 & 49.79 & 53.67 & 06.91 & 07.04 \\
    & Max. Var. & 0.6994 & 0.7164 & 21.31 & 27.09 & 45.57 & 49.20 & 06.83 & 07.01 \\
    & BALD & 0.7593 & 0.7765 & 24.27 & 28.27 & 57.69 & 61.29 & 07.60 & 07.69 \\
    \midrule
    \multicolumn{10}{c}{SegFormer B-5} \\
    Base & Var. Ratio & 0.8830 & 0.9087 & 24.60 & 24.07 & 82.10 & 86.94 & 10.90 & 10.95 \\
    & Prob. Margin & 0.8892 & 0.9142 & 23.40 & 22.87 & 83.65 & 88.29 & 11.68 & 11.73 \\
    & Entropy & 0.8997 & 0.9234 & 21.43 & 20.89 & 86.20 & 90.47 & 13.14 & 13.20 \\
    Noise & Var.Ratio & 0.8833 & 0.9089 & 24.53 & 24.00 & 82.22 & 87.02 & 10.95 & 11.00 \\
    & Prob. Margin & 0.8896 & 0.9143 & 23.33 & 22.81 & 83.77 & 88.37 & 11.73 & 11.78 \\
    & Entropy & 0.9002 & 0.9236 & 21.37 & 20.84 & 86.34 & 90.55 & 13.20 & 13.26 \\
    & Avg. Var. & 0.6972 & 0.7147 & \textbf{40.19} & \textbf{40.11} & 41.61 & 45.11 & 03.38 & 03.40 \\
    & Max. Var. & 0.6965 & 0.7136 & 40.18 & 40.07 & 41.46 & 44.90 & 03.37 & 03.39 \\
    & BALD & 0.7520 & 0.7697 & 34.08 & 34.03 & 53.80 & 57.30 & 05.16 & 05.19 \\
    Scale & Var. Ratio & 0.9212 & 0.9391 & 20.33 & 19.77 & 90.13 & 93.35 & 14.48 & 14.56 \\
    & Prob. Margin & 0.9245 & 0.9418 & 19.21 & 18.67 & 91.02 & 94.10 & 15.48 & 15.56 \\
    & Entropy & \textbf{0.9290} & \textbf{0.9459} & 17.37 & 16.85 & \textbf{92.46} & \textbf{95.29} & 17.39 & 17.48 \\
    & Avg. Var. & 0.8678 & 0.8826 & 28.23 & 27.55 & 78.95 & 81.98 & 09.13 & 09.19 \\
    & Max. Var. & 0.8687 & 0.8832 & 28.04 & 27.36 & 79.20 & 82.18 & 09.23 & 09.28 \\
    & BALD & 0.8947 & 0.9091 & 21.36 & 20.81 & 86.42 & 89.33 & 13.22 & 13.28 \\
    Drop & Var. Ratio & 0.8842 & 0.9097 & 24.44 & 23.91 & 82.38 & 87.17 & 11.01 & 11.06 \\
    & Prob. Margin & 0.8904 & 0.9151 & 23.24 & 22.72 & 83.92 & 88.50 & 11.80 & 11.85 \\
    & Entropy & 0.9008 & 0.9242 & 21.28 & 20.74 & 86.46 & 90.66 & 13.27 & 13.33 \\
    & Avg. Var. & 0.7677 & 0.7931 & 37.02 & 36.66 & 56.64 & 61.71 & 05.00 & 05.02 \\
    & Max. Var. & 0.7660 & 0.7915 & 37.06 & 36.69 & 56.31 & 61.38 & 04.96 & 04.98 \\
    & BALD & 0.8255 & 0.8519 & 28.97 & 28.73 & 70.12 & 75.25 & 07.91 & 07.93 \\
    \bottomrule
    \caption{Micro and macro-averaged (\(\mu\) and M, respectively) Area Under Receiver Operating Characteristic (AUROC) and Precision, Recall and the fraction of pixels selected for largest difference thresholding on the Cityscapes dataset.}
    \label{tab:cityscapes_auroc_d}
\end{longtable}

\begin{longtable}{llcccccccc}
    \toprule
    Scenario & Uncertainty & \multicolumn{2}{c}{AUROC \(\uparrow\)} & \multicolumn{6}{c}{Largest Difference Thresholding} \\
    & Metric & & & \multicolumn{2}{c}{Precision \(\uparrow\)} & \multicolumn{2}{c}{Recall \(\uparrow\)} & \multicolumn{2}{c}{Pixel \%} \\
    & & \(\mu\) Avg. & M Avg. & \(\mu\) Avg. & M Avg. & \(\mu\) Avg. & M Avg. & \(\mu\) Avg. & M Avg. \\
    \midrule
    \multicolumn{10}{c}{OneFormer ConvNeXt-L} \\
    Base & Var. Ratio & 0.6067 & 0.6941 & 25.45 & 28.92 & 50.31 & 63.73 & 29.11 & 30.34 \\
    & Prob. Margin & 0.7048 & 0.7750 & 25.15 & 29.12 & 63.92 & 75.96 & 37.43 & 38.25 \\
    & Entropy & 0.5588 & 0.6164 & 23.07 & 27.39 & 41.38 & 54.30 & 26.42 & 27.64 \\
    Noise & Var. Ratio & 0.6052 & 0.6927 & 25.73 & 28.89 & 50.62 & 63.82 & 28.97 & 30.43 \\
    & Prob. Margin & 0.7039 & 0.7727 & 25.48 & 29.24 & 63.66 & 75.54 & 36.79 & 37.66 \\
    & Entropy & 0.5576 & 0.6151 & 23.34 & 27.23 & 41.43 & 54.34 & 26.13 & 27.57 \\
    & Avg. Var. & 0.6328 & 0.6235 & \textbf{30.67} & \textbf{35.27} & 41.61 & 45.27 & 19.98 & 22.40 \\
    & Max. Var. & 0.6184 & 0.6084 & 29.52 & 34.66 & 38.94 & 42.18 & 19.42 & 21.76 \\
    & BALD & 0.6218 & 0.6672 & 26.34 & 29.24 & 54.24 & 58.38 & 30.32 & 33.04 \\
    Scale & Var. Ratio & 0.6243 & 0.7078 & 24.23 & 28.00 & 52.29 & 65.56 & 31.78 & 32.76 \\
    & Prob. Margin & 0.7285 & \textbf{0.7883} & 24.87 & 28.95 & 65.42 & 76.88 & 38.74 & 39.19 \\
    & Entropy & 0.5693 & 0.6304 & 22.37 & 26.39 & 42.37 & 55.78 & 27.89 & 29.02 \\
    & Avg. Var. & 0.6880 & 0.6728 & 27.69 & 30.54 & 61.78 & 66.51 & 32.85 & 35.02 \\
    & Max. Var. & 0.6635 & 0.6489 & 26.92 & 29.84 & 58.47 & 62.64 & 31.98 & 34.25 \\
    & BALD & \textbf{0.7473} & 0.7279 & 28.66 & 30.16 & \textbf{73.48} & \textbf{77.93} & 37.76 & 39.69 \\
    Drop & Var. Ratio & 0.6136 & 0.6956 & 25.19 & 28.55 & 50.22 & 64.06 & 29.36 & 30.82 \\
    & Prob. Margin & 0.7151 & 0.7778 & 25.10 & 29.06 & 63.51 & 76.03 & 37.26 & 38.10 \\
    & Entropy & 0.5618 & 0.6169 & 22.72 & 26.59 & 40.91 & 54.33 & 26.51 & 27.99 \\
    & Avg. Var. & 0.6732 & 0.6636 & 29.53 & 33.02 & 54.03 & 59.30 & 26.95 & 29.28 \\
    & Max. Var. & 0.6517 & 0.6415 & 28.65 & 32.46 & 50.24 & 54.95 & 25.82 & 28.23 \\
    & BALD & 0.7359 & 0.7258 & 29.66 & 31.50 & 68.32 & 74.11 & 33.92 & 36.23 \\
    \midrule
    \multicolumn{10}{c}{OneFormer Swin-L} \\
    Base & Var. Ratio & 0.6115 & 0.6977 & 24.79 & 28.04 & 52.66 & 64.95 & 30.16 & 31.21 \\
    & Prob. Margin & 0.7066 & 0.7770 & 25.18 & 28.13 & 66.49 & 76.84 & 37.49 & 38.22 \\
    & Entropy & 0.5661 & 0.6222 & 22.96 & 26.26 & 43.63 & 55.65 & 26.97 & 28.53 \\
    Noise & Var. Ratio & 0.6110 & 0.6958 & 24.86 & 27.82 & 53.28 & 65.05 & 30.42 & 31.46 \\
    & Prob. Margin & 0.7066 & 0.7748 & 24.83 & 27.98 & 65.73 & 76.59 & 37.58 & 38.38 \\
    & Entropy & 0.5652 & 0.6208 & 22.82 & 26.20 & 43.62 & 55.28 & 27.13 & 28.37 \\
    & Avg. Var. & 0.6331 & 0.6281 & 28.40 & \textbf{34.11} & 43.92 & 47.86 & 21.95 & 24.05 \\
    & Max. Var. & 0.6180 & 0.6114 & 27.42 & 33.52 & 40.58 & 44.08 & 21.01 & 23.09 \\
    & BALD & 0.6510 & 0.6806 & 26.38 & 29.88 & 57.68 & 62.16 & 31.04 & 33.33 \\
    Scale & Var. Ratio & 0.6296 & 0.7142 & 23.84 & 27.27 & 55.82 & 67.39 & 33.23 & 33.60 \\
    & Prob. Margin & 0.7322 & \textbf{0.7933} & 24.05 & 27.69 & 68.38 & 78.83 & 40.37 & 40.47 \\
    & Entropy & 0.5742 & 0.6358 & 21.99 & 25.40 & 45.23 & 57.34 & 29.20 & 30.27 \\
    & Avg. Var. & 0.6994 & 0.6891 & 27.97 & 30.37 & 64.48 & 68.74 & 32.72 & 34.31 \\
    & Max. Var. & 0.6735 & 0.6643 & 26.78 & 29.50 & 60.42 & 64.77 & 32.02 & 33.75 \\
    & BALD & \textbf{0.7563} & 0.7427 & 28.37 & 29.68 & \textbf{74.23} & \textbf{79.18} & 37.15 & 38.73 \\
    Drop & Var. Ratio & 0.6173 & 0.7005 & 24.31 & 27.57 & 53.69 & 65.76 & 31.35 & 32.33 \\
    & Prob. Margin & 0.7156 & 0.7804 & 24.41 & 28.04 & 65.93 & 76.85 & 38.34 & 38.83 \\
    & Entropy & 0.5693 & 0.6241 & 22.34 & 25.76 & 43.40 & 55.83 & 27.57 & 28.91 \\
    & Avg. Var. & 0.6665 & 0.6565 & 27.86 & 31.72 & 55.14 & 59.57 & 28.10 & 30.11 \\
    & Max. Var. & 0.6459 & 0.6333 & 26.98 & 31.18 & 51.19 & 54.98 & 26.93 & 28.93 \\
    & BALD & 0.7302 & 0.7182 & \textbf{28.76} & 30.76 & 69.12 & 73.75 & 34.11 & 35.92 \\
    \midrule
    \multicolumn{10}{c}{SegFormer B-5} \\
    Base & Var. Ratio & 0.8000 & 0.8388 & 35.54 & 33.27 & 80.01 & 87.38 & 36.81 & 38.62 \\
    & Prob. Margin & 0.8024 & 0.8397 & 34.56 & 32.45 & 81.51 & 88.36 & 38.56 & 40.28 \\
    & Entropy & 0.8113 & 0.8484 & 32.20 & 30.34 & 84.63 & 90.97 & 42.96 & 44.99 \\
    Noise & Var. Ratio & 0.8356 & 0.8628 & 31.86 & 30.41 & 88.13 & 93.03 & 45.22 & 46.72 \\
    & Prob. Margin & 0.8353 & 0.8615 & 30.93 & 29.67 & 88.88 & 93.51 & 46.97 & 48.33 \\
    & Entropy & \textbf{0.8384} & \textbf{0.8661} & 28.80 & 27.87 & \textbf{90.24} & \textbf{94.82} & 51.21 & 52.85 \\
    & Avg. Var. & 0.7929 & 0.8096 & 39.47 & 37.55 & 77.79 & 83.11 & 32.22 & 33.33 \\
    & Max. Var. & 0.7929 & 0.8089 & 39.28 & 37.40 & 78.06 & 83.31 & 32.48 & 33.60 \\
    & BALD & 0.8141 & 0.8266 & 32.97 & 31.46 & 86.96 & 91.18 & 43.11 & 44.50 \\
    Scale & Var. Ratio & 0.8003 & 0.8388 & 35.55 & 33.26 & 80.02 & 87.38 & 36.79 & 38.59 \\
    & Prob. Margin & 0.8028 & 0.8397 & 34.57 & 32.44 & 81.51 & 88.35 & 38.54 & 40.24 \\
    & Entropy & 0.8115 & 0.8483 & 32.12 & 30.34 & 84.27 & 90.91 & 42.89 & 44.93 \\
    & Avg. Var. & 0.7123 & 0.7490 & \textbf{48.46} & \textbf{45.09} & 52.74 & 61.89 & 17.79 & 19.46 \\
    & Max. Var. & 0.7116 & 0.7483 & 48.38 & 45.03 & 52.65 & 61.80 & 17.79 & 19.45 \\
    & BALD & 0.7627 & 0.7952 & 41.45 & 38.38 & 68.51 & 76.96 & 27.02 & 29.59 \\
    Drop & Var. Ratio & 0.8016 & 0.8398 & 35.32 & 33.08 & 80.47 & 87.70 & 37.24 & 39.05 \\
    & Prob. Margin & 0.8039 & 0.8407 & 34.34 & 32.26 & 81.92 & 88.65 & 38.99 & 40.71 \\
    & Entropy & 0.8125 & 0.8491 & 31.92 & 30.16 & 84.70 & 91.18 & 43.37 & 45.43 \\
    & Avg. Var. & 0.7521 & 0.7955 & 44.41 & 41.56 & 63.76 & 73.73 & 23.47 & 24.96 \\
    & Max. Var. & 0.7520 & 0.7955 & 44.35 & 41.52 & 63.85 & 73.85 & 23.54 & 25.04 \\
    & BALD & 0.7914 & 0.8299 & 36.50 & 34.17 & 78.07 & 86.38 & 34.96 & 37.22 \\
    \bottomrule
    \caption{Micro and macro-averaged (\(\mu\) and M, respectively) Area Under Receiver Operating Characteristic (AUROC) and Precision, Recall and the fraction of pixels selected for largest difference thresholding on the ADE20K dataset.}
    \label{tab:ade20k_auroc_d}
\end{longtable}

Finally, Table~\ref{tab:cityscapes_max_pixel} and Table~\ref{tab:ade20k_max_pixel} show Precision and Recall for thresholding based on a maximum fraction of pixels allowed on Cityscapes and ADE20K datasets, respectively. We show results across three threshold values.

As expected, we see that as we increase the maximum number of pixels allowed, we get higher Recall at the cost of Precision. This observation holds true across all of the experiments. Again, we note that Scale often has the best performance with simple uncertainty metrics across both datasets.
\begin{longtable}{llcccccccccccc}
    \toprule
    Scenario & Unc. & \multicolumn{4}{c}{Max 5\% pixels} & \multicolumn{4}{c}{Max 10\% pixels} & \multicolumn{4}{c}{Max 15\% pixels} \\
    & Metric & \multicolumn{2}{c}{Precision \(\uparrow\)} & \multicolumn{2}{c}{Recall \(\uparrow\)} & \multicolumn{2}{c}{Precision \(\uparrow\)} & \multicolumn{2}{c}{Recall \(\uparrow\)} & \multicolumn{2}{c}{Precision \(\uparrow\)} & \multicolumn{2}{c}{Recall \(\uparrow\)} \\
    & & \(\mu\) & M & \(\mu\) & M & \(\mu\) & M & \(\mu\) & M & \(\mu\) & M & \(\mu\) & M \\
    \midrule
    \multicolumn{14}{c}{DRN D-22} \\
    Base & VR & 45.19 & 45.40 & 41.52 & 50.74 & 35.04 & 35.18 & 64.13 & 73.92 & 28.48 & 28.45 & 76.44 & 84.64 \\
    & PM & 44.66 & 44.83 & 41.35 & 50.58 & 34.77 & 34.92 & 64.13 & 73.95 & 28.14 & 28.19 & 76.51 & 84.74 \\
    & E & 44.04 & 44.23 & 39.76 & 48.74 & 34.80 & 35.03 & 63.64 & 73.61 & 28.11 & 28.23 & 76.42 & 84.73 \\
    Noise & VR & 45.15 & 45.35 & 41.51 & 50.75 & 35.01 & 35.16 & 64.10 & 73.94 & 28.46 & 28.44 & 76.39 & 84.64 \\
    & PM & 44.60 & 44.78 & 41.27 & 50.56 & 34.76 & 34.91 & 64.07 & 73.93 & 28.15 & 28.18 & 76.50 & 84.77 \\
    & E & 44.02 & 44.20 & 39.75 & 48.77 & 34.79 & 35.02 & 63.64 & 73.64 & 28.12 & 28.23 & 76.45 & 84.78 \\
    & AV & 40.06 & 39.93 & 32.11 & 38.45 & 38.03 & 37.47 & 50.03 & 55.62 & 37.92 & 36.97 & 55.23 & 58.76 \\
    & MV & 39.79 & 39.66 & 31.88 & 38.17 & 37.81 & 37.22 & 49.85 & 55.45 & 37.65 & 36.70 & 55.34 & 58.86 \\
    & B & 36.39 & 36.38 & 30.65 & 36.02 & 32.57 & 32.42 & 50.90 & 57.46 & 30.13 & 29.82 & 63.10 & 68.56 \\
    Scale & VR & \textbf{47.90} & \textbf{48.21} & 43.77 & 52.63 & 36.67 & 36.94 & 67.41 & 76.23 & 29.11 & 29.28 & 79.80 & \textbf{86.84} \\
    & PM & 47.17 & 47.44 & 43.46 & 52.41 & 36.33 & 36.59 & 67.23 & 76.08 & 28.92 & 29.10 & \textbf{79.84} & 86.82 \\
    & E & 43.68 & 43.86 & 39.53 & 47.72 & 35.26 & 35.54 & 63.98 & 73.68 & 28.67 & 28.92 & 78.57 & 86.18 \\
    & AV & 32.12 & 32.29 & 29.36 & 34.35 & 30.02 & 30.11 & 54.35 & 62.02 & 27.29 & 27.15 & 71.54 & 78.61 \\
    & MV & 31.42 & 31.55 & 28.76 & 33.67 & 29.60 & 29.67 & 53.63 & 61.29 & 27.02 & 26.87 & 71.09 & 78.23 \\
    & B & 31.13 & 31.32 & 28.39 & 32.45 & 28.38 & 28.55 & 51.55 & 57.83 & 25.34 & 25.41 & 68.24 & 74.57 \\
    \midrule
    \multicolumn{14}{c}{DRN D-105} \\
    Base & VR & 39.63 & 39.78 & 49.13 & 59.67 & 29.24 & 29.21 & 70.10 & 80.33 & 24.61 & 24.23 & 79.31 & 87.50 \\
    & PM & 39.32 & 39.46 & 48.98 & 59.54 & 28.83 & 28.86 & 70.23 & 80.53 & 23.91 & 23.60 & 79.69 & 87.97 \\
    & E & 39.72 & 39.95 & 48.70 & 59.33 & 28.85 & 28.94 & 70.48 & 80.80 & 23.15 & 22.99 & 80.26 & 88.63 \\
    Noise & VR & 39.57 & 39.70 & 49.01 & 59.57 & 29.24 & 29.19 & 70.09 & 80.33 & 24.62 & 24.24 & 79.34 & 87.55 \\
    & PM & 39.23 & 39.35 & 48.89 & 59.48 & 28.82 & 28.83 & 70.23 & 80.55 & 23.94 & 23.62 & 79.72 & 88.01 \\
    & E & 39.66 & 39.90 & 48.58 & 59.24 & 28.83 & 28.92 & 70.51 & 80.82 & 23.17 & 23.00 & 80.32 & 88.67 \\
    & AV & 40.51 & 40.32 & 40.62 & 48.14 & 39.37 & 38.74 & 50.96 & 56.09 & 39.41 & 38.64 & 52.14 & 56.53 \\
    & MV & 40.36 & 40.17 & 40.64 & 48.16 & 39.28 & 38.64 & 51.00 & 56.08 & 39.34 & 38.55 & 52.17 & 56.51 \\
    & B & 37.20 & 37.19 & 40.30 & 47.75 & 32.99 & 32.68 & 59.12 & 66.07 & 32.29 & 31.73 & 64.62 & 69.48 \\
    Scale & VR & \textbf{42.41} & \textbf{42.67} & 52.51 & 62.94 & 30.32 & 30.47 & 74.36 & 83.62 & 23.97 & 23.82 & 84.09 & 91.08 \\
    & PM & 42.07 & 42.31 & 52.42 & 62.88 & 30.07 & 30.23 & 74.33 & 83.65 & 23.48 & 23.45 & \textbf{84.15} & 91.18 \\
    & E & 41.11 & 41.40 & 49.76 & 60.38 & 30.00 & 30.29 & 73.84 & 83.50 & 23.16 & 23.24 & 84.14 & \textbf{91.32} \\
    & AV. & 35.74 & 35.87 & 43.55 & 51.63 & 30.48 & 30.24 & 68.84 & 77.10 & 28.21 & 27.52 & 79.36 & 84.93 \\
    & MV. & 35.35 & 35.49 & 43.03 & 51.07 & 30.32 & 30.07 & 68.69 & 76.96 & 28.03 & 27.36 & 79.33 & 84.95 \\
    & B & 33.45 & 33.64 & 40.73 & 47.25 & 27.51 & 27.57 & 65.47 & 73.22 & 23.51 & 23.33 & 79.31 & 85.81 \\
    \midrule
    \multicolumn{14}{c}{OneFormer ConvNeXt-L} \\
    Base & VR & 32.16 & 32.31 & 47.73 & 57.40 & 24.77 & 24.84 & 67.92 & 77.28 & 20.80 & 21.03 & 74.47 & 82.96 \\
    & PM & 33.41 & 33.78 & 50.18 & 60.30 & 24.95 & 24.99 & 71.11 & 80.37 & 20.58 & 20.68 & 79.04 & 86.67 \\
    & E & 30.89 & 31.47 & 44.48 & 54.09 & 23.82 & 24.11 & 64.18 & 74.15 & 20.07 & 20.51 & 70.79 & 79.86 \\
    Noise & VR & 32.43 & 32.39 & 48.17 & 57.91 & 24.70 & 24.74 & 67.97 & 77.59 & 20.72 & 20.89 & 74.96 & 83.38 \\
    & PM & 33.53 & 33.86 & 50.60 & 60.76 & 24.90 & 25.00 & 70.69 & 80.41 & 20.82 & 20.85 & 79.28 & 86.78 \\
    & E & 31.29 & 31.41 & 45.12 & 54.88 & 24.19 & 24.24 & 65.25 & 74.79 & 20.46 & 20.73 & 71.70 & 80.37 \\
    & AV & 28.32 & 30.65 & 23.29 & 27.49 & 25.39 & 29.84 & 27.17 & 30.78 & 24.23 & 29.78 & 28.87 & 32.20 \\
    & MV & 26.90 & 29.38 & 21.05 & 24.74 & 23.40 & 28.55 & 24.70 & 27.96 & 22.19 & 28.49 & 26.33 & 29.30 \\
    & B & 29.88 & 31.11 & 27.14 & 31.75 & 26.94 & 30.33 & 31.97 & 35.82 & 25.64 & 30.16 & 34.28 & 37.59 \\
    Scale & VR & 33.06 & 33.12 & 49.59 & 59.05 & 24.31 & 24.37 & 70.19 & 79.48 & 20.03 & 20.18 & 77.58 & 85.63 \\
    & PM & \textbf{34.95} & \textbf{35.08} & 53.07 & 62.79 & 24.83 & 24.94 & 73.38 & 82.40 & 20.01 & 20.04 & \textbf{81.97} & \textbf{88.92} \\
    & E & 31.49 & 31.81 & 46.19 & 55.45 & 23.63 & 23.77 & 66.65 & 76.48 & 19.21 & 19.49 & 73.98 & 82.95 \\
    & AV & 26.24 & 26.37 & 34.74 & 41.71 & 23.57 & 24.22 & 50.06 & 57.43 & 22.03 & 23.46 & 54.38 & 61.20 \\
    & MV & 24.47 & 24.64 & 31.43 & 38.02 & 22.32 & 23.08 & 46.09 & 53.41 & 20.73 & 22.42 & 50.57 & 57.34 \\
    & B & 27.69 & 27.71 & 38.17 & 45.61 & 24.49 & 24.68 & 55.98 & 63.55 & 22.87 & 23.67 & 60.90 & 67.73 \\
    Drop & VR & 32.26 & 32.34 & 48.26 & 57.81 & 24.32 & 24.33 & 68.79 & 78.27 & 19.99 & 20.11 & 76.25 & 84.80 \\
    & PM & 33.41 & 33.65 & 50.55 & 60.60 & 24.66 & 24.76 & 71.27 & 80.90 & 20.26 & 20.21 & 80.70 & 87.94 \\
    & E & 31.03 & 31.47 & 45.20 & 54.88 & 23.63 & 23.72 & 65.94 & 75.90 & 19.41 & 19.59 & 72.95 & 82.24 \\
    & AV & 27.22 & 28.30 & 29.97 & 37.42 & 25.57 & 27.51 & 39.85 & 47.02 & 24.53 & 27.53 & 41.76 & 48.59 \\
    & MV & 25.91 & 26.88 & 27.07 & 33.79 & 24.04 & 26.38 & 35.73 & 42.43 & 23.02 & 26.34 & 37.48 & 43.91 \\
    & B & 28.84 & 29.05 & 34.44 & 42.68 & 26.69 & 27.84 & 46.91 & 54.52 & 25.52 & 27.71 & 49.72 & 56.72 \\
    \midrule
    \multicolumn{14}{c}{OneFormer Swin-L} \\
    Base & VR & 32.94 & 33.23 & 49.05 & 57.91 & 24.85 & 24.96 & 69.57 & 77.97 & 20.84 & 20.77 & 78.74 & 84.70 \\
    & PM & 34.55 & 34.74 & 51.98 & 61.02 & 25.30 & 25.40 & 72.51 & 80.80 & 20.86 & 20.73 & 81.77 & 87.42 \\
    & E & 31.81 & 32.27 & 45.96 & 54.81 & 24.24 & 24.56 & 66.79 & 75.25 & 20.25 & 20.52 & 75.56 & 81.89 \\
    Noise & VR & 33.11 & 33.47 & 49.38 & 58.24 & 24.97 & 25.05 & 70.12 & 78.27 & 20.99 & 20.94 & 79.10 & 84.90 \\
    & PM & 34.73 & 34.88 & 52.24 & 61.27 & 25.36 & 25.39 & 72.93 & 80.98 & 20.70 & 20.58 & 81.90 & 87.55 \\
    & E & 32.04 & 32.58 & 46.64 & 55.36 & 24.33 & 24.57 & 67.07 & 75.50 & 20.33 & 20.55 & 76.08 & 82.28 \\
    & AV & 29.87 & 31.00 & 26.77 & 30.33 & 27.68 & 30.23 & 31.40 & 34.14 & 26.58 & 30.07 & 32.79 & 35.08 \\
    & MV & 28.28 & 29.58 & 24.38 & 27.57 & 26.22 & 28.98 & 28.79 & 31.26 & 25.39 & 28.89 & 30.04 & 32.11 \\
    & B & 31.24 & 31.81 & 30.48 & 34.60 & 28.89 & 30.78 & 36.27 & 39.23 & 27.91 & 30.66 & 38.01 & 40.31 \\
    Scale & VR & 33.94 & 34.32 & 50.84 & 59.78 & 25.02 & 25.08 & 73.38 & 80.83 & 20.08 & 19.99 & 82.28 & 87.82 \\
    & PM & \textbf{35.94} & \textbf{36.36} & 54.53 & 63.51 & 25.64 & 25.65 & 76.29 & 83.36 & 19.98 & 19.84 & \textbf{84.67} & \textbf{90.01} \\
    & E & 32.44 & 32.95 & 47.77 & 56.50 & 24.29 & 24.38 & 70.42 & 78.41 & 19.44 & 19.53 & 79.54 & 85.80 \\
    & AV & 27.75 & 27.90 & 37.81 & 44.46 & 24.89 & 25.01 & 57.01 & 62.72 & 23.77 & 24.16 & 63.79 & 67.53 \\
    & MV & 25.97 & 26.06 & 34.85 & 40.95 & 23.92 & 24.09 & 53.77 & 59.28 & 22.74 & 23.24 & 60.39 & 64.22 \\
    & B & 29.08 & 29.14 & 41.04 & 47.95 & 25.57 & 25.34 & 61.88 & 67.78 & 24.06 & 24.11 & 69.03 & 72.82 \\
    Drop & VR & 32.99 & 33.47 & 49.35 & 58.22 & 24.84 & 24.91 & 70.67 & 78.76 & 20.58 & 20.55 & 79.87 & 85.54 \\
    & PM & 34.51 & 34.86 & 52.25 & 61.35 & 25.26 & 25.31 & 73.34 & 81.33 & 20.42 & 20.32 & 82.53 & 88.16 \\
    & E & 31.84 & 32.43 & 46.51 & 55.32 & 23.95 & 24.13 & 67.69 & 76.26 & 19.89 & 19.97 & 77.17 & 83.17 \\
    & AV & 29.79 & 30.33 & 33.67 & 40.77 & 27.84 & 29.11 & 43.91 & 50.12 & 26.93 & 28.82 & 46.11 & 51.68 \\
    & MV & 28.13 & 28.72 & 30.60 & 37.09 & 26.50 & 27.96 & 40.00 & 45.77 & 25.88 & 27.83 & 41.88 & 47.13 \\
    & B & 31.21 & 31.36 & 38.42 & 46.14 & 28.73 & 29.50 & 50.96 & 57.41 & 27.71 & 29.06 & 53.38 & 59.08 \\
    \midrule
    \multicolumn{14}{c}{SegFormer B-5} \\
    Base & VR & 36.62 & 36.67 & 54.75 & 63.87 & 27.19 & 26.95 & 74.08 & 81.94 & 24.94 & 24.49 & 80.02 & 86.05 \\
    & PM & 36.39 & 36.44 & 54.79 & 63.96 & 26.68 & 26.48 & 74.42 & 82.37 & 23.88 & 23.45 & 81.16 & 87.19 \\
    & E & 36.61 & 36.71 & 54.78 & 63.98 & 26.09 & 25.99 & 74.94 & 83.01 & 22.42 & 22.02 & 82.52 & 88.66 \\
    Noise & VR & 36.50 & 36.54 & 54.61 & 63.79 & 27.16 & 26.91 & 74.01 & 81.90 & 24.87 & 24.42 & 80.04 & 86.09 \\
    & PM & 36.29 & 36.34 & 54.60 & 63.82 & 26.65 & 26.45 & 74.41 & 82.36 & 23.82 & 23.39 & 81.15 & 87.21 \\
    & E & 36.49 & 36.60 & 54.55 & 63.83 & 26.04 & 25.94 & 74.84 & 82.98 & 22.36 & 21.96 & 82.56 & 88.71 \\
    & AV & \textbf{40.42} & \textbf{40.39} & 38.53 & 43.44 & 40.25 & 40.13 & 41.31 & 45.00 & 40.19 & 40.11 & 41.61 & 45.11 \\
    & MV & \textbf{40.42} & 40.36 & 38.38 & 43.21 & 40.19 & 40.07 & 41.25 & 44.84 & 40.18 & 40.07 & 41.46 & 44.90 \\
    & B & 35.96 & 36.06 & 42.17 & 48.35 & 34.23 & 34.20 & 52.27 & 56.64 & 34.07 & 34.05 & 53.50 & 57.22 \\
    Scale & VR & 38.52 & 38.66 & 57.84 & 66.40 & 27.03 & 26.98 & 78.45 & 85.47 & 22.33 & 21.92 & 86.20 & 91.23 \\
    & PM & 38.36 & 38.49 & 57.89 & 66.44 & 26.68 & 26.68 & 78.68 & 85.68 & 21.55 & 21.23 & 86.60 & 91.67 \\
    & E & 37.88 & 38.15 & 56.12 & 65.23 & 26.49 & 26.57 & 78.44 & 85.68 & 20.75 & 20.54 & \textbf{86.91} & \textbf{92.12} \\
    & AV & 34.28 & 34.28 & 49.87 & 57.04 & 29.54 & 29.07 & 72.25 & 78.02 & 28.45 & 27.81 & 77.68 & 81.50 \\
    & MV & 34.09 & 34.09 & 49.70 & 56.83 & 29.38 & 28.91 & 72.37 & 78.12 & 28.26 & 27.62 & 77.88 & 81.69 \\
    & B & 31.46 & 31.51 & 46.09 & 52.15 & 25.41 & 25.31 & 71.04 & 77.64 & 22.48 & 22.11 & 81.72 & 86.61 \\
    Drop & VR & 36.62 & 36.67 & 54.79 & 63.92 & 27.17 & 26.93 & 74.15 & 82.01 & 24.81 & 24.35 & 80.25 & 86.25 \\
    & PM & 36.42 & 36.47 & 54.80 & 63.96 & 26.64 & 26.45 & 74.51 & 82.45 & 23.76 & 23.33 & 81.35 & 87.36 \\
    & E & 36.61 & 36.73 & 54.74 & 63.96 & 26.05 & 25.96 & 74.96 & 83.06 & 22.31 & 21.92 & 82.68 & 88.80 \\
    & AV & 38.44 & 38.37 & 47.21 & 54.70 & 37.02 & 36.69 & 56.13 & 61.53 & 37.02 & 36.66 & 56.64 & 61.71 \\
    & MV & 38.39 & 38.33 & 46.98 & 54.43 & 37.06 & 36.72 & 55.80 & 61.20 & 37.06 & 36.69 & 56.31 & 61.38 \\
    & B & 34.73 & 34.82 & 45.35 & 52.84 & 29.60 & 29.53 & 65.38 & 72.44 & 28.96 & 28.77 & 69.43 & 75.02 \\
    \bottomrule
    \caption{Micro and macro-averaged (\(\mu\) and M, respectively) Precision and Recall for thresholding based on a maximum fraction of pixels allowed on the Cityscapes dataset.}
    \label{tab:cityscapes_max_pixel}
\end{longtable}

\begin{longtable}{llcccccccccccc}
    \toprule
    Scenario & Unc. & \multicolumn{4}{c}{Max 5\% pixels} & \multicolumn{4}{c}{Max 10\% pixels} & \multicolumn{4}{c}{Max 15\% pixels} \\
    & Metric & \multicolumn{2}{c}{Precision \(\uparrow\)} & \multicolumn{2}{c}{Recall \(\uparrow\)} & \multicolumn{2}{c}{Precision \(\uparrow\)} & \multicolumn{2}{c}{Recall \(\uparrow\)} & \multicolumn{2}{c}{Precision \(\uparrow\)} & \multicolumn{2}{c}{Recall \(\uparrow\)} \\
    & & \(\mu\) & M & \(\mu\) & M & \(\mu\) & M & \(\mu\) & M & \(\mu\) & M & \(\mu\) & M \\
    \midrule
    \multicolumn{14}{c}{OneFormer ConvNeXt-L} \\
    Base & VR & 47.06 & 48.64 & 13.71 & 26.63 & 41.24 & 42.80 & 22.95 & 39.37 & 37.98 & 39.10 & 30.06 & 47.02 \\
    & PM & 50.10 & 52.72 & 13.67 & 27.27 & 44.99 & 47.46 & 24.82 & 42.06 & 41.45 & 43.45 & 34.19 & 51.95 \\
    & E & 43.42 & 45.63 & 11.38 & 23.14 & 38.09 & 40.24 & 19.18 & 34.30 & 34.36 & 36.72 & 24.40 & 40.23 \\
    Noise & VR & 46.82 & 48.40 & 13.60 & 26.50 & 40.80 & 42.62 & 22.69 & 39.25 & 37.48 & 38.92 & 29.51 & 46.74 \\
    & PM & 49.72 & 51.84 & 13.56 & 26.82 & 44.48 & 46.83 & 24.65 & 41.97 & 41.35 & 43.13 & 34.27 & 51.96 \\
    & E & 43.29 & 45.73 & 11.36 & 23.11 & 37.44 & 39.87 & 18.91 & 34.23 & 34.22 & 36.50 & 24.11 & 40.07 \\
    & AV & 35.66 & 37.79 & 08.32 & 14.39 & 34.12 & 36.78 & 14.30 & 21.51 & 32.70 & 36.13 & 18.94 & 26.12 \\
    & MV & 34.75 & 37.29 & 07.96 & 13.66 & 33.20 & 36.16 & 13.66 & 20.13 & 31.75 & 35.70 & 17.86 & 24.32 \\
    & B & 37.41 & 37.75 & 09.20 & 14.77 & 34.55 & 35.70 & 16.07 & 23.16 & 33.63 & 34.57 & 22.48 & 29.42 \\
    Scale & VR & 48.64 & 50.01 & 14.37 & 27.55 & 41.80 & 43.48 & 23.98 & 40.57 & 38.22 & 39.40 & 31.20 & 48.36 \\
    & PM & \textbf{51.85} & \textbf{54.22} & 14.63 & 28.57 & 45.91 & 48.29 & 26.21 & 43.75 & 41.99 & 44.14 & \textbf{35.70} & \textbf{53.54} \\
    & E & 44.02 & 46.12 & 11.83 & 23.81 & 37.77 & 39.70 & 19.77 & 35.24 & 34.16 & 35.94 & 25.65 & 41.55 \\
    & AV & 33.02 & 33.00 & 08.90 & 15.87 & 31.13 & 32.04 & 16.00 & 25.62 & 30.02 & 31.46 & 22.35 & 32.39 \\
    & MV & 31.15 & 31.46 & 08.22 & 14.56 & 29.56 & 30.60 & 14.87 & 23.66 & 28.26 & 30.01 & 20.47 & 29.77 \\
    & B & 37.05 & 37.00 & 10.57 & 18.79 & 34.67 & 34.90 & 19.06 & 30.30 & 32.94 & 33.54 & 26.15 & 38.40 \\
    Drop & VR & 47.23 & 48.70 & 13.88 & 26.96 & 41.36 & 42.79 & 23.23 & 39.93 & 37.82 & 38.97 & 30.23 & 47.41 \\
    & PM & 50.62 & 52.55 & 14.06 & 27.53 & 45.01 & 47.40 & 25.44 & 42.56 & 41.51 & 43.43 & 34.83 & 52.39 \\
    & E & 43.48 & 45.72 & 11.61 & 23.58 & 37.55 & 39.90 & 19.32 & 34.64 & 34.41 & 36.36 & 25.01 & 40.94 \\
    & AV & 32.91 & 34.23 & 08.39 & 15.68 & 31.96 & 33.95 & 15.30 & 24.70 & 30.97 & 33.53 & 21.32 & 31.35 \\
    & MV & 31.38 & 33.16 & 07.89 & 14.49 & 30.31 & 32.86 & 14.19 & 22.54 & 29.59 & 32.70 & 19.63 & 28.57 \\
    & B & 35.57 & 36.53 & 09.66 & 18.33 & 34.25 & 35.34 & 18.07 & 29.48 & 33.37 & 34.64 & 25.62 & 38.02 \\
    \midrule
    \multicolumn{14}{c}{OneFormer Swin-L} \\
    Base & VR & 46.97 & 48.39 & 14.18 & 27.16 & 41.28 & 42.31 & 23.92 & 40.21 & 38.27 & 38.78 & 31.45 & 47.94 \\
    & PM & 49.35 & 51.85 & 14.19 & 27.81 & 44.07 & 46.13 & 25.34 & 42.57 & 41.18 & 42.29 & 35.43 & 52.68 \\
    & E & 43.36 & 45.43 & 11.87 & 23.72 & 37.71 & 39.28 & 20.01 & 34.95 & 34.84 & 35.71 & 26.00 & 41.07 \\
    Noise & VR & 46.76 & 48.02 & 14.18 & 26.98 & 41.26 & 42.21 & 24.00 & 40.25 & 38.13 & 38.55 & 31.40 & 47.83 \\
    & PM & 49.39 & 51.30 & 14.30 & 27.85 & 44.03 & 45.67 & 25.47 & 42.60 & 40.67 & 41.89 & 35.15 & 52.55 \\
    & E & 42.99 & 45.13 & 11.89 & 23.66 & 37.56 & 39.34 & 19.86 & 34.81 & 34.67 & 35.65 & 26.07 & 41.10 \\
    & AV & 33.66 & 35.65 & 08.40 & 14.71 & 32.44 & 35.30 & 14.46 & 22.00 & 30.69 & 34.96 & 19.15 & 27.13 \\
    & MV & 32.74 & 34.93 & 07.97 & 13.72 & 31.41 & 34.61 & 13.67 & 20.48 & 29.65 & 34.08 & 18.00 & 25.09 \\
    & B & 36.48 & 37.19 & 09.60 & 16.23 & 33.93 & 35.36 & 17.18 & 25.26 & 32.10 & 34.11 & 23.16 & 31.76 \\
    Scale & VR & 48.38 & 49.56 & 14.85 & 28.10 & 41.74 & 42.90 & 25.00 & 41.66 & 38.39 & 39.05 & 32.99 & 49.81 \\
    & PM & \textbf{51.02} & \textbf{52.71} & 15.09 & 28.78 & 45.13 & 46.90 & 27.08 & 44.40 & 41.36 & 42.87 & \textbf{36.75} & \textbf{54.48} \\
    & E & 43.48 & 45.43 & 12.25 & 24.32 & 37.57 & 39.17 & 20.63 & 35.87 & 34.33 & 35.33 & 26.97 & 42.45 \\
    & AV & 34.14 & 34.12 & 09.65 & 17.18 & 32.33 & 32.78 & 17.57 & 27.72 & 30.42 & 31.87 & 23.84 & 34.88 \\
    & MV & 32.14 & 32.40 & 08.90 & 15.81 & 30.23 & 31.10 & 16.14 & 25.39 & 28.56 & 30.32 & 21.88 & 31.85 \\
    & B & 37.85 & 37.71 & 11.38 & 20.03 & 35.90 & 35.87 & 20.74 & 32.48 & 33.65 & 34.27 & 28.15 & 41.15 \\
    Drop & VR & 47.08 & 48.53 & 14.39 & 27.37 & 41.44 & 42.34 & 24.48 & 40.77 & 38.36 & 38.68 & 31.94 & 48.42 \\
    & PM & 50.04 & 52.30 & 14.62 & 28.21 & 44.52 & 46.35 & 26.12 & 43.17 & 41.48 & 42.47 & 36.28 & 53.25 \\
    & E & 43.07 & 45.14 & 12.06 & 23.97 & 37.85 & 39.12 & 20.53 & 35.48 & 34.68 & 35.35 & 26.65 & 41.70 \\
    & AV & 32.02 & 33.24 & 08.56 & 15.45 & 30.49 & 32.83 & 15.41 & 24.45 & 29.64 & 32.45 & 21.13 & 30.64 \\
    & MV & 30.58 & 31.90 & 07.99 & 14.26 & 29.13 & 31.69 & 14.23 & 22.27 & 28.17 & 31.49 & 19.49 & 27.95 \\
    & B & 34.35 & 35.55 & 09.68 & 17.86 & 33.17 & 34.43 & 18.12 & 29.11 & 32.11 & 33.38 & 25.36 & 37.35 \\
    \midrule
    \multicolumn{14}{c}{SegFormer B-5} \\
    Base & VR & 55.19 & 55.90 & 16.38 & 29.25 & 48.88 & 49.68 & 29.08 & 46.12 & 44.74 & 45.38 & 39.35 & 57.38 \\
    & PM & 53.48 & 54.15 & 15.97 & 28.81 & 48.18 & 48.93 & 28.83 & 45.86 & 44.27 & 44.92 & 39.22 & 57.25 \\
    & E & 54.69 & 55.50 & 16.07 & 28.51 & 48.63 & 49.74 & 28.67 & 45.76 & 44.33 & 45.40 & 39.16 & 57.46 \\
    Noise & VR & 55.21 & 55.90 & 16.39 & 29.24 & 48.90 & 49.68 & 29.10 & 46.14 & 44.73 & 45.37 & 39.34 & 57.38 \\
    & PM & 53.50 & 54.13 & 15.97 & 28.81 & 48.19 & 48.93 & 28.82 & 45.87 & 44.26 & 44.89 & 39.19 & 57.25 \\
    & E & 54.61 & 55.45 & 16.02 & 28.48 & 48.60 & 49.71 & 28.66 & 45.74 & 44.29 & 45.35 & 39.13 & 57.45 \\
    & AV & 49.94 & 50.55 & 13.87 & 24.11 & 48.42 & 48.36 & 24.92 & 38.74 & 47.80 & 47.03 & 33.24 & 47.63 \\
    & MV & 49.83 & 50.43 & 13.79 & 23.94 & 48.28 & 48.24 & 24.78 & 38.57 & 47.61 & 46.92 & 33.19 & 47.52 \\
    & B & 48.63 & 49.49 & 13.84 & 23.18 & 45.82 & 46.51 & 25.20 & 38.80 & 44.00 & 44.23 & 34.62 & 49.73 \\
    Scale & VR & \textbf{57.22} & \textbf{58.41} & 16.86 & 29.97 & 50.55 & 51.87 & 30.04 & 47.26 & 45.83 & 47.10 & \textbf{40.67} & \textbf{58.91} \\
    & PM & 54.98 & 56.10 & 16.23 & 29.29 & 49.43 & 50.66 & 29.52 & 46.70 & 45.15 & 46.38 & 40.34 & 58.50 \\
    & E & 55.19 & 56.35 & 16.16 & 28.46 & 49.43 & 50.91 & 29.07 & 46.08 & 45.05 & 46.61 & 39.71 & 58.12 \\
    & AV & 44.61 & 46.20 & 13.17 & 22.04 & 43.34 & 44.56 & 25.23 & 38.40 & 42.16 & 42.98 & 35.75 & 50.37 \\
    & MV & 44.17 & 45.76 & 13.04 & 21.84 & 42.86 & 44.17 & 24.99 & 38.12 & 41.82 & 42.68 & 35.53 & 50.13 \\
    & B & 45.28 & 47.00 & 13.38 & 21.49 & 42.71 & 44.29 & 25.22 & 37.42 & 40.59 & 41.95 & 35.64 & 49.61 \\
    Drop & VR & 55.26 & 55.97 & 16.40 & 29.26 & 48.92 & 49.72 & 29.12 & 46.18 & 44.71 & 45.40 & 39.38 & 57.45 \\
    & PM & 53.53 & 54.18 & 15.98 & 28.81 & 48.20 & 48.95 & 28.85 & 45.89 & 44.27 & 44.93 & 39.23 & 57.30 \\
    & E & 54.62 & 55.46 & 16.06 & 28.46 & 48.65 & 49.78 & 28.68 & 45.75 & 44.32 & 45.42 & 39.14 & 57.47 \\
    & AV & 51.84 & 52.71 & 15.04 & 26.50 & 48.66 & 49.12 & 27.29 & 43.02 & 46.72 & 46.54 & 37.01 & 53.72 \\
    & MV & 51.47 & 52.41 & 14.92 & 26.34 & 48.43 & 48.90 & 27.09 & 42.82 & 46.49 & 46.39 & 36.88 & 53.62 \\
    & B & 51.86 & 52.96 & 15.10 & 25.75 & 47.11 & 48.29 & 27.14 & 42.33 & 43.76 & 44.77 & 37.09 & 53.92 \\
    \bottomrule
    \caption{Micro and macro-averaged (\(\mu\) and M, respectively) Precision and Recall for thresholding based on a maximum fraction of pixels allowed on the ADE20K dataset.}
    \label{tab:ade20k_max_pixel}
\end{longtable}

Considering all these results, we would suggest using either Base or Scale to gauge the trustworthiness of a network. Base is computationally much cheaper, but Scale generally provides better results. As for the uncertainty metric, it is best to use the simpler metrics like Variation Ratio, Probability Margin and Entropy since they seem to provide the best results in most cases. It also seems that for in-domain data, thresholding based on a maximum fraction of pixels is better than using the largest difference.

\subsection*{Dark Zurich}
We now provide all results on the Dark Zurich dataset. Since Dark Zurich was used to test how well the metrics work when there is a shift in the input domain, all networks perform poorly on the dataset. Table~\ref{tab:dz_perf} shows the base performance of the models on the dataset.

DRN performs very poorly, while OneFormer and SegFormer provide much more respectable results. Overall, the results are quite poor and show that, in general, these networks do not handle large changes in input domain well.

\begin{table}[h!]
    \centering
    \begin{tabular}{lcccccc}
        \toprule
        Model & \multicolumn{2}{c}{mIoU \(\uparrow\)} & \multicolumn{2}{c}{ECE \(\downarrow\)} & \multicolumn{2}{c}{BS \(\downarrow\)} \\
        & \(\mu\) Avg. & M Avg. & \(\mu\) Avg. & M Avg. & \(\mu\) Avg. & M Avg. \\
        \midrule
        DRN D-22 & 0.0710 & 0.0737 & 0.3067 & 0.5142 & 1.1303 & 1.1361 \\
        DRN D-105 & 0.0951 & 0.0990 & 0.2877 & 0.4173 & 0.9259 & 0.9315 \\
        OneFormer ConvNeXt-L & \textbf{0.3994} & 0.3752 & 0.5964 & 0.5755 & 0.8340 & 0.8345 \\
        OneFormer Swin-L & \textbf{0.3994} & \textbf{0.3949} & 0.5988 & 0.6678 & 0.8259 & 0.8263 \\
        SegFormer B-5 & 0.3128 & 0.2686 & \textbf{0.2212} & \textbf{0.2715} & \textbf{0.6053} & \textbf{0.6063} \\
        \bottomrule
    \end{tabular}
    \caption{Micro and macro-averaged (\(\mu\) and M, respectively) Mean Intersection Over Union (mIoU), Expected Calibration Error (ECE) and Brier Score (BS) on the Dark Zurich dataset.}
    \label{tab:dz_perf}
\end{table}

Table~\ref{tab:dark_zurich_auroc_d} shows the AUROC and the performance of thresholding based on the largest difference, and Table~\ref{tab:dark_zurich_max_pixel} shows the results when we use a maximum fraction of pixels as the thresholding metric.

Even with such poor performance, we observe that the largest difference thresholding is able to achieve AUROC of 0.7 or greater. Precision suffers a bit in this scenario, but Recall is excellent, and while a large fraction of pixels are selected, it is often correct since most pixels are actually misclassified. Scale with Probability Margin as the uncertainty metric often performs the best, similar to earlier experiments.

On the other hand, thresholding based on a maximum fraction of pixels is less useful in this case, with Recall being much lower than the largest difference method. This is because, as explained earlier, most pixels are actually misclassified and restricting it to a maximum of 15\% actually hurts performance.

\begin{longtable}{llcccccccc}
    \toprule
    Scenario & Uncertainty & \multicolumn{2}{c}{AUROC \(\uparrow\)} & \multicolumn{6}{c}{Largest Difference Thresholding} \\
    & Metric & & & \multicolumn{2}{c}{Precision \(\uparrow\)} & \multicolumn{2}{c}{Recall \(\uparrow\)} & \multicolumn{2}{c}{Pixel \%} \\
    & & \(\mu\) Avg. & M Avg. & \(\mu\) Avg. & M Avg. & \(\mu\) Avg. & M Avg. & \(\mu\) Avg. & M Avg. \\
    \midrule
    \multicolumn{10}{c}{DRN D-22} \\
    Base & Var. Ratio & 0.6060 & 0.6061 & 70.14 & 70.63 & 74.73 & 75.07 & 70.05 & 70.01 \\
    & Prob. Margin & 0.6056 & 0.6067 & 69.67 & 70.18 & 77.53 & 77.85 & 73.16 & 73.11 \\
    & Entropy & 0.6054 & 0.6062 & 68.92 & 69.46 & 80.93 & 81.24 & 77.21 & 77.15 \\
    Noise & Var. Ratio & 0.6027 & 0.6038 & 69.96 & 70.49 & 74.81 & 75.17 & 70.31 & 70.26 \\
    & Prob. Margin & 0.6021 & 0.6041 & 69.50 & 70.05 & 77.57 & 77.91 & 73.39 & 73.33 \\
    & Entropy & 0.6024 & 0.6038 & 68.77 & 69.35 & 80.95 & 81.28 & 77.39 & 77.32 \\
    & Avg. Var. & 0.5892 & 0.5918 & \textbf{75.12} & \textbf{75.56} & 46.70 & 46.99 & 40.87 & 40.89 \\
    & Max. Var. & 0.5898 & 0.5922 & 75.04 & 75.47 & 47.31 & 47.60 & 41.46 & 41.47 \\
    & BALD & 0.6113 & 0.6185 & 73.39 & 74.13 & 63.25 & 63.62 & 56.66 & 56.62 \\
    Scale & Var. Ratio & 0.6835 & 0.6853 & 69.46 & 70.14 & 94.25 & 94.42 & 89.22 & 89.21 \\
    & Prob. Margin & \textbf{0.6849} & \textbf{0.6864} & 68.47 & 69.26 & \textbf{95.12} & \textbf{95.40} & 91.34 & 91.31 \\
    & Entropy & 0.6680 & 0.6714 & 71.92 & 72.34 & 79.24 & 79.24 & 72.45 & 72.67 \\
    & Avg. Var. & 0.6769 & 0.6771 & 71.45 & 71.78 & 91.80 & 91.93 & 84.48 & 84.48 \\
    & Max. Var. & 0.6763 & 0.6766 & 71.16 & 71.51 & 92.49 & 92.61 & 85.46 & 85.46 \\
    & BALD & 0.6808 & 0.6809 & 69.02 & 69.77 & 93.96 & 94.32 & 89.51 & 89.41 \\
    \midrule
    \multicolumn{10}{c}{DRN D-105} \\
    Base & Var. Ratio & 0.6909 & 0.6722 & 65.35 & 64.55 & 77.75 & 77.90 & 64.00 & 64.30 \\
    & Prob. Margin & 0.6944 & 0.6749 & 64.45 & 63.68 & 80.71 & 80.82 & 67.35 & 67.66 \\
    & Entropy & 0.6954 & 0.6756 & 63.41 & 62.68 & 83.74 & 83.88 & 71.03 & 71.35 \\
    Noise & Var. Ratio & 0.6951 & 0.6765 & 65.42 & 64.54 & 78.16 & 78.25 & 64.26 & 64.55 \\
    & Prob. Margin & 0.6986 & 0.6793 & 64.50 & 63.65 & 81.08 & 81.13 & 67.61 & 67.91 \\
    & Entropy & 0.6994 & 0.6795 & 63.44 & 62.66 & 84.12 & 84.18 & 71.31 & 71.62 \\
    & Avg. Var. & 0.6597 & 0.6499 & \textbf{72.46} & \textbf{71.56} & 53.21 & 53.57 & 39.50 & 39.70 \\
    & Max. Var. & 0.6594 & 0.6496 & 72.30 & 71.41 & 53.50 & 53.90 & 39.80 & 40.00 \\
    & BALD & 0.6951 & 0.6820 & 68.93 & 68.17 & 69.57 & 69.91 & 54.28 & 54.56 \\
    Scale & Var. Ratio & 0.7514 & 0.7291 & 61.32 & 60.82 & 91.13 & 91.52 & 79.94 & 80.29 \\
    & Prob. Margin & \textbf{0.7520} & \textbf{0.7295} & 60.48 & 60.08 & \textbf{91.99} & \textbf{92.49} & 81.80 & 82.15 \\
    & Entropy & 0.7410 & 0.7184 & 59.96 & 60.21 & 88.06 & 89.04 & 78.99 & 79.12 \\
    & Avg. Var. & 0.7369 & 0.7168 & 65.89 & 64.95 & 83.86 & 83.77 & 68.46 & 68.80 \\
    & Max. Var. & 0.7368 & 0.7149 & 65.51 & 64.57 & 84.64 & 84.53 & 69.49 & 69.83 \\
    & BALD & 0.7433 & 0.7225 & 62.10 & 61.33 & 90.92 & 90.90 & 78.75 & 79.09 \\
    \midrule
    \multicolumn{10}{c}{OneFormer ConvNeXt-L} \\
    Base & Var. Ratio & 0.5855 & 0.6168 & 38.21 & 39.15 & 60.89 & 64.90 & 46.06 & 46.13 \\
    & Prob. Margin & 0.6221 & 0.6604 & 40.17 & 41.14 & 66.90 & 70.26 & 48.14 & 48.38 \\
    & Entropy & 0.5667 & 0.5949 & 38.41 & 39.55 & 56.71 & 60.16 & 42.67 & 42.79 \\
    Noise & Var. Ratio & 0.5810 & 0.6297 & 41.12 & 41.79 & 62.40 & 64.63 & 43.86 & 44.11 \\
    & Prob. Margin & 0.6235 & 0.6731 & 44.26 & 44.16 & 70.58 & 71.58 & 46.09 & 46.43 \\
    & Entropy & 0.5641 & 0.6036 & 41.33 & 42.15 & 56.58 & 58.65 & 39.57 & 39.84 \\
    & Avg. Var. & 0.6185 & 0.5884 & 45.33 & 43.47 & 49.41 & 46.04 & 31.51 & 31.94 \\
    & Max. Var. & 0.6061 & 0.5770 & 44.69 & 42.99 & 46.14 & 42.97 & 29.84 & 30.22 \\
    & BALD & 0.6408 & 0.6098 & 46.09 & 44.01 & 54.68 & 51.24 & 34.29 & 34.82 \\
    Scale & Var. Ratio & 0.6645 & 0.7152 & 42.57 & 43.90 & 77.82 & 80.84 & 52.83 & 52.83 \\
    & Prob. Margin & 0.7136 & 0.7483 & 42.30 & 42.50 & \textbf{86.97} & \textbf{88.44} & 59.42 & 59.58 \\
    & Entropy & 0.6377 & 0.7012 & 42.47 & 43.69 & 73.21 & 77.11 & 49.82 & 49.79 \\
    & Avg. Var. & 0.7535 & 0.7529 & 47.01 & 47.51 & 79.58 & 79.77 & 48.93 & 49.27 \\
    & Max. Var. & 0.7375 & 0.7389 & \textbf{46.93} & 47.73 & 75.94 & 76.41 & 46.78 & 47.11 \\
    & BALD & \textbf{0.7782} & \textbf{0.7702} & 46.55 & \textbf{48.37} & 84.16 & 83.93 & 52.26 & 52.63 \\
    Drop & Var. Ratio & 0.5911 & 0.6229 & 38.94 & 39.56 & 62.37 & 65.38 & 46.30 & 46.49 \\
    & Prob. Margin & 0.6249 & 0.6608 & 40.57 & 41.36 & 70.08 & 73.29 & 49.93 & 50.09 \\
    & Entropy & 0.5755 & 0.6089 & 39.10 & 40.23 & 57.44 & 61.24 & 42.46 & 42.61 \\
    & Avg. Var. & 0.6007 & 0.5568 & 41.51 & 42.21 & 50.29 & 46.64 & 35.02 & 35.36 \\
    & Max. Var. & 0.5884 & 0.5424 & 39.94 & 40.71 & 47.21 & 42.98 & 34.16 & 34.50 \\
    & BALD & 0.6246 & 0.5831 & 43.23 & 43.76 & 61.41 & 57.44 & 41.06 & 41.61 \\
    \midrule
    \multicolumn{10}{c}{OneFormer Swin-L} \\
    Base & Var. Ratio & 0.6779 & 0.7245 & 30.32 & 31.88 & 76.77 & 79.86 & 51.50 & 51.69 \\
    & Prob. Margin & 0.7280 & 0.7592 & 30.33 & 31.68 & 82.61 & 85.43 & 55.40 & 55.62 \\
    & Entropy & 0.6555 & 0.7020 & 30.87 & 31.99 & 72.48 & 75.92 & 47.75 & 48.00 \\
    Noise & Var. Ratio & 0.6483 & 0.6946 & 29.73 & 31.70 & 75.11 & 78.96 & 51.38 & 51.44 \\
    & Prob. Margin & 0.6926 & 0.7253 & 30.44 & 31.83 & 82.02 & 84.50 & 54.80 & 54.92 \\
    & Entropy & 0.6293 & 0.6770 & 29.85 & 31.36 & 70.65 & 74.42 & 48.13 & 48.15 \\
    & Avg. Var. & 0.6410 & 0.6416 & 33.52 & 35.01 & 56.22 & 56.33 & 34.11 & 34.37 \\
    & Max. Var. & 0.6307 & 0.6288 & 33.44 & 35.05 & 53.37 & 53.36 & 32.46 & 32.70 \\
    & BALD & 0.6670 & 0.6635 & \textbf{34.19} & 35.52 & 63.24 & 62.72 & 37.62 & 37.98 \\
    Scale & Var. Ratio & 0.7211 & 0.7414 & 29.35 & 30.90 & 87.33 & 89.37 & 60.52 & 60.78 \\
    & Prob. Margin & \textbf{0.7577} & \textbf{0.7662} & 30.10 & 31.51 & \textbf{93.48} & \textbf{94.12} & 63.17 & 63.49 \\
    & Entropy & 0.7008 & 0.7259 & 29.76 & 31.32 & 80.13 & 82.50 & 54.77 & 54.87 \\
    & Avg. Var. & 0.6925 & 0.6958 & 29.62 & 31.29 & 82.56 & 83.32 & 56.69 & 57.03 \\
    & Max. Var. & 0.6752 & 0.6806 & 29.78 & 31.03 & 78.57 & 79.66 & 53.65 & 53.92 \\
    & BALD & 0.7250 & 0.7221 & 30.32 & 31.52 & 85.74 & 86.40 & 57.50 & 57.76 \\
    Drop & Var. Ratio & 0.6956 & 0.7290 & 29.63 & 31.56 & 79.85 & 83.26 & 54.81 & 54.88 \\
    & Prob. Margin & 0.7379 & 0.7591 & 29.56 & 31.19 & 86.79 & 88.97 & 59.71 & 59.94 \\
    & Entropy & 0.6736 & 0.7098 & 30.07 & 31.84 & 74.97 & 78.39 & 50.70 & 50.75 \\
    & Avg. Var. & 0.6046 & 0.6040 & 31.75 & 35.61 & 49.89 & 50.91 & 31.95 & 31.96 \\
    & Max. Var. & 0.5834 & 0.5822 & 30.39 & 34.46 & 44.06 & 45.19 & 29.49 & 29.42 \\
    & BALD & 0.6471 & 0.6387 & 33.59 & \textbf{36.63} & 59.41 & 59.76 & 35.97 & 36.06 \\
    \midrule
    \multicolumn{10}{c}{SegFormer B-5} \\
    Base & Var. Ratio & 0.7999 & 0.7978 & 61.71 & 60.47 & 83.55 & 83.24 & 48.95 & 49.18 \\
    & Prob. Margin & 0.8053 & 0.8033 & 60.56 & 59.34 & 85.81 & 85.53 & 51.23 & 51.46 \\
    & Entropy & 0.8155 & 0.8156 & 58.55 & 57.43 & 89.42 & 89.24 & 55.22 & 55.47 \\
    Noise & Var. Ratio & 0.7949 & 0.7903 & 62.18 & 60.54 & 82.73 & 82.46 & 48.10 & 48.26 \\
    & Prob. Margin & 0.8012 & 0.7966 & 61.02 & 59.40 & 85.26 & 84.98 & 50.52 & 50.69 \\
    & Entropy & 0.8130 & 0.8092 & 58.85 & 57.44 & 89.09 & 88.89 & 54.74 & 54.93 \\
    & Avg. Var. & 0.6740 & 0.6721 & 67.72 & 67.33 & 47.74 & 47.73 & 25.48 & 25.56 \\
    & Max. Var. & 0.6743 & 0.6724 & 67.80 & 67.37 & 47.72 & 47.71 & 25.45 & 25.51 \\
    & BALD & 0.7243 & 0.7217 & 65.13 & 64.02 & 63.62 & 63.44 & 35.31 & 35.39 \\
    Scale & Var. Ratio & 0.8474 & 0.8490 & 56.87 & 55.85 & 94.71 & 94.56 & 60.21 & 60.54 \\
    & Prob. Margin & 0.8483 & 0.8491 & 55.76 & 54.76 & 95.73 & 95.62 & 62.06 & 62.38 \\
    & Entropy & \textbf{0.8519} & \textbf{0.8566} & 53.81 & 52.90 & \textbf{97.22} & \textbf{97.22} & 65.32 & 65.63 \\
    & Avg. Var. & 0.8064 & 0.8045 & 62.63 & 61.51 & 84.98 & 84.79 & 49.06 & 49.35 \\
    & Max. Var. & 0.8088 & 0.8060 & 62.39 & 61.25 & 85.59 & 85.37 & 49.60 & 49.90 \\
    & BALD & 0.8247 & 0.8249 & 57.76 & 56.68 & 93.09 & 92.86 & 58.27 & 58.59 \\
    Drop & Var. Ratio & 0.8013 & 0.7990 & 61.58 & 60.33 & 83.92 & 83.58 & 49.27 & 49.50 \\
    & Prob. Margin & 0.8065 & 0.8043 & 60.42 & 59.19 & 86.15 & 85.83 & 51.55 & 51.78 \\
    & Entropy & 0.8165 & 0.8163 & 58.37 & 57.24 & 89.68 & 89.48 & 55.55 & 55.80 \\
    & Avg. Var. & 0.6980 & 0.6990 & 69.68 & 68.43 & 51.47 & 51.90 & 26.70 & 26.78 \\
    & Max. Var. & 0.6979 & 0.6976 & \textbf{70.03} & \textbf{68.67} & 51.20 & 51.42 & 26.43 & 26.50 \\
    & BALD & 0.7539 & 0.7532 & 65.18 & 64.02 & 69.57 & 69.40 & 38.59 & 38.75 \\
    \bottomrule
    \caption{Micro and macro-averaged (\(\mu\) and M, respectively) Area Under Receiver Operating Characteristic (AUROC) and Precision, Recall and the fraction of pixels selected for largest difference thresholding on the Dark Zurich dataset.}
    \label{tab:dark_zurich_auroc_d}
\end{longtable}

\begin{longtable}{llcccccccccccc}
    \toprule
    Scenario & Unc. & \multicolumn{4}{c}{Max 5\% pixels} & \multicolumn{4}{c}{Max 10\% pixels} & \multicolumn{4}{c}{Max 15\% pixels} \\
    & Metric & \multicolumn{2}{c}{Precision \(\uparrow\)} & \multicolumn{2}{c}{Recall \(\uparrow\)} & \multicolumn{2}{c}{Precision \(\uparrow\)} & \multicolumn{2}{c}{Recall \(\uparrow\)} & \multicolumn{2}{c}{Precision \(\uparrow\)} & \multicolumn{2}{c}{Recall \(\uparrow\)} \\
    & & \(\mu\) & M & \(\mu\) & M & \(\mu\) & M & \(\mu\) & M & \(\mu\) & M & \(\mu\) & M \\
    \midrule
    \multicolumn{14}{c}{DRN D-22} \\
    Base & VR & 80.81 & 80.86 & 05.86 & 05.94 & 79.04 & 79.14 & 11.65 & 11.79 & 77.96 & 78.11 & 17.41 & 17.61 \\
    & PM & 78.91 & 79.08 & 05.74 & 05.81 & 78.24 & 78.41 & 11.63 & 11.76 & 77.62 & 77.77 & 17.41 & 17.59 \\
    & E & 81.07 & 81.12 & 05.92 & 06.01 & 78.88 & 78.96 & 11.57 & 11.74 & 77.51 & 77.58 & 17.03 & 17.23 \\
    Noise & VR & 79.99 & 80.04 & 05.75 & 05.83 & 78.31 & 78.44 & 11.49 & 11.63 & 77.40 & 77.56 & 17.24 & 17.43 \\
    & PM & 77.93 & 78.09 & 05.63 & 05.69 & 77.36 & 77.56 & 11.45 & 11.58 & 76.98 & 77.16 & 17.27 & 17.45 \\
    & E & 80.71 & 80.78 & 05.92 & 06.01 & 78.48 & 78.56 & 11.59 & 11.74 & 77.12 & 77.22 & 16.98 & 17.19 \\
    & AV & 77.78 & 77.89 & 05.65 & 05.69 & 77.46 & 77.62 & 11.13 & 11.20 & 77.28 & 77.46 & 16.58 & 16.68 \\
    & MV & 77.75 & 77.81 & 05.64 & 05.68 & 77.27 & 77.53 & 11.18 & 11.25 & 77.17 & 77.43 & 16.57 & 16.69 \\
    & B & 77.99 & 78.09 & 05.65 & 05.70 & 77.99 & 78.30 & 11.22 & 11.32 & 77.93 & 78.23 & 16.78 & 16.92 \\
    Scale & VR & 83.99 & 84.07 & 06.03 & 06.13 & 82.50 & 82.54 & 11.97 & 12.16 & 81.54 & 81.62 & 17.83 & 18.10 \\
    & PM & 82.04 & 82.13 & 05.49 & 05.56 & 81.63 & 81.73 & 11.75 & 11.92 & 81.09 & 81.22 & \textbf{17.87} & \textbf{18.13} \\
    & E & \textbf{84.60} & \textbf{84.61} & 06.08 & 06.18 & 81.69 & 81.80 & 11.88 & 12.08 & 80.24 & 80.34 & 17.67 & 17.94 \\
    & AV & 76.09 & 76.20 & 05.45 & 05.50 & 76.80 & 76.91 & 11.22 & 11.33 & 77.37 & 77.53 & 16.96 & 17.13 \\
    & MV & 75.39 & 75.64 & 05.48 & 05.52 & 76.65 & 76.87 & 11.27 & 11.37 & 77.42 & 77.68 & 17.04 & 17.21 \\
    & B & 81.40 & 81.51 & 05.87 & 05.94 & 79.81 & 79.95 & 11.60 & 11.73 & 79.19 & 79.32 & 17.43 & 17.61 \\
    \midrule
    \multicolumn{14}{c}{DRN D-105} \\
    Base & VR & 79.49 & 79.57 & 06.99 & 07.54 & 77.15 & 77.21 & 13.95 & 14.92 & 75.53 & 75.57 & 20.69 & 21.99 \\
    & PM & 78.21 & 78.21 & 07.00 & 07.49 & 76.79 & 76.80 & 14.02 & 14.95 & 75.38 & 75.42 & 20.73 & 22.02 \\
    & E & 78.96 & 79.07 & 07.03 & 07.59 & 76.02 & 76.04 & 13.50 & 14.44 & 74.43 & 74.49 & 19.73 & 21.04 \\
    Noise & VR & 79.45 & 79.67 & 07.01 & 07.55 & 77.69 & 77.87 & 14.08 & 15.07 & 76.24 & 76.40 & 20.83 & 22.21 \\
    & PM & 78.74 & 78.86 & 07.01 & 07.54 & 77.46 & 77.60 & 14.10 & 15.06 & 76.13 & 76.30 & 20.91 & 22.28 \\
    & E & 78.57 & 78.76 & 06.98 & 07.54 & 76.16 & 76.30 & 13.49 & 14.45 & 74.78 & 74.96 & 19.91 & 21.27 \\
    & AV & 82.54 & 82.55 & 07.32 & 07.80 & 80.39 & 80.35 & 14.29 & 15.18 & 78.53 & 78.51 & 20.94 & 22.18 \\
    & MV & 82.35 & 82.37 & 07.34 & 07.82 & 80.21 & 80.12 & 14.26 & 15.14 & 78.36 & 78.36 & 20.82 & 22.01 \\
    & B & 83.58 & 83.48 & 07.41 & 07.91 & 81.03 & 81.08 & 14.36 & 15.31 & 79.27 & 79.24 & 21.10 & 22.38 \\
    Scale & VR & \textbf{85.07} & \textbf{85.04} & 07.44 & 08.09 & 83.39 & 83.39 & 14.81 & 15.95 & 81.64 & 81.69 & 21.89 & 23.48 \\
    & PM & 83.96 & 83.95 & 07.22 & 07.84 & 82.98 & 82.98 & 14.79 & 15.93 & 81.75 & 81.75 & \textbf{22.28} & \textbf{23.85} \\
    & E & 81.47 & 81.53 & 07.09 & 07.70 & 78.75 & 78.91 & 13.96 & 15.06 & 76.88 & 77.00 & 20.53 & 21.96 \\
    & AV & 80.17 & 80.07 & 07.04 & 07.52 & 79.30 & 79.18 & 14.21 & 15.06 & 78.35 & 78.17 & 21.11 & 22.31 \\
    & MV & 79.46 & 79.40 & 07.02 & 07.49 & 78.23 & 78.17 & 14.07 & 14.93 & 77.44 & 77.35 & 20.89 & 22.10 \\
    & B & 83.78 & 83.56 & 07.35 & 07.95 & 81.43 & 81.31 & 14.51 & 15.53 & 79.77 & 79.67 & 21.41 & 22.75 \\
    \midrule
    \multicolumn{14}{c}{OneFormer ConvNeXt-L} \\
    Base & VR & 55.89 & 55.76 & 08.77 & 09.96 & 50.95 & 51.16 & 16.68 & 18.75 & 49.33 & 49.48 & 24.12 & 27.14 \\
    & PM & 56.90 & 57.26 & 08.92 & 10.33 & 52.74 & 52.69 & 17.00 & 19.15 & 50.50 & 50.72 & 25.34 & 28.66 \\
    & E & 53.66 & 57.98 & 07.10 & 07.95 & 50.51 & 52.77 & 15.43 & 17.23 & 48.45 & 48.76 & 23.19 & 26.27 \\
    Noise & VR & 50.88 & 50.21 & 08.01 & 09.27 & 48.60 & 48.81 & 15.88 & 18.17 & 48.02 & 48.27 & 23.52 & 26.45 \\
    & PM & 53.23 & 53.87 & 08.42 & 09.64 & 51.40 & 51.91 & 16.85 & 19.14 & 51.08 & 51.35 & 25.31 & 28.26 \\
    & E & 49.02 & 51.82 & 06.52 & 07.39 & 49.29 & 48.66 & 15.02 & 17.19 & 48.25 & 48.43 & 22.81 & 25.61 \\
    & AV & 46.85 & 45.04 & 06.36 & 07.00 & 44.08 & 43.39 & 12.61 & 13.80 & 43.62 & 43.65 & 17.48 & 19.20 \\
    & MV & 45.42 & 43.87 & 06.00 & 06.62 & 43.74 & 42.74 & 11.67 & 12.69 & 42.86 & 42.59 & 16.37 & 17.96 \\
    & B & 48.90 & 47.01 & 07.34 & 08.14 & 46.53 & 45.87 & 13.75 & 15.10 & 46.43 & 46.20 & 20.19 & 22.12 \\
    Scale & VR & 52.04 & 51.60 & 07.88 & 08.93 & 48.94 & 49.20 & 15.42 & 17.39 & 50.30 & 50.41 & 24.27 & 27.58 \\
    & PM & 46.41 & 47.38 & 06.59 & 07.49 & 48.77 & 48.99 & 15.34 & 17.39 & 50.86 & 50.86 & 24.99 & 28.57 \\
    & E & 51.66 & 50.42 & 06.92 & 08.08 & 51.24 & 51.17 & 14.99 & 17.19 & 51.33 & 51.43 & 23.35 & 26.61 \\
    & AV & 52.01 & 51.73 & 07.79 & 08.84 & 56.87 & 55.85 & 16.98 & 19.21 & 57.48 & 56.90 & 25.91 & 29.41 \\
    & MV & 51.34 & 49.82 & 06.96 & 07.83 & 55.74 & 53.54 & 14.94 & 17.66 & 57.64 & 55.91 & 24.48 & 28.40 \\
    & B & \textbf{57.51} & \textbf{57.99} & 09.13 & 10.25 & 58.43 & 58.94 & 18.77 & 21.02 & 58.65 & 59.66 & \textbf{28.41} & \textbf{31.66} \\
    Drop & VR & 51.98 & 51.69 & 08.09 & 09.25 & 49.13 & 49.10 & 16.15 & 18.31 & 48.15 & 48.37 & 23.93 & 26.97 \\
    & PM & 51.11 & 51.05 & 07.99 & 09.14 & 49.69 & 49.70 & 16.39 & 18.56 & 49.00 & 49.12 & 24.54 & 27.56 \\
    & E & 52.52 & 50.47 & 06.85 & 07.76 & 50.17 & 48.08 & 15.28 & 17.34 & 47.78 & 47.82 & 22.84 & 25.84 \\
    & AV & 38.57 & 37.21 & 04.57 & 05.25 & 39.58 & 38.91 & 09.37 & 10.53 & 37.29 & 36.68 & 13.84 & 15.12 \\
    & MV & 38.98 & 36.21 & 04.45 & 05.15 & 37.04 & 34.95 & 08.49 & 09.52 & 36.13 & 35.49 & 13.12 & 14.32 \\
    & B & 39.47 & 38.96 & 04.86 & 05.60 & 39.04 & 40.32 & 10.22 & 11.43 & 39.46 & 39.81 & 15.44 & 17.16 \\
    \midrule
    \multicolumn{14}{c}{OneFormer Swin-L} \\
    Base & VR & 56.31 & 57.18 & 12.60 & 13.72 & 52.36 & 52.90 & 24.12 & 26.45 & 48.69 & 49.42 & 33.07 & 36.21 \\
    & PM & \textbf{56.72} & \textbf{57.57} & 11.37 & 13.35 & 55.15 & 55.24 & 25.23 & 28.12 & 51.94 & 52.22 & \textbf{36.61} & \textbf{40.07} \\
    & E & 52.69 & 55.74 & 09.64 & 11.01 & 50.94 & 53.44 & 21.54 & 23.85 & 47.18 & 49.77 & 29.67 & 32.64 \\
    Noise & VR & 47.88 & 49.03 & 10.66 & 11.51 & 45.92 & 46.34 & 21.15 & 22.78 & 42.87 & 43.36 & 29.80 & 31.88 \\
    & PM & 47.28 & 49.55 & 10.06 & 11.08 & 46.03 & 46.35 & 21.56 & 23.40 & 44.13 & 44.37 & 31.53 & 33.76 \\
    & E & 47.89 & 50.21 & 09.52 & 10.16 & 45.17 & 47.59 & 19.38 & 21.13 & 42.15 & 42.67 & 28.22 & 30.05 \\
    & AV & 40.68 & 39.68 & 08.21 & 08.92 & 37.55 & 38.22 & 15.33 & 16.84 & 37.95 & 38.76 & 22.46 & 24.85 \\
    & MV & 37.52 & 38.19 & 07.38 & 08.13 & 37.55 & 38.72 & 14.63 & 16.60 & 35.51 & 36.92 & 21.04 & 23.13 \\
    & B & 42.65 & 42.53 & 09.14 & 09.80 & 41.62 & 41.63 & 17.70 & 19.21 & 39.95 & 40.22 & 25.51 & 27.80 \\
    Scale & VR & 48.65 & 49.65 & 10.46 & 11.12 & 46.42 & 47.06 & 20.92 & 22.38 & 45.17 & 45.57 & 31.19 & 33.47 \\
    & PM & 43.23 & 46.15 & 08.91 & 09.53 & 44.32 & 44.66 & 20.25 & 21.83 & 45.21 & 45.28 & 31.92 & 34.26 \\
    & E & 46.59 & 48.57 & 09.34 & 10.13 & 46.59 & 48.38 & 19.30 & 21.02 & 44.81 & 45.35 & 29.41 & 31.71 \\
    & AV & 36.17 & 35.70 & 08.00 & 08.83 & 36.06 & 35.52 & 16.22 & 17.54 & 36.04 & 36.23 & 24.61 & 26.69 \\
    & MV & 33.22 & 32.79 & 07.13 & 08.06 & 34.45 & 33.58 & 14.99 & 16.49 & 35.30 & 34.60 & 23.07 & 25.30 \\
    & B & 42.63 & 42.15 & 09.55 & 09.89 & 41.67 & 41.58 & 19.41 & 20.26 & 40.87 & 41.04 & 28.37 & 29.89 \\
    Drop & VR & 53.68 & 54.48 & 11.89 & 12.78 & 50.24 & 50.64 & 23.10 & 25.08 & 47.42 & 47.96 & 33.15 & 35.84 \\
    & PM & 50.47 & 51.85 & 10.43 & 11.89 & 51.13 & 51.63 & 23.81 & 26.40 & 49.95 & 50.33 & 35.11 & 38.40 \\
    & E & 52.31 & 56.24 & 09.76 & 10.67 & 49.54 & 52.10 & 21.05 & 23.13 & 46.24 & 48.85 & 30.17 & 33.00 \\
    & AV. & 32.22 & 36.01 & 05.65 & 06.56 & 35.33 & 37.06 & 12.58 & 14.02 & 33.60 & 35.12 & 16.58 & 18.30 \\
    & MV. & 31.64 & 35.13 & 05.12 & 05.95 & 32.18 & 33.85 & 10.85 & 12.03 & 30.54 & 32.84 & 15.57 & 17.30 \\
    & B & 36.19 & 37.58 & 07.08 & 08.17 & 38.70 & 40.07 & 13.87 & 15.61 & 37.65 & 39.60 & 19.59 & 21.82 \\
    \midrule
    \multicolumn{14}{c}{SegFormer B-5} \\
    Base & VR & 72.70 & 72.67 & 09.58 & 11.22 & 71.27 & 71.21 & 19.25 & 22.15 & 69.95 & 69.87 & 28.38 & 32.22 \\
    & PM & 72.24 & 72.19 & 09.73 & 11.32 & 71.05 & 71.00 & 19.32 & 22.18 & 69.75 & 69.64 & 28.57 & 32.32 \\
    & E & 72.93 & 72.80 & 09.47 & 11.01 & 71.84 & 71.79 & 18.83 & 21.77 & 70.47 & 70.46 & 28.09 & 32.03 \\
    Noise & VR & 70.16 & 70.08 & 09.29 & 10.96 & 69.24 & 69.22 & 18.66 & 21.70 & 68.26 & 68.20 & 27.78 & 31.80 \\
    & PM & 69.36 & 69.36 & 09.30 & 10.92 & 69.00 & 68.94 & 18.82 & 21.81 & 68.08 & 68.02 & 27.87 & 31.87 \\
    & E & 71.56 & 71.48 & 09.34 & 10.99 & 70.72 & 70.54 & 18.61 & 21.78 & 68.74 & 68.77 & 27.30 & 31.54 \\
    & AV & 69.60 & 69.58 & 08.94 & 10.26 & 68.89 & 68.89 & 17.37 & 19.75 & 68.13 & 68.16 & 25.14 & 28.08 \\
    & MV & 69.83 & 69.78 & 09.03 & 10.39 & 68.77 & 68.92 & 17.49 & 20.04 & 68.30 & 68.16 & 25.17 & 28.07 \\
    & B & 68.81 & 68.80 & 08.88 & 10.21 & 67.70 & 67.64 & 16.99 & 19.23 & 66.81 & 66.66 & 24.81 & 27.69 \\
    Scale & VR & \textbf{77.93} & \textbf{77.94} & 10.22 & 11.91 & 76.02 & 76.08 & 20.21 & 23.45 & 74.41 & 74.45 & \textbf{30.01} & \textbf{34.19} \\
    & PM & 75.47 & 75.56 & 09.87 & 11.43 & 74.68 & 74.77 & 20.04 & 23.06 & 73.60 & 73.63 & 29.95 & 34.02 \\
    & E & 75.50 & 75.60 & 09.76 & 11.42 & 75.39 & 75.33 & 19.60 & 22.63 & 74.43 & 74.41 & 29.45 & 33.66 \\
    & AV. & 67.95 & 68.02 & 08.96 & 10.33 & 68.42 & 68.45 & 18.39 & 21.15 & 69.37 & 69.32 & 28.13 & 31.94 \\
    & MV. & 66.71 & 66.83 & 08.83 & 10.13 & 68.17 & 68.16 & 18.40 & 21.12 & 69.45 & 69.33 & 28.25 & 31.95 \\
    & B & 69.76 & 69.68 & 09.21 & 10.68 & 68.56 & 68.62 & 18.32 & 21.08 & 68.93 & 68.88 & 27.78 & 31.49 \\
    Drop & VR & 72.80 & 72.78 & 09.66 & 11.28 & 71.46 & 71.37 & 19.38 & 22.24 & 70.06 & 69.97 & 28.56 & 32.37 \\
    & PM & 72.48 & 72.43 & 09.75 & 11.30 & 71.25 & 71.16 & 19.40 & 22.19 & 69.84 & 69.74 & 28.59 & 32.34 \\
    & E & 72.92 & 72.78 & 09.47 & 10.98 & 71.92 & 71.81 & 18.99 & 21.88 & 70.55 & 70.56 & 27.98 & 31.86 \\
    & AV & 75.17 & 74.96 & 09.59 & 10.93 & 73.69 & 73.33 & 18.39 & 20.67 & 72.07 & 71.87 & 26.53 & 29.74 \\
    & MV & 75.22 & 74.96 & 09.58 & 10.84 & 73.62 & 73.33 & 18.30 & 20.62 & 72.26 & 72.06 & 26.56 & 29.70 \\
    & B & 74.30 & 73.99 & 09.48 & 10.74 & 72.36 & 72.05 & 18.21 & 20.52 & 70.46 & 70.54 & 26.30 & 29.47 \\
    \bottomrule
    \caption{Micro and macro-averaged (\(\mu\) and M, respectively) Precision and Recall for thresholding based on a maximum fraction of pixels allowed on the Dark Zurich dataset.}
    \label{tab:dark_zurich_max_pixel}
\end{longtable}

The results on Dark Zurich suggest much the same things as those on Cityscapes and ADE20K that Scale with simpler metrics is probably the best choice. One major difference is that if domain shift is expected, it is better to use the largest difference thresholding method than a maximum fraction of allowed pixels.

\subsection*{Experiments With Noisy Inputs}
Here, we provide the results of our experiments with noisy inputs. This experiment was only performed on the Cityscapes dataset with DRN D-22 and OneFormer ConvNeXt-L models. Table~\ref{tab:noise_perf} shows the base performances across varying noise levels.

As noise levels increase, we see a decrease in performance for both models, although, as seen earlier DRN gives much worse outputs than OneFormer. DRN's calibration also worsens with each noise level, whereas OneFormer's calibration stays mostly the same.
\begin{longtable}{lcccccc}
    \toprule
    Noise Level & \multicolumn{2}{c}{mIoU \(\uparrow\)} & \multicolumn{2}{c}{ECE \(\downarrow\)} & \multicolumn{2}{c}{BS \(\downarrow\)} \\
    & \(\mu\) Avg. & M Avg. & \(\mu\) Avg. & M Avg. & \(\mu\) Avg. & M Avg. \\
    \midrule
    \multicolumn{7}{c}{DRN D-22} \\
    None & \textbf{0.6790} & \textbf{0.5479} & 0.0380 & \textbf{0.0177} & \textbf{0.0808} & \textbf{0.0824} \\
    Std. 5 & 0.4961 & 0.4497 & 0.0897 & 0.0619 & 0.1810 & 0.1828 \\
    Std. 10 & 0.3066 & 0.2928 & \textbf{0.0288} & 0.1733 & 0.4327 & 0.4325 \\
    Std. 25 & 0.1071 & 0.1034 & 0.0700 & 0.3995 & 0.9390 & 0.9349 \\
    Std. 50 & 0.0311 & 0.0312 & 0.0917 & 0.5574 & 1.2908 & 1.2878 \\
    \midrule
    \multicolumn{7}{c}{OneFormer ConvNeXt-L} \\
    None & \textbf{0.8287} & \textbf{0.6733} & 0.7887 & 0.8383 & 0.8012 & 0.8011 \\
    Std. 5 & 0.8149 & 0.6639 & 0.7912 & 0.8365 & 0.8014 & 0.8014 \\
    Std. 10 & 0.7979 & 0.6525 & 0.7874 & 0.8325 & 0.8013 & 0.8013 \\
    Std. 25 & 0.7413 & 0.6053 & 0.7693 & 0.8176 & 0.8012 & 0.8009 \\
    Std. 50 & 0.6202 & 0.5179 & \textbf{0.7505} & \textbf{0.7770} & \textbf{0.7996} & \textbf{0.7992} \\
    \bottomrule
    \caption{Micro and macro-averaged (\(\mu\) and M, respectively) Mean Intersection Over Union (mIoU), Expected Calibration Error (ECE) and Brier Score (BS) on the Cityscapes dataset for noisy inputs.}
    \label{tab:noise_perf}
\end{longtable}

Table~\ref{tab:noise_auroc_d} shows the results of using the largest difference as a thresholding technique on noisy inputs and AUROC values. Table~\ref{tab:noise_max_pixel} shows the results of using a maximum fraction of pixels allowed as thresholding.

As seen with Dark Zurich, the largest thresholding technique becomes better as noise levels increase, with better Precision and Recall. In contrast, a maximum fraction of pixels thresholding works better at lower noise.

\begin{longtable}{llcccccccc}
    \toprule
    Noise Level & Uncertainty & \multicolumn{2}{c}{AUROC \(\uparrow\)} & \multicolumn{6}{c}{Largest Difference Thresholding} \\
    & Metric & & & \multicolumn{2}{c}{Precision \(\uparrow\)} & \multicolumn{2}{c}{Recall \(\uparrow\)} & \multicolumn{2}{c}{Pixel \%} \\
    & & \(\mu\) Avg. & M Avg. & \(\mu\) Avg. & M Avg. & \(\mu\) Avg. & M Avg. & \(\mu\) Avg. & M Avg. \\
    \midrule
    \multicolumn{10}{c}{DRN D-22} \\
    None & Var. Ratio & 0.9118 & 0.9277 & 21.10 & 20.18 & 92.13 & 94.71 & 23.29 & 23.51 \\
    & Prob. Margin & 0.9152 & 0.9302 & 19.71 & 18.84 & 93.31 & 95.62 & 25.26 & 25.48 \\
    & Entropy & \textbf{0.9205} & \textbf{0.9337} & 17.52 & 16.74 & 95.04 & \textbf{96.89} & 28.94 & 29.21 \\
    Std. 5 & Var. Ratio & 0.8413 & 0.8618 & 30.23 & 29.08 & 84.43 & 87.70 & 31.78 & 31.86 \\
\end{longtable}

\begin{longtable}{llcccccccccccc}
    \toprule
    Noise & Unc. & \multicolumn{4}{c}{Max 5\% pixels} & \multicolumn{4}{c}{Max 10\% pixels} & \multicolumn{4}{c}{Max 15\% pixels} \\
    Level & Metric & \multicolumn{2}{c}{Precision \(\uparrow\)} & \multicolumn{2}{c}{Recall \(\uparrow\)} & \multicolumn{2}{c}{Precision \(\uparrow\)} & \multicolumn{2}{c}{Recall \(\uparrow\)} & \multicolumn{2}{c}{Precision \(\uparrow\)} & \multicolumn{2}{c}{Recall \(\uparrow\)} \\
    & & \(\mu\) & M & \(\mu\) & M & \(\mu\) & M & \(\mu\) & M & \(\mu\) & M & \(\mu\) & M \\
    \midrule
    \multicolumn{14}{c}{DRN D-22} \\
    None & VR & 45.19 & 45.40 & 41.52 & 50.74 & 35.04 & 35.18 & 64.13 & 73.92 & 28.48 & 28.45 & 76.44 & 84.64 \\
    & PM & 44.66 & 44.83 & 41.35 & 50.58 & 34.77 & 34.92 & 64.13 & 73.95 & 28.14 & 28.19 & \textbf{76.51} & \textbf{84.74} \\
    & E & 44.04 & 44.23 & 39.76 & 48.74 & 34.80 & 35.03 & 63.64 & 73.61 & 28.11 & 28.23 & 76.42 & 84.73 \\
    Std. 5 & VR & 54.97 & 55.12 & 23.54 & 29.69 & 47.63 & 47.73 & 41.16 & 49.70 & 42.10 & 42.13 & 54.23 & 63.15 \\
\end{longtable}

\subsection*{Classwise Results}
Finally, we provide classwise results for some of the experiments. Table~\ref{tab:cls_city_iou} shows classwise Mean Intersection Over Union (mIoU) for all tested networks on the Cityscapes dataset, and Table~\ref{tab:cls_ade_iou} shows the same for ADE20K dataset. We observe that the best and worst classes in each dataset are consistent across models.

\begin{longtable}[!h]{lcccccccccc}
    \toprule
    Class & \multicolumn{2}{c}{DRN D-22} & \multicolumn{2}{c}{DRN D-105} & \multicolumn{2}{c}{OF ConvNeXt-L} & \multicolumn{2}{c}{OF Swin-L} & \multicolumn{2}{c}{Segformer B5} \\
    & \(\mu\) Avg. & M Avg. & \(\mu\) Avg. & M Avg. & \(\mu\) Avg. & M Avg. & \(\mu\) Avg. & M Avg. & \(\mu\) Avg. & M Avg. \\
    \midrule
    road          & \G{0.9721} & \G{0.9390} & \G{0.9815} & \G{0.9492} & \G{0.9858} & \G{0.9553} & \G{0.9847} & \G{0.9561} & \G{0.9847} & \G{0.9524} \\
sidewalk      & 0.7966                      & 0.6962                      & 0.8498                      & 0.7424                      & 0.8775                      & 0.7710                      & 0.8673                      & 0.7646                      & 0.8735                      & 0.7671                      \\
building      & \G{0.9019} & \G{0.8461} & \G{0.9249} & \G{0.8744} & \G{0.9371} & \G{0.8926} & \G{0.9401} & \G{0.8965} & \G{0.9375} & \G{0.8947} \\
wall          & \R{0.3805} & \R{0.1526} & \R{0.4826} & \R{0.2291} & \R{0.5081} & \R{0.3053} & \R{0.6663} & \R{0.2973} & \R{0.6960} & \R{0.3196} \\
fence         & \R{0.4654} & \R{0.1716} & \R{0.5877} & \R{0.2515} & \R{0.6937} & \R{0.2809} & \R{0.6951} & \R{0.2886} & \R{0.6754} & \R{0.3224} \\
pole          & 0.5877                      & 0.5258                      & 0.6632                      & 0.6062                      & \R{0.7235} & 0.6665                      & \R{0.7217} & 0.6683                      & \R{0.6964} & 0.6454                      \\
\end{longtable}

\begin{longtable}[!h]{lcccccc}
\toprule
    Class & \multicolumn{2}{c}{OF ConvNeXt-L} & \multicolumn{2}{c}{OF Swin-L} & \multicolumn{2}{c}{Segformer B5} \\
    & \(\mu\) Avg. & M Avg. & \(\mu\) Avg. & M Avg. & \(\mu\) Avg. & M Avg. \\
    \midrule
wall            & 0.8067                      & 0.6516                      & 0.8145                      & 0.6618                      & 0.7856                      & 0.6381                      \\
building        & 0.8439                      & 0.6402                      & 0.8566                      & 0.6479                      & 0.8214                      & 0.6160                      \\
sky             & \G{0.9483} & \G{0.8759} & \G{0.9506} & \G{0.8790} & \G{0.9453} & \G{0.8648} \\
floor           & 0.8459                      & 0.7415                      & 0.8456                      & 0.7434                      & 0.8220                      & 0.7262                      \\
tree            & 0.7759                      & 0.6058                      & 0.7753                      & 0.6108                      & 0.7477                      & 0.5786                      \\
ceiling         & 0.8583                      & 0.7567                      & 0.8542                      & 0.7483                      & 0.8490                      & \G{0.7379} \\
\end{longtable}

Table~\ref{tab:cls_city_auroc} and Table~\ref{tab:cls_ade_auroc} show the Area Under Receiver Operating Characteristic (AUROC) on Cityscapes and ADE20K datasets across a few scenarios. All of the results use entropy as the uncertainty metric. Here also we observe that the best and worst performing classes are generally consistent across different scenarios, especially for Cityscapes.

\begin{longtable}[!h]{lcccccccc}
\toprule
Class & \multicolumn{2}{c}{DRN D-105 Base} & \multicolumn{2}{c}{DRN D-105 Scale} & \multicolumn{2}{c}{Segformer B5 Base} & \multicolumn{2}{c}{Segformer B5 Scale} \\
    & \(\mu\) Avg. & M Avg. & \(\mu\) Avg. & M Avg. & \(\mu\) Avg. & M Avg. & \(\mu\) Avg. & M Avg. \\
\midrule
road          & \G{0.9241} & \G{0.9783} & \G{0.9615} & \G{0.9861} & 0.9118                      & \G{0.9693} & 0.9335                      & \G{0.9801} \\
sidewalk      & 0.8606                      & 0.8756                      & 0.8996                      & 0.8742                      & 0.8642                      & 0.8826                      & 0.8962                      & 0.8795                      \\
building      & \G{0.9235} & \G{0.9332} & 0.9349                      & \G{0.9346} & \G{0.9148} & \G{0.9293} & 0.9280                      & \G{0.9353} \\
wall          & \R{0.6513} & \R{0.4907} & \R{0.6980} & \R{0.4717} & \R{0.7152} & \R{0.5491} & \R{0.7167} & \R{0.5132} \\
fence         & \R{0.6946} & 0.5896                      & \R{0.7400} & 0.5685                      & \R{0.7387} & \R{0.6148} & \R{0.8026} & \R{0.5933} \\
pole          & \R{0.7529} & 0.7270                      & 0.7550                      & 0.7205                      & \R{0.7565} & 0.7447                      & \R{0.7680} & 0.7438                      \\
\end{longtable}

\begin{longtable}[!h]{lcccccccc}
\toprule
Class & \multicolumn{2}{c}{OF ConvNeXt-L Base} & \multicolumn{2}{c}{OF ConvNeXt-L Scale} & \multicolumn{2}{c}{Segformer B5 Base} & \multicolumn{2}{c}{Segformer B5 Scale} \\
    & \(\mu\) Avg. & M Avg. & \(\mu\) Avg. & M Avg. & \(\mu\) Avg. & M Avg. & \(\mu\) Avg. & M Avg. \\
\midrule
wall            & 0.5739                      & 0.6680                      & 0.5838                      & 0.6911                      & 0.8406                      & 0.8069                      & 0.8743                      & 0.8182                      \\
building        & 0.5608                      & 0.6452                      & 0.5531                      & 0.6701                      & 0.8300                      & 0.8045                      & 0.8605                      & 0.8039                      \\
sky             & 0.5670                      & 0.7676                      & 0.5786                      & 0.7708                      & 0.9233                      & \G{0.9184} & 0.9339                      & \G{0.9159} \\
floor           & 0.5547                      & 0.7259                      & 0.5663                      & 0.7346                      & 0.8556                      & 0.8752                      & 0.8674                      & 0.8720                      \\
tree            & 0.6203                      & 0.7113                      & 0.6428                      & 0.7331                      & 0.8665                      & 0.7693                      & 0.8745                      & 0.7638                      \\
ceiling         & 0.5189                      & 0.7149                      & 0.5099                      & 0.7141                      & 0.8375                      & 0.8589                      & 0.8665                      & 0.8536                      \\
\end{longtable}

\subsection*{Qualitative results on ADE20K}
\begin{figure}[!h]
    \begin{subfigure}{0.163\textwidth}
    \includegraphics[width=\textwidth]{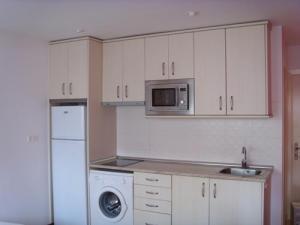}
    \end{subfigure}
    \begin{subfigure}{0.163\textwidth}
    \includegraphics[width=\textwidth]{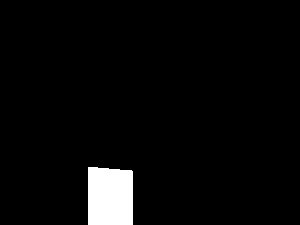}
    \end{subfigure}
    \begin{subfigure}{0.163\textwidth}
    \includegraphics[width=\textwidth]{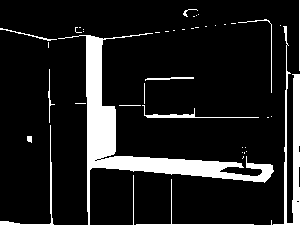}
    \end{subfigure}
    \begin{subfigure}{0.163\textwidth}
    \includegraphics[width=\textwidth]{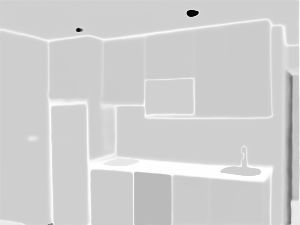}
    \end{subfigure}
    \begin{subfigure}{0.163\textwidth}
    \includegraphics[width=\textwidth]{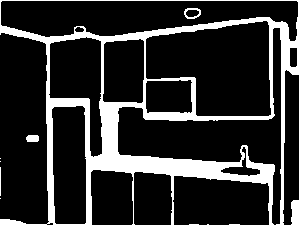}
    \end{subfigure}
    \begin{subfigure}{0.163\textwidth}
    \includegraphics[width=\textwidth]{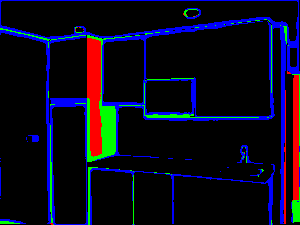}
    \end{subfigure} \\
    \begin{subfigure}{0.163\textwidth}
    \includegraphics[width=\textwidth]{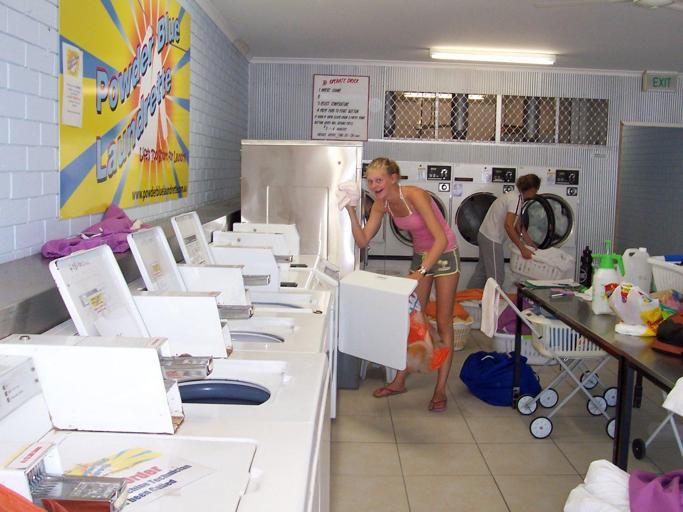}
    \end{subfigure}
    \begin{subfigure}{0.163\textwidth}
    \includegraphics[width=\textwidth]{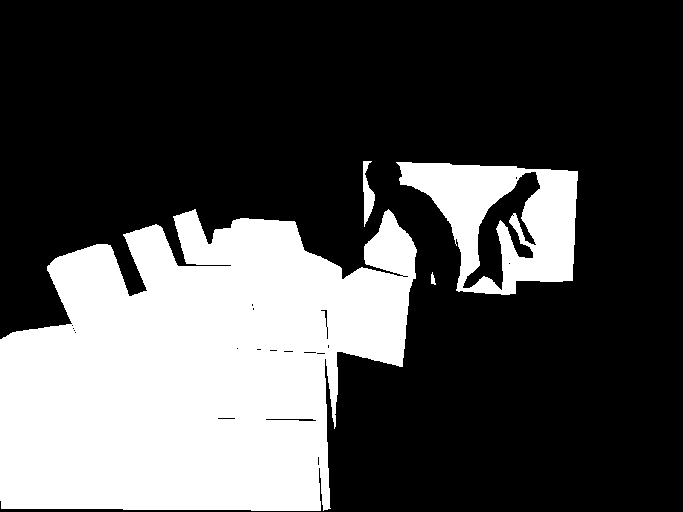}
    \end{subfigure}
    \begin{subfigure}{0.163\textwidth}
    \includegraphics[width=\textwidth]{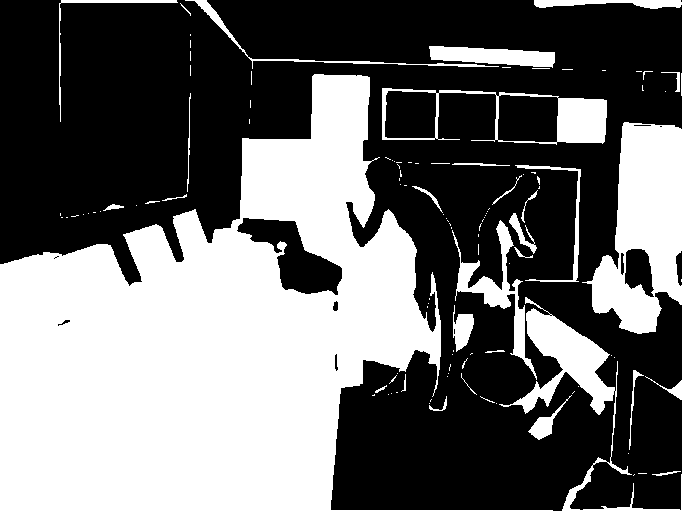}
    \end{subfigure}
    \begin{subfigure}{0.163\textwidth}
    \includegraphics[width=\textwidth]{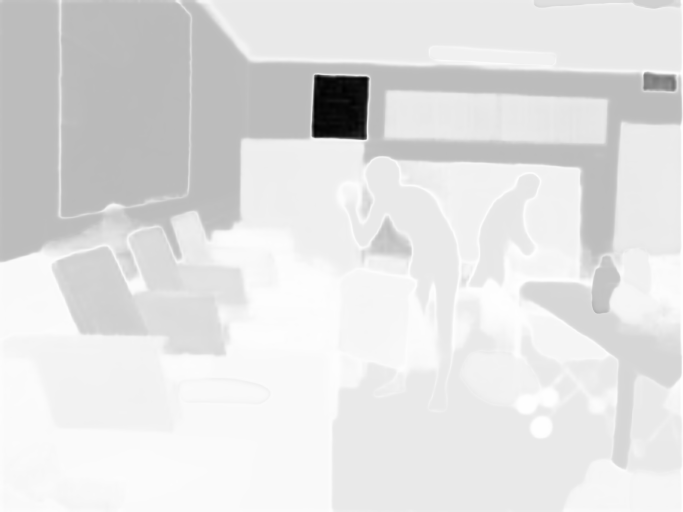}
    \end{subfigure}
    \begin{subfigure}{0.163\textwidth}
    \includegraphics[width=\textwidth]{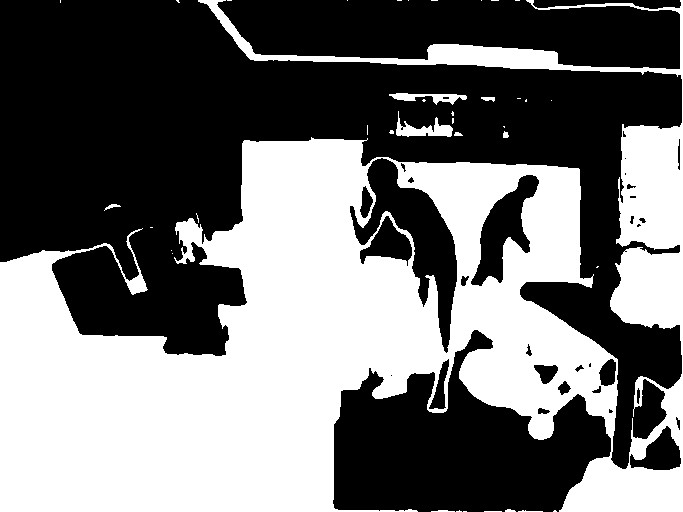}
    \end{subfigure}
    \begin{subfigure}{0.163\textwidth}
    \includegraphics[width=\textwidth]{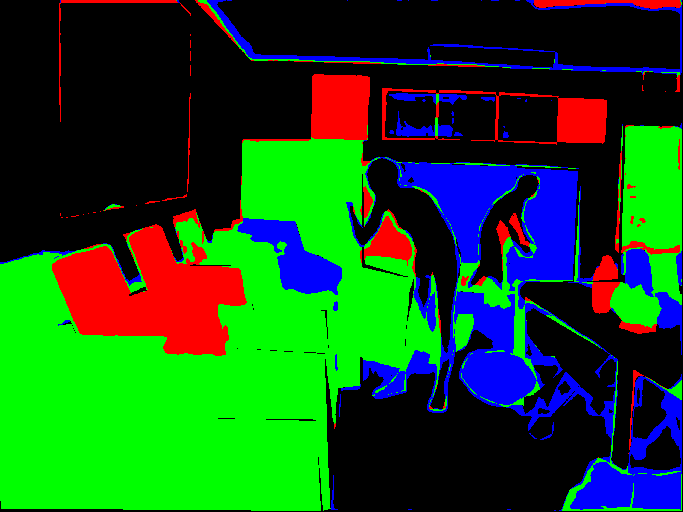}
    \end{subfigure} \\\\\\
    \begin{subfigure}{0.163\textwidth}
    \includegraphics[width=\textwidth]{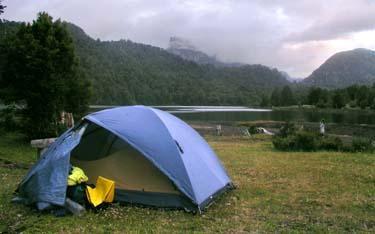}
    \end{subfigure}
    \begin{subfigure}{0.163\textwidth}
    \includegraphics[width=\textwidth]{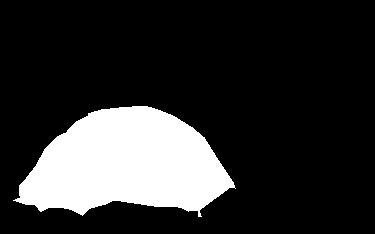}
    \end{subfigure}
    \begin{subfigure}{0.163\textwidth}
    \includegraphics[width=\textwidth]{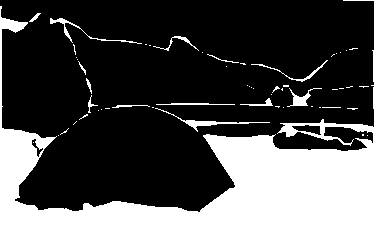}
    \end{subfigure}
    \begin{subfigure}{0.163\textwidth}
    \includegraphics[width=\textwidth]{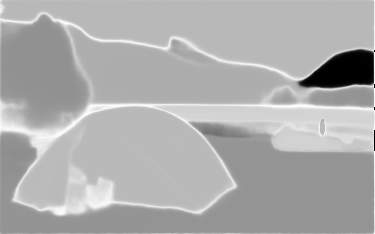}
    \end{subfigure}
    \begin{subfigure}{0.163\textwidth}
    \includegraphics[width=\textwidth]{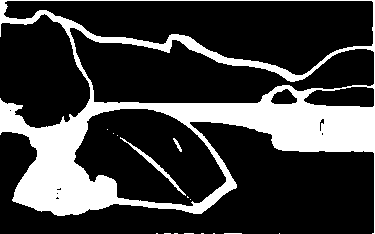}
    \end{subfigure}
    \begin{subfigure}{0.163\textwidth}
    \includegraphics[width=\textwidth]{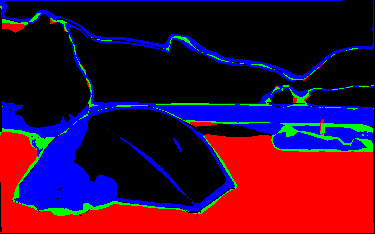}
    \end{subfigure} \\
    \begin{subfigure}{0.163\textwidth}
    \includegraphics[width=\textwidth]{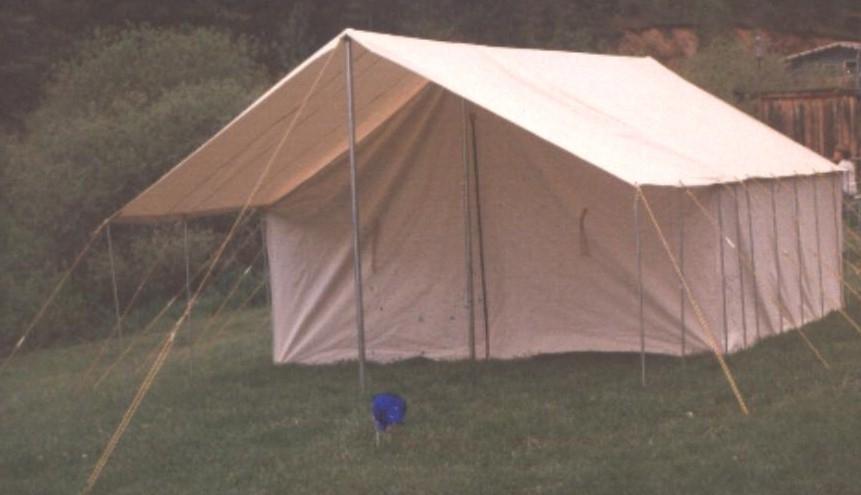}
    \end{subfigure}
    \begin{subfigure}{0.163\textwidth}
    \includegraphics[width=\textwidth]{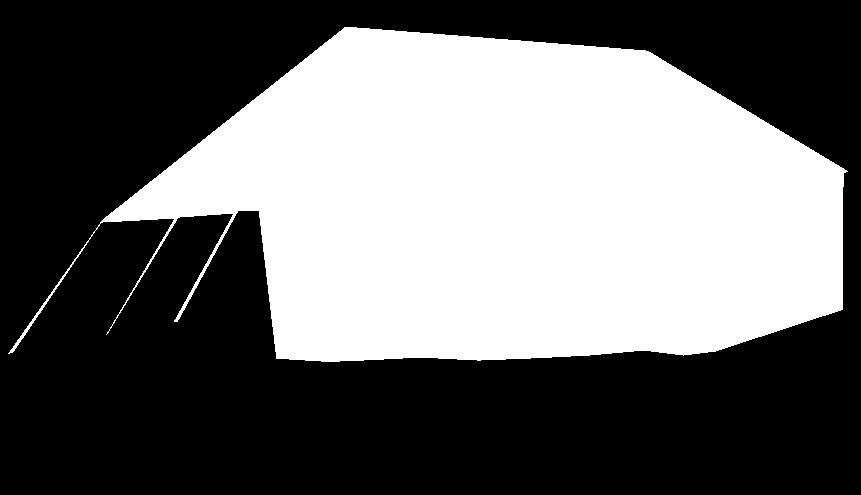}
    \end{subfigure}
    \begin{subfigure}{0.163\textwidth}
    \includegraphics[width=\textwidth]{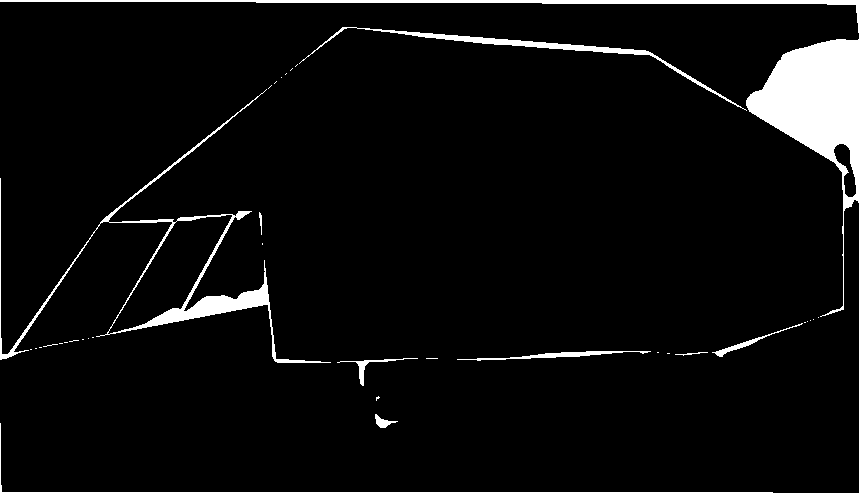}
    \end{subfigure}
    \begin{subfigure}{0.163\textwidth}
    \includegraphics[width=\textwidth]{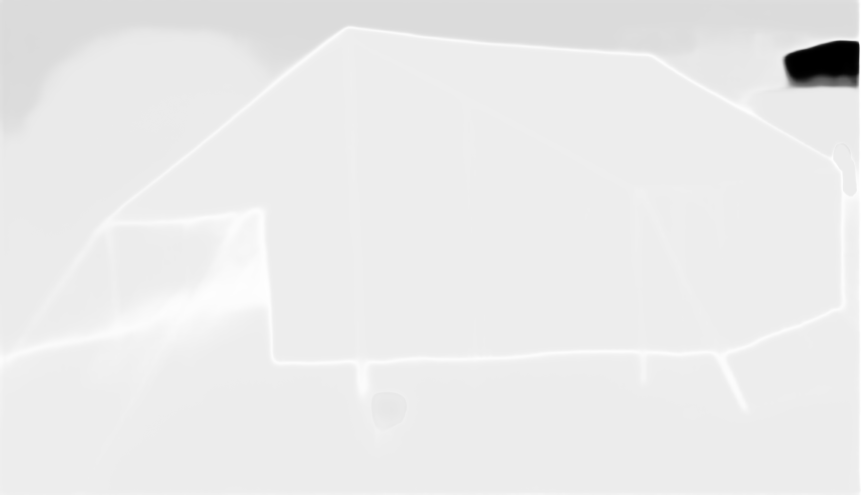}
    \end{subfigure}
    \begin{subfigure}{0.163\textwidth}
    \includegraphics[width=\textwidth]{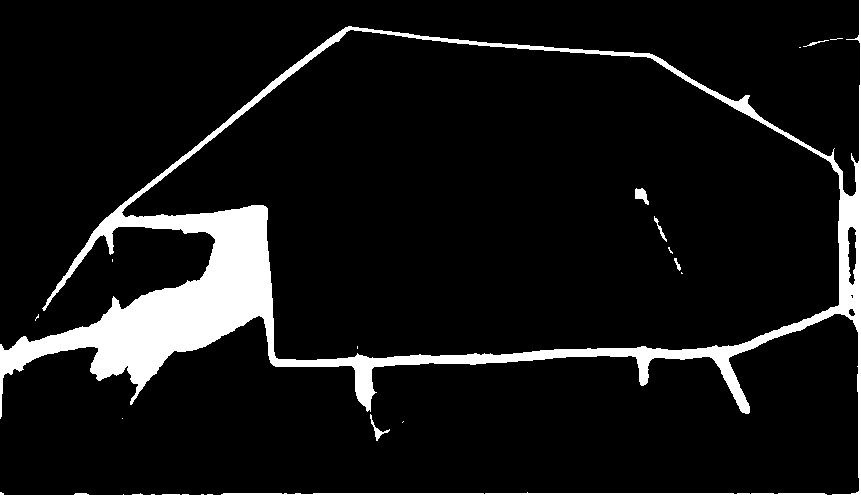}
    \end{subfigure}
    \begin{subfigure}{0.163\textwidth}
    \includegraphics[width=\textwidth]{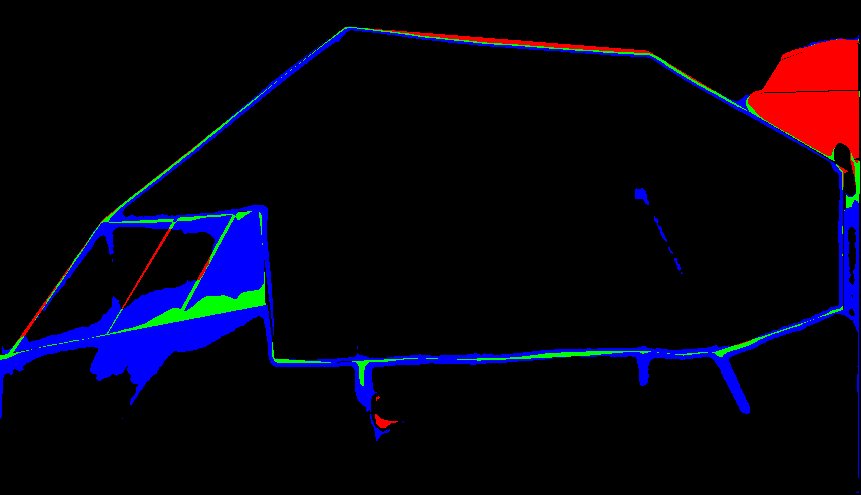}
    \end{subfigure}
    \caption{A few images showing a couple of best performing classes in the ADE20K dataset. The first two images include the washing machine class while the last two include the tent class. For each image, we show from left to right, the image, the highlighted class, misclassified pixels, entropy, thresholded entropy, detection mask.}
    \label{fig:best_classes}
\end{figure}

\begin{figure}[!h]
    \begin{subfigure}{0.163\textwidth}
    \includegraphics[width=\textwidth]{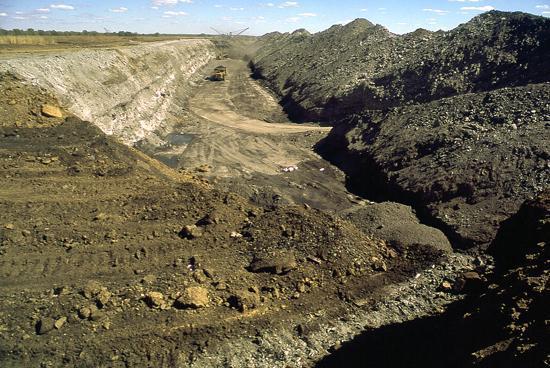}
    \end{subfigure}
    \begin{subfigure}{0.163\textwidth}
    \includegraphics[width=\textwidth]{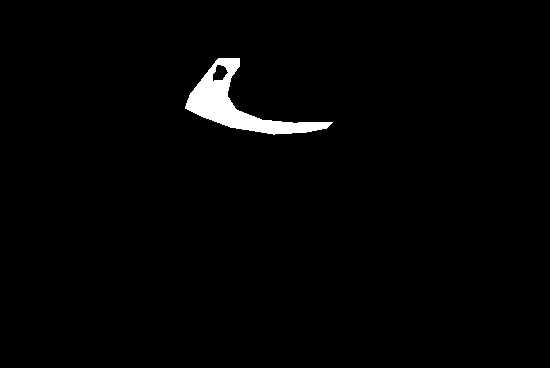}
    \end{subfigure}
    \begin{subfigure}{0.163\textwidth}
    \includegraphics[width=\textwidth]{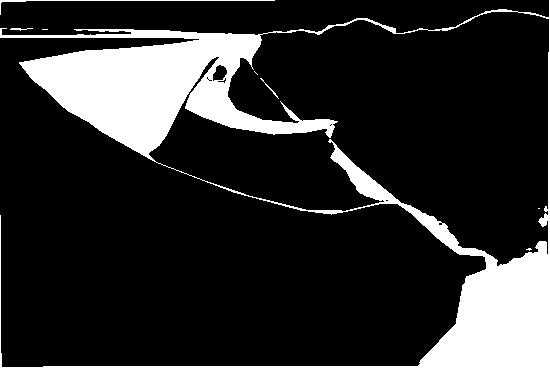}
    \end{subfigure}
    \begin{subfigure}{0.163\textwidth}
    \includegraphics[width=\textwidth]{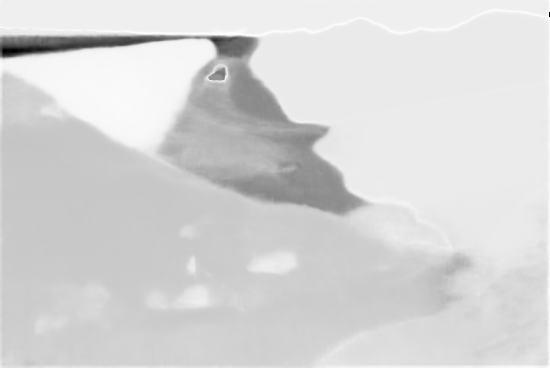}
    \end{subfigure}
    \begin{subfigure}{0.163\textwidth}
    \includegraphics[width=\textwidth]{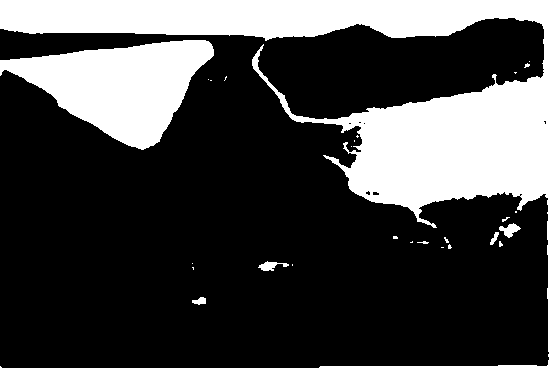}
    \end{subfigure}
    \begin{subfigure}{0.163\textwidth}
    \includegraphics[width=\textwidth]{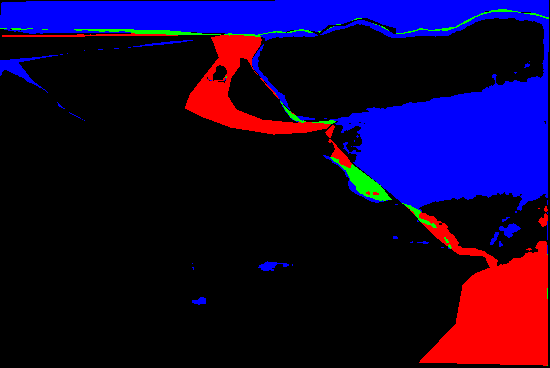}
    \end{subfigure} \\
    \begin{subfigure}{0.163\textwidth}
    \includegraphics[width=\textwidth]{images/dt91_2.jpg}
    \end{subfigure}
    \begin{subfigure}{0.163\textwidth}
    \includegraphics[width=\textwidth]{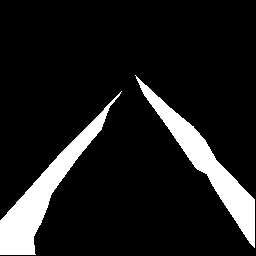}
    \end{subfigure}
    \begin{subfigure}{0.163\textwidth}
    \includegraphics[width=\textwidth]{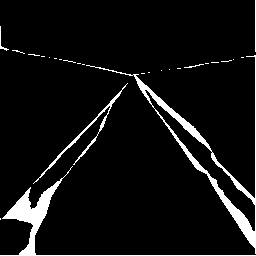}
    \end{subfigure}
    \begin{subfigure}{0.163\textwidth}
    \includegraphics[width=\textwidth]{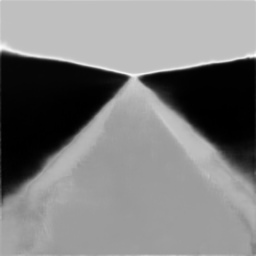}
    \end{subfigure}
    \begin{subfigure}{0.163\textwidth}
    \includegraphics[width=\textwidth]{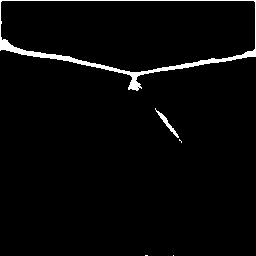}
    \end{subfigure}
    \begin{subfigure}{0.163\textwidth}
    \includegraphics[width=\textwidth]{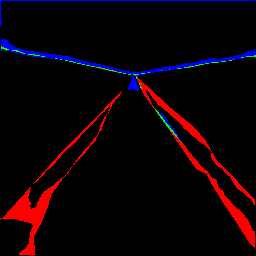}
    \end{subfigure} \\\\\\
    \begin{subfigure}{0.163\textwidth}
    \includegraphics[width=\textwidth]{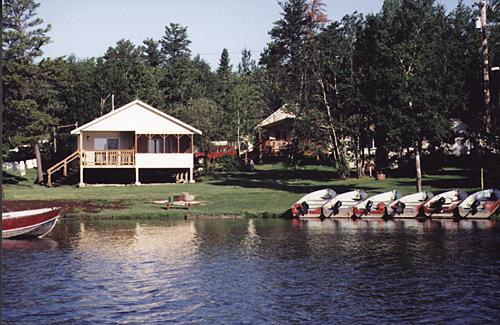}
    \end{subfigure}
    \begin{subfigure}{0.163\textwidth}
    \includegraphics[width=\textwidth]{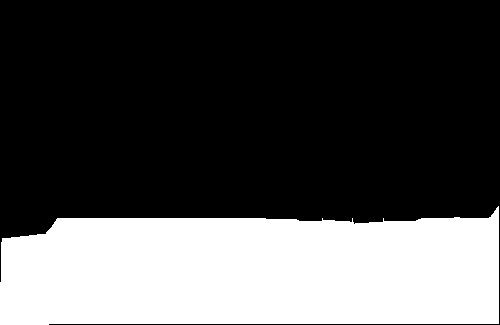}
    \end{subfigure}
    \begin{subfigure}{0.163\textwidth}
    \includegraphics[width=\textwidth]{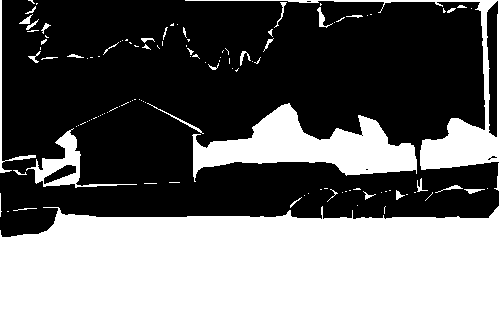}
    \end{subfigure}
    \begin{subfigure}{0.163\textwidth}
    \includegraphics[width=\textwidth]{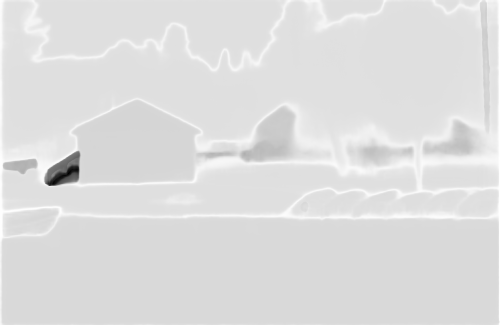}
    \end{subfigure}
    \begin{subfigure}{0.163\textwidth}
    \includegraphics[width=\textwidth]{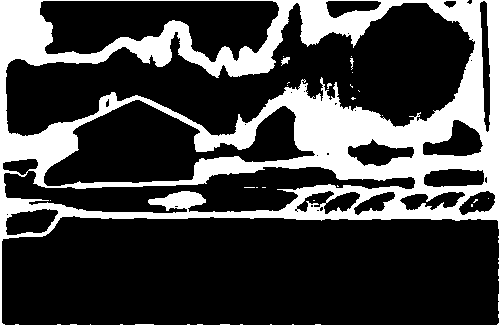}
    \end{subfigure}
    \begin{subfigure}{0.163\textwidth}
    \includegraphics[width=\textwidth]{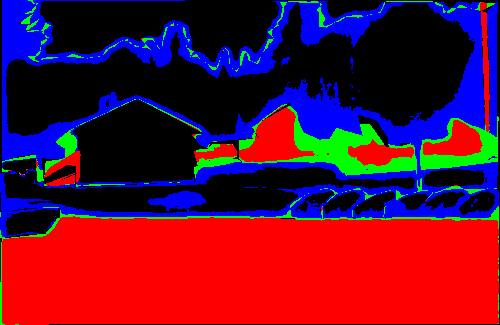}
    \end{subfigure} \\
    \begin{subfigure}{0.163\textwidth}
    \includegraphics[width=\textwidth]{images/lake128_2.jpg}
    \end{subfigure}
    \begin{subfigure}{0.163\textwidth}
    \includegraphics[width=\textwidth]{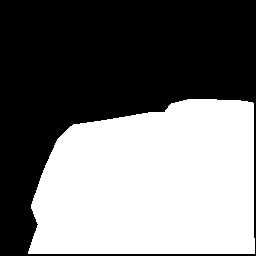}
    \end{subfigure}
    \begin{subfigure}{0.163\textwidth}
    \includegraphics[width=\textwidth]{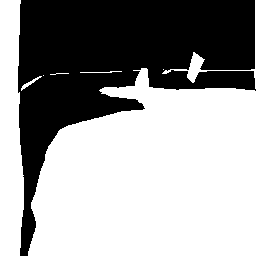}
    \end{subfigure}
    \begin{subfigure}{0.163\textwidth}
    \includegraphics[width=\textwidth]{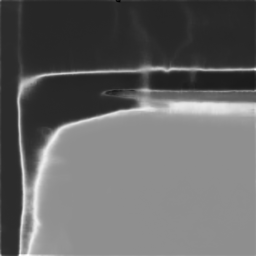}
    \end{subfigure}
    \begin{subfigure}{0.163\textwidth}
    \includegraphics[width=\textwidth]{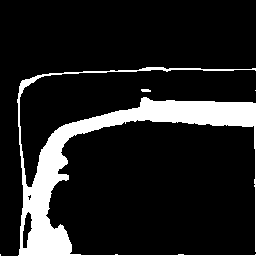}
    \end{subfigure}
    \begin{subfigure}{0.163\textwidth}
    \includegraphics[width=\textwidth]{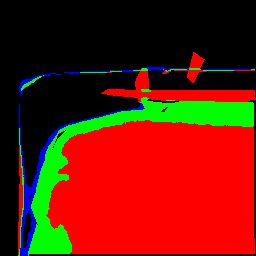}
    \end{subfigure}
    \caption{A few images showing a couple of worst performing classes in the ADE20K dataset. The first two images include dirt track class while the last two include the lake class. For each image, we show from left to right, the image, the highlighted class, misclassified pixels, entropy, thresholded entropy, detection mask.}
    \label{fig:worst_classes}
\end{figure}

We show a few qualitative results on the ADE20K dataset for a couple of best and worst performing classes in Figure~\ref{fig:best_classes} and Figure~\ref{fig:worst_classes}, respectively. The figures show the original image, the class label being looked at, misclassified pixels, entropy, thresholded entropy values (using the largest difference technique), and the detection performance (green areas show correctly identified misclassified pixels, blue areas are false positives and red areas are false negatives). All of these results are from the Base setting of OneFormer ConvNeXt-L model.

\end{document}